%% file: arxiv-ready.tex
\documentclass[10pt,twocolumn,letterpaper,nonatbib]{article}

\usepackage[pagenumbers]{cvpr}              

\usepackage{graphicx}
\usepackage{amsmath}
\usepackage{amssymb}
\usepackage{booktabs}
\usepackage{soul}

\usepackage{dsfont}
\usepackage{multirow}
\usepackage{enumerate}
\usepackage{enumitem}
\usepackage{xcolor}
\makeatletter
\newcommand{\dontusepackage}[2][]{%
  \@namedef{ver@#2.sty}{9999/12/31}%
  \@namedef{opt@#2.sty}{#1}}
\makeatother
\dontusepackage[bar,baz]{algorithmic}
\usepackage[noend]{algpseudocode}
\usepackage{algorithm}

\definecolor{dark-red}{rgb}{0.8,0.0,0}

\newcommand{\SKIP}[1]{}  
\newcommand{\eq}[1]{Eq.\;\eqref{eq:#1}}

\newcommand{\COB}{\color{black!10!blue}}
\newcommand{\COG}{\color{black!10!green}}
\newcommand{\COR}{\color{black!10!red}}

\newcommand{\vr}{\boldsymbol{r}}
\newcommand{\vmu}{\boldsymbol{\mu}}
\newcommand{\vphi}{\boldsymbol{\phi}}
\newcommand{\vvv}{\boldsymbol{v}}
\newcommand{\vv}{\boldsymbol{v}}
\newcommand{\mV}{\boldsymbol{V}}

\newcommand{\tM}{{\mathcal{M}}}
\newcommand{\mM}{\boldsymbol{M}}
\newcommand{\vm}{\boldsymbol{m}}
\newcommand{\vx}{\boldsymbol{x}}
\newcommand{\vy}{\boldsymbol{y}}
\newcommand{\tvx}{\boldsymbol{\tilde{x}}}

\newcommand{\mX}{\boldsymbol{X}}
\newcommand{\tmX}{\boldsymbol{\tilde{X}}}
\newcommand{\bmM}{\boldsymbol{\bar{M}}}
\newcommand{\vz}{\boldsymbol{z}}
\newcommand{\mbr}[1]{\mathbb{R}^{#1}}
\DeclareMathOperator*{\argmin}{arg\,min}

\newcommand{\valpha}{\boldsymbol{\alpha}}
\newcommand{\vtheta}{\boldsymbol{\theta}}

\newcommand{\vf}{\boldsymbol{f}}

\newcommand{\vh}{\boldsymbol{h}}
\newcommand{\vepsilon}{\boldsymbol{\epsilon}}
\DeclareMathOperator*{\spann}{Span}
\DeclareMathOperator*{\sign}{Sign}

\newtheorem{prop}{Proposition}

\newcommand{\appropto}{\mathrel{\vcenter{
  \offinterlineskip\halign{\hfil$##$\cr
    \propto\cr\noalign{\kern0.5pt}\sim\cr\noalign{\kern-2pt}}}}}

\makeatletter
\DeclareRobustCommand\onedot{\futurelet\@let@token\bmv@onedotaux}
\def\bmv@onedotaux{\ifx\@let@token.\else.\null\fi\xspace}
\def\eg{\emph{e.g}\onedot} 
\def\ie{\emph{i.e}\onedot} 
\def\cf{\emph{c.f}\onedot} 
\def\etc{\emph{etc}\onedot} \def\vs{\emph{vs}\onedot}
\def\wrt{w.r.t\onedot}

\def\etal{\emph{et al}\onedot}

\makeatother

\usepackage[pagebackref,breaklinks,colorlinks,bookmarks=false,urlcolor=black
,citecolor={green!60!black}]{hyperref}

\makeatletter
\@namedef{ver@everyshi.sty}{}
\makeatother
\usepackage[most]{tcolorbox}

\definecolor{beaublue}{rgb}{0.75, 0.85, 0.95}
\definecolor{blackish}{rgb}{0.2, 0.2, 0.2}

\definecolor{beaublue2}{rgb}{0.84, 0.9, 0.95}
\definecolor{blackish2}{rgb}{0.2, 0.2, 0.2}

\makeatletter
\def\ps@myheadings{%
    \let\@oddfoot\@empty\let\@evenfoot\@empty
    \def\@evenhead{\thepage\hfil\slshape\leftmark}%
    \def\@oddhead{{\slshape\rightmark}\hfil\thepage}%
    \let\@mkboth\@gobbletwo
    \let\sectionmark\@gobble
    \let\subsectionmark\@gobble
    }
  \if@titlepage
  \renewcommand\maketitle{\begin{titlepage}%
  \let\footnotesize\small
  \let\footnoterule\relax
  \let \footnote \thanks
  \null\vfil
  \vskip 60\p@
  \begin{center}%
    {\LARGE \@title \par}%
    \vskip 3em%
    {\large
     \lineskip .75em%
      \begin{tabular}[t]{c}%
        \@author
      \end{tabular}\par}%
      \vskip 1.5em%
    {\large \@date \par}
  \end{center}\par
  \@thanks
  \vfil\null
  \end{titlepage}%
  \setcounter{footnote}{0}%
}
\else
\renewcommand\maketitle{\par
  \begingroup
    \renewcommand\thefootnote{\@fnsymbol\c@footnote}%
    \def\@makefnmark{\rlap{\@textsuperscript{\normalfont\color{black}\@thefnmark}}}%
    \long\def\@makefntext##1{\parindent 1em\noindent
            \hb@xt@1.8em{%
                \hss\@textsuperscript{\normalfont\@thefnmark}}##1}%
    \if@twocolumn
      \ifnum \col@number=\@ne
        \@maketitle
      \else
        \twocolumn[\@maketitle]%
      \fi
    \else
      \newpage
      \global\@topnum\z@   
      \@maketitle
    \fi
    \thispagestyle{plain}\@thanks
  \endgroup
  \setcounter{footnote}{0}%
}
\makeatother

\makeatletter
\newcommand\fs@nobottomruled{\def\@fs@cfont{\bfseries}\let\@fs@capt\floatc@ruled
  \def\@fs@pre{}
  \def\@fs@post{}
  \def\@fs@mid{\kern2pt\hrule\kern2pt}%
  \let\@fs@iftopcapt\iftrue}
\makeatother

\usepackage[capitalize]{cleveref}
\crefname{section}{Sec.}{Secs.}
\Crefname{section}{Section}{Sections}
\Crefname{table}{Table}{Tables}
\crefname{table}{Tab.}{Tabs.}

\def\cvprPaperID{3798} 
\def\confName{CVPR}
\def\confYear{2022}

\DeclareSymbolFont{extraup}{U}{zavm}{m}{n}
\DeclareMathSymbol{\varheart}{\mathalpha}{extraup}{86}
\DeclareMathSymbol{\vardiamond}{\mathalpha}{extraup}{87}

\begin{document}

\title{Manifold Learning Benefits GANs\vspace{-0.3cm}}

\author{%
  Yao Ni\textsuperscript{\textasteriskcentered}$^{\!, \dagger}$, \quad Piotr Koniusz\thanks{Equal contribution.$\qquad$Accepted by CVPR 2022.$\qquad\qquad\qquad\quad$\linebreak\indent$\,^{_\clubsuit}\!$Brain team (richardnock@google.com).}$\;^{,\S,\dagger}$, \quad Richard Hartley$^{\dagger,\vardiamond}$, \quad Richard Nock$^{\vardiamond\!,\clubsuit,\dagger}$\\\vspace{0.3cm}
  $^{\dagger}$The Australian National University \quad 
   $^\S$Data61/CSIRO  \quad $^{\vardiamond}$Google Research \\
	\vspace{-0.5cm}
  firstname.lastname@anu.edu.au \\
	\vspace{-0.3cm}
}

\maketitle

\begin{abstract}

In this paper\footnote{Code: \url{https://github.com/MaxwellYaoNi/LCSAGAN}.}, we improve Generative Adversarial Networks by incorporating a
manifold learning step into the discriminator. We consider locality-constrained
linear and subspace-based manifolds\footnote{The coding spaces considered in this paper are loosely termed
manifolds.  In most cases they are not manifolds in the strict
mathematical sense, but rather topological spaces such as 
varieties, or simplicial complexes.
The word will be used only in an informal sense.
}, and locality-constrained non-linear
manifolds. In our design, the manifold learning and coding steps are intertwined
with layers of the discriminator, with the goal of attracting intermediate
feature representations onto manifolds. We adaptively balance the discrepancy
between feature representations and their manifold view, which is a
trade-off between denoising on the manifold and refining the manifold. We
find that locality-constrained non-linear manifolds outperform 
linear manifolds due to their non-uniform density and smoothness. 
We also substantially outperform  state-of-the-art baselines.

\end{abstract}

\vspace{-0.2cm}
\section{Introduction}\label{sec:intro}
Generative Adversarial Networks (GANs) \cite{GAN} are 
powerful models for image generation \cite{PGAN,BigGAN,StyleGANV2}, sound generation \cite{WaveGAN}, image stylization
\cite{li2016precomputed} and destylization \cite{fatima_dicta,fatima_wacv18,ShiriYPHK19,fatima_ijcv}, super-resolution
\cite{yu2018face}, feature generation \cite{xian2018feature, MetaGAN}, \etc. %
The original GAN  learns to generate  images 
\cite{wang2018high,PGAN,BigGAN,StyleGANV2,StyleGAN} 
by performing the following min-max game:
\vspace{-0.2cm}
\begin{equation}
    \min\limits_{\vtheta_{G}} \max\limits_{\vtheta_{D}}
\mathcal{J}(D_{|\vtheta_{D}}, G_{|\vtheta_{G}}),
    \label{eq:gan1}
\vspace{-0.2cm}
\end{equation}
where $\mathcal{J}(\cdot)\!=\!\mathds{E}_{\vx\sim
p_x(\vx)}\!\log(D(\vx;\vtheta_{D}))+\mathds{E}_{\vz\sim
p_z(\vz)}\!\log(1\!-\!D(G(\vz)))$. \eq{gan1} updates parameters
$\vtheta_{D}$ of discriminator $D(\vx;\vtheta_{D})$ to discriminate between
samples from the data distribution $p_x(\vx)$  and generative distributions $p_g(G)$. Simultaneously, parameters $\vtheta_{G}$ of  generator
$G(\vz;\vtheta_{G})$ are updated to fool the discriminator $D$. Thus, the
noise distribution $p_z(\vz)$ becomes mapped to $p_x(\vx)$ via generator $G$.


\begin{figure}
\includegraphics[width=1\linewidth]{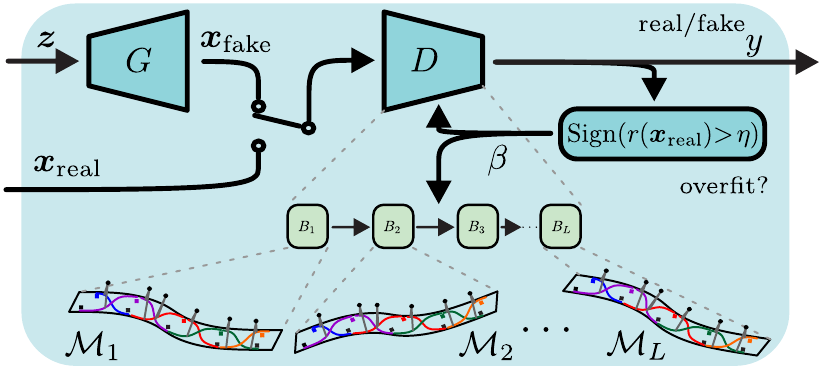}
%
\vspace{-0.5cm}
\caption{Our GAN pipeline. We equip the discriminator with residual blocks
$B_1,\cdots,B_L$, each containing standard CNN operations, \eg convolutions,
ReLU, downsampling, residual link \etc, and the manifold learner
$\mathcal{M}_l$. Metaparameter $\beta$ controls the degree of mixing block conv.
features with their view recovered from the manifold. The `overfit?' detector
increases $\beta$ when overfitting to $\vx_\text{real}$ is suspected, which
boosts the impact of manifold learners.}
\label{fig:gan_pipe}
\vspace{-0.5cm}
\end{figure}

However, GANs typically suffer from three problems: \textbf{1)} the training
instability \cite{kodali2017convergence}, \textbf{2)} the so-called mode
collapse \cite{roth2017stabilizing}, and \textbf{3)}  overfitting of the
discriminator \cite{Webster_2019_CVPR}.

The training instability is an imbalanced competition of the
generator and the discriminator due to 
non-overlapping support between the model distribution and the data distribution
\cite{kodali2017convergence, PGAN,Chu2020Smoothness}, leading to poor quality
of generated data. Mode collapse has to do with sharply rising
gradient around undesirable local equilibria \cite{roth2017stabilizing,kodali2017convergence}, resulting in generation of the
same image. Finally, 
discriminator overfitting leads to excessive
memorization and poor generalization. 

Indeed, with an excessive number of parameters, the discriminator may memorize
the training data instead of learning a meaningful distribution, 
leading to a high
real/fake classification accuracy on the training dataset 
and a low accuracy on the validation split  
\cite{BigGAN, ADA, DiffAug}. Webster \etal \cite{Webster_2019_CVPR} argue such a phenomenon mainly affects the discriminator and is
undetectable in the generator, with the exception of hybrid adversarial and
non-adversarial methods \cite{pmlr-v80-bojanowski18a} 
which impose  
the so-called consistency loss  on a generator. 

We also observed discriminator overfitting in baseline models \eg, doubling the number of parameters of discriminator resulted in training and validation FID scores of baseline GANs
diverging at some intermediate training stage. 

Thus, to reduce overfitting of the discriminator, we propose  a
data-driven feature manifold-learning step and intertwine it with layers of the discriminator. In
this way, the discriminator learns the feature manifold at different levels of
object abstraction, from fine to coarse, which limits the complexity of
parameter space and separates the signal from noise as 
both generated and real data are expressed 
on a common manifold. As a result, the generator diversifies the generated
patterns according to their view on the manifold, on which the discriminator
operates. The min-max game operates on a gradually learnt manifold (see Fig. \ref{fig:gan_pipe}). 
%
%
%

\vspace{0.1cm}
Our contributions are threefold:
\renewcommand{\labelenumi}{\roman{enumi}.}
\vspace{-0.15cm}
\hspace{-1.0cm}
\begin{enumerate}[leftmargin=0.6cm]
\item We intertwine  locality-constrained and subspace-based feature
encoding and dictionary learning steps  
\cite{liu_sadefense,me_ATN} with blocks of the GAN discriminator to exploit manifold learning in an end-to-end scenario.
\vspace{-0.25cm}
\item We employ a balancing term to help blocks of the discriminator
 learn the data-driven manifold from the 
encoder intertwined with them, while permitting some degree of freedom in the
vicinity of that manifold (\S \ref{sec:pro}).$\!\!\!\!\!\!$
\vspace{-0.25cm}
\item 
We  show that  
locality-constrained soft assignment coding (the best coder in our experiments) acts as a locally flexible denoiser \cite{denoising_JMLR} due to its Lipschitz continuity which we  control to vary its operating mode between the ordinary
k-means quantization and locality-constrained linear coding. This
setting admits quantization of some feature space parts while
approximately preserving linearity of other feature space parts (\S\ref{sec:lcsa}). 
\end{enumerate}


\vspace{-0.15cm}
For contribution (i), 
we investigate Sparse Coding (SC) 
\cite{lee_sparse, yang_sparse}, 
Non-negative Sparse Coding (SC\textsubscript{+}) \cite{nnsc},
Orthogonal Matching Pursuit (OMP) \cite{pati_omp,dmallat_omp},
Locality-constrained Linear Coding (LLC) \cite{wang_llc}, Soft Assignment (SA)
\cite{bilmes_gmm,gemert_kernel}, and Locality-constrained Soft Assignment (LCSA)
\cite{liu_sadefense,me_SAO,me_ATN,me_tensor_tech_rep,me_tensor}, and Hard Assignment (HA)
\cite{steinhaus_ha,csurka04_bovw}. We provide formulations and discussion on properties  of each
 coder 
in \S \ref{sec:pre}.

\section{Problem Formulation}
\label{sec:pro}

Figure \ref{fig:gan_pipe} shows our pipeline (we skip conditional
cues for brevity). We build on BigGAN \cite{BigGAN}, OmniGAN
\cite{zhou2020omni}, MSG-StyleGAN \cite{karnewar2020msg}, StyleGAN2\cite{StyleGANV2} but we equip the discriminator with the manifold learner
which is metacontrolled to reduce overfitting. 
%
%
In \S \ref{sup:limited_data} of the supplementary material, we also study the combination of our method with DA \cite{DiffAug}, ADA \cite{ADA} and LeCamGAN \cite{lecamgan} in limited data scenario.

\begin{figure}
\includegraphics[width=1\linewidth]{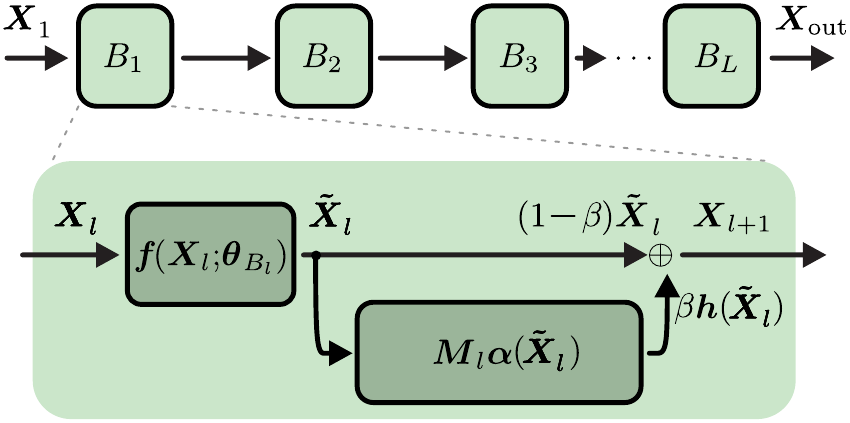}
%
\vspace{-0.5cm}
\caption{Blocks $B_1,\cdots,B_L$ of our discriminator contain a standard block denoted by $\vf$ intertwined with the manifold
learner. Metaparameter $\beta$ controls the mixing balance between $\vf$ and
$\vh$.}
\label{fig:res_blcoks}
\vspace{-0.4cm}
\end{figure}


The discriminator $D(\vx;\vtheta_{D})$ of the GAN in \eq{gan1} classifies
input images as real or fake. Many architectures 
exist in  the literature \eg, GAN \cite{GAN} uses the discriminator
based on a convolutional network, whereas recent architectures \eg, 
BigGAN \cite{BigGAN}, OmniGAN \cite{zhou2020omni}, MSG-StyleGAN\cite{karnewar2020msg} and StyleGAN2\cite{StyleGANV2} use residual
discriminators with $L$ residual blocks \eg, see BigGAN \cite{BigGAN} (their Fig. 16).

Let\footnote{Our notations are explained in \S\ref{sup:notations} of the
supplementary material.} $\vf:\mbr{d\times N}\!\times\mbr{|\vtheta_{B}|}\rightarrow\mbr{d'\times N'}$
(where $|\vtheta_B|$ is the size of the set of parameters $\vtheta_B$)
be a function realized by a single discriminator block with  parameters
$\vtheta_{B}$, where $d$ and $d'$ are the number of input/output channels,
$N\!=\!WH$ and $N'\!=\!W'H'$ are the number of input/output spatial locations in
feature maps of 
the block.  Often, $N'$ may equal $N$.

We introduce an encoding function $\vh\!:\!\mbr{d'\times
N'}\!\rightarrow\!\mbr{d'\times {N'}}$ that maps $\mbr{d'\times
N'}$ into a subset, usually nowhere dense or of small volume
in $\mbr{d'\times N'}$, which we sometimes refer to as the {\em feature space}.
In the cases we consider this mapping derives from a mapping
$\mbr{d'}\rightarrow \mbr{d'}$ applied independently and equally over the
second dimension $\mbr{N'}$.  The encoding introduces an error, measured by
\(
\|\vh (\mX) - \mX\|_F \le \epsilon
\)
where $\epsilon$ is called the {reconstruction error}.

In dictionary-based encoding, the function $\vh$ relies on a dictionary
$\mM\!=\![\vm_1,\cdots,\vm_k]\!\in\!\mbr{d'\times k}$ containing $k$ column
vectors, the so-called dictionary atoms (sometimes called anchors),
defining the underlying manifold $\tM$, and $k\!\gg\!d'$ ensures the dictionary
is overcomplete.
%
Then, after solving the optimization problem, 
\vspace{-0.1cm}
\begin{equation}
(\valpha,\mM)=\argmin_{\valpha',\mM'}  \|\mX\!-\!\mM'\valpha'\|_F^2+\kappa\,
\Omega(\valpha', \mM', \mX),
\label{eq:enc1}
\vspace{-0.3cm}
\end{equation}
where $\valpha\!\equiv\![\valpha_1,\cdots,\valpha_{N'}]\!\in\!\mbr{k\times
N'}\!$, the function $\vh$ is defined by $\vh (\mX) = \mM \valpha$.
Since $\valpha$ depends on $\mX$, we shall commonly write it as $\valpha(\mX)$.
The mapping $\vh$ maps $\mbr{d'}$ into a subset $\tM$ of $\mbr{d'}$,
that we will call the feature manifold (or simply manifold).

%

The choice of $\Omega(\valpha', \mM', \mX)$ realizes some
desired constraints via regularization (with $\kappa\!>\!0$) for example, $\Omega(\valpha',
\mM', \mX)\!=\!\|\valpha'\|_1$ encourages sparsity of $\valpha$,
%
%
while $\Omega(\valpha', \mM', \mX)\!=\!
\sum_{n}[\|\vx_n-\vm_1\|_2^2,\cdots,\|\vx_n-\vm_k\|_2^2]^T]\valpha_{n}|$
encourages locality to express each $\valpha_n$ \wrt
$\spann(\vm_1,\cdots,\vm_{k'})$, where $\vm_1,\cdots,\vm_{k'}$ are the $k'$
nearest neighbors of $\vx_n$.  

We intertwine the encoding step with blocks of the discriminator as
follows:
\vspace{-0.2cm}
\begin{equation}
\mX_{l+1}\!=\!
(1\!-\!\beta)\tmX_{l} +\beta\, \vh_{l}(\tmX_{l})
\label{eq:enc2}
\vspace{-0.2cm}
\end{equation}
%
where $\tmX_{l} = \vf(\mX_{l};\vtheta_{B_l})$
and $\vh_{l}$ is the encoding function introduced just above, expressed in terms of a dictionary 
$\mM_{l}$, 
whereas
$\{\vtheta_{B_l}\}_{l=1}^L$ and $\{\mM_l\}_{l=1}^L$ are parameters of blocks of
the discriminator and dictionaries for layers $1, \ldots, L$
respectively. Figure \ref{fig:res_blcoks} illustrates \eq{enc2}
applied to blocks $B_1,\cdots,B_L$ of the discriminator.

At the same time, we prevent \eq{enc2} from becoming a residual link by
adding a reconstruction loss to the GAN objective:
{
\setlength{\belowdisplayskip}{0.20cm}
\vspace{-0.7cm}
\begin{align}
\begin{split}
\mathcal{J}_\text{prox}\!&=\!\frac{\gamma}{L}
\sum_{l=1}^L\vepsilon(\tmX_{l}; \mM_{l}),\quad\text{ where }\\
\;\vepsilon(\tmX_{l}; \mM_{l})\!&=
\| \tmX_{l} - \vh_{l}(\tmX_{l}) \|_F^2 ~.
\label{eq:prox}
\end{split}
\end{align}
Metaparameters $(\beta,\gamma)$ control the mixing balance and the proximity
between $\vf$ and $\vh$. 
}

\vspace{0.05cm}
\noindent\textbf{Meta-adaptation of $\boldsymbol{(\beta,\gamma)}$.}
Discriminator overfitting can be detected with a  hypothesis test on the
expectation over decisions $r(\vx_\text{real})\!=\!\mathds{E}[\sign(D(\vx_
\text{real}))]$ \wrt samples $\vx_\text{real}$, defined as 
$\sign(r(\vx_\text{real})\!>\!\eta)\!\in\!\{-1,0,1\}$, where $\eta\!=\!0.5$ is
the threshold  whose violation indicates potential overfitting, as the discriminator
becomes increasingly good at distinguishing real datapoints \cite{ADA}. Thus, to
update $(\beta,\gamma)$, we apply:
{\setlength{\abovedisplayskip}{0.10cm}
\setlength{\belowdisplayskip}{0.2cm}
\vspace{-0.3cm}
\begin{align}
& \beta_{t+1}\!=\!\beta_t\!+\!\Delta_\beta\!\cdot\!\sign\left(r(\vx_\text{real})
\!>\!\eta\right),\label{eq:beta}\\
& \gamma_{t+1}\!=\!\gamma_0\!+\!\Delta_\gamma\!\cdot\!\beta_{t+1},
\label{eq:gamma}
\end{align}
where $\beta_0\!=\!0.1$. $\Delta_\beta\!=\!0.001$ ensures a gradual
change of $\beta$ by increasing contributions from $\vh(\tmX_{l})$ in \eq{enc2}
when overfitting is detected, and increasing contributions from $\tmX_{l}$  in \eq{enc2} when overfitting vanishes. By setting $\gamma_0\!=\!0.1$, we
ensure that the proximity loss in \eq{prox} is always enabled, and
$0.01\!\leq\!\Delta_\gamma\!\leq\!3$ controls the strength of proximity.}

\begin{figure}
\begin{center}
   \includegraphics[width=0.95\linewidth]{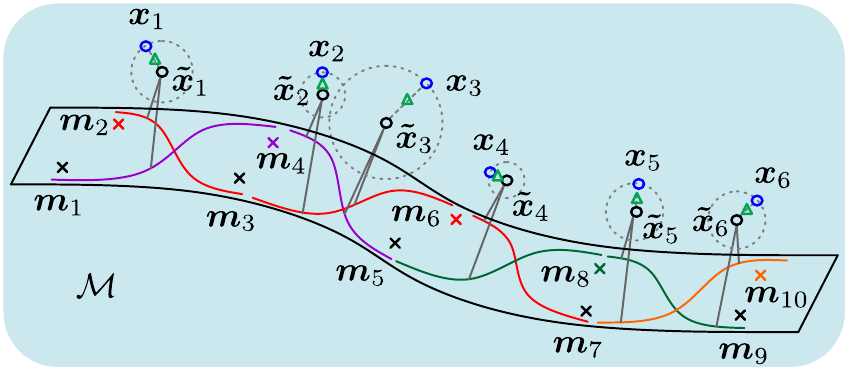} 
\end{center}
%
\vspace{-0.5cm}
\caption{Our manifold setting. Atoms
$\vm_1,\cdots,\vm_k$ (crosses $\times$) define the geometry of the manifold. Samples
$\vx_1,\cdots,\vx_{N'}$ (blue {\COB$\circ$}) are projected onto the manifold
$\tM$ via function $\boldsymbol{\alpha}(\vx)$ and then recovered via $\vh(\vx)$, which produces
recovered samples  $\tvx_1,\cdots,\tvx_{N'}$ (black $\circ$). The grey circles
indicate  $L_2$ balls imposed by the proximity operator in \eq{prox} while green triangles {\COG $\triangle$} are trade-off samples
between $\vf$ and $\vh$, controlled by $\beta$ within $L_2$ balls. We note
that radii of $L_2$ balls are controlled by $\gamma$ but each ball can be
larger or smaller depending whether this benefits the discrimination loss. Thus,
where desired, trade-off samples are refined by  $\vf$ \wrt $\vh$ within
$L_2$ balls. Curly lines (sigmoid-like {\COR$\backsim$}) indicate we use
locally the sigmoid non-linearity.}
\label{fig:res_man1}
\vspace{-0.5cm}
\end{figure}

\vspace{0.05cm}
\noindent\textbf{Discussion. } Figure \ref{fig:res_man1} shows our manifold
setting, which exploits the interplay between \eq{enc2} and
\eq{prox}. We alternately learn encoding of samples $\mX_l$ on manifold
$\tM_l$ and refine  dictionary 
$\mM_l$. The proximity operator in
\eq{prox} encourages samples $\tmX_l$ to stay in the proximity of
their recovered view $\vh(\tmX_l)$ by controlling $L_2$ balls around $\vh(\tmX_l)$. \eq{enc2} interpolates between $\vh(\tmX_l)$ and $\tmX_l$ within $L_2$ balls.
We opt for such a design as (i) $\vf$ may refine $\vh$ if
the discriminator loss spotting real/fake inputs deems it useful, (ii)
$\vh$ is preferred when overfitting is detected, whereas $\vf$ introduces
refined patterns otherwise, (iii) $\vf$ is encouraged to learn from the piecewise-smooth $\vh$ 
(see \S \ref{sec:lcsa}). 

\section{Related Works}

\vspace{0.05cm}
\noindent\textbf{Modern GANs.} Recent GANs build on GAN \cite{GAN} or DCGAN \cite{radford2015unsupervised} by improving generator or 
discriminator. 
%
A residual model  \cite{gulrajani2017improved} was improved by adding a self-attention block 
\cite{SAGAN}.  Progressive GAN \cite{PGAN} 
uses several levels of layers for 
increasingly finer image resolution. StyleGAN \cite{StyleGAN} maps the input to
an intermediate latent  space and controls the generator through
adaptive instance normalization (so-called AdaIN).  MSG-GAN 
passes multi-scale gradients from the discriminator to the generator
\cite{karnewar2020msg}. 

\vspace{0.05cm}
\noindent\textbf{Improving GANs.} To address training instability, mode collapse
and overfitting, researchers study (i) loss
and distance formulations, (ii) regularization mechanisms and penalties, and (iii) architectural modifications.

 Wasserstein GAN (Arjovsky \etal \cite{arjovsky2017wasserstein})  
 enjoys a good training stability. 
It was further improved by Gulrajani \etal \cite{gulrajani2017improved} 
by penalizing the norm of gradient of the
critic. 
Mean \& Covariance GAN
\cite{mroueh2017mcgan}  matches  the generated and real data
distributions with first- and second-order statistics. 
A Maximum
Mean Discrepancy GAN \cite{NIPS2017_dfd7468a}  matches 
distributions in the Reproducing Kernel Hilbert Space (RKHS). 


Spectral Normalization GAN \cite{SNGAN} applies   normalization on weights to stabilize  the
discriminator. Spectral Regularization GAN \cite{liu2019spectral}
performs detection of so-called spectral collapse. 
Disconnected Manifold GAN \cite{khayatkhoei2018disconnected} assumes that
natural images lie on a union of disjoint manifolds. 
 Feature Quantization GAN (FQGAN) \cite{FQGAN} quantizes 
features of discriminator into a k-means based dictionary. The Denoising Feature Matching GAN \cite{warde2016improving} encourages proximity between the output of the
generator and a denoising auto-encoder. 

Our work differs from the Disconnected Manifold GAN which models an entire
image distribution as a union of  non-explicit manifolds (a collection of generators).  
FQGAN  imposes quantization on features of the discriminator and Denoising Feature Matching GAN learns a denoising
auto-encoder on 
real images to apply it on
generated images.

In contrast, we model coarse-to-fine features extracted from multiple blocks of
the discriminator, which capture different semantic levels of abstraction. We
encourage these features to lie on explicit locality-constrained
non-linear manifolds (each block of our discriminator has its own learner). 
We adaptively control the mixing levels of features and their
views recovered from manifolds, and the smoothness of manifolds to  
prevent 
overfitting.


\vspace{0.05cm}
\noindent\textbf{Limiting Overfitting.}  Augmentations (rotations, clipping) \cite{CRGAN, DiffAug, jeong2021training, tran2021data} can limit overfitting, 
however, augmentation artifacts  leak into the generated images
\cite{ICRGAN, ADA}. Injecting noises into
the discriminator \cite{CAGAN, CTGAN} via 
dropout \cite{srivastava2014dropout} forms an  ensemble network, whereas we equip the discriminator with a data-manifold learner whose smoothness we  control. 
%

\section{Preliminaries}
\label{sec:pre}

Below, we explain  GAN pipelines on which we build, and feature
encoding and dictionary learning, our 
key tools.

\subsection{Baseline GANs}
\label{sec:base_gans}

\vspace{0.05cm}
\noindent\textbf{BigGAN} \cite{BigGAN} 
combines a
projection-based loss \cite{miyato2018cgans}, spectral normalization \cite{SNGAN},
and self-attention \cite{SAGAN}. 
The projection-based score is a
trade-off between a class-wise cosine similarity and a class independent term:
\vspace{-0.2cm}
\begin{equation}
    s(\vx, \vy) = \vy^T \mV D(\vx;\vtheta_D) + \vf'(D(\vx;\vtheta_D);
\vtheta_{D'}),
\vspace{-0.2cm}
\end{equation}
where $\vy\!\in\!\{0,1\}^C$ and $\|\vy\|_1\!=\!1$, $\mV\!\in\!\mbr{C\times d'}$
is a bilinear compatibility matrix that associates output features
$\mX_\text{out}$ from $D$ with the class label. As $\vy$ is a one-hot vector,
$\mV\!\equiv\![\vvv_1,\cdots,\vvv_C]^T$ contains linear projectors $\vvv_c$, one
per class $c\!\in\{0,\cdots,C\!-\!1\}$. Function $\vf':\mbr{d'}\!\rightarrow\!1$
is  realized by an FC layer with parameters $\vtheta_{D'}$. Scores
$s(\vx, \vy)$ are  passed to a hinge-based   
loss with two components:
{\setlength{\abovedisplayskip}{0.05cm}
\setlength{\belowdisplayskip}{0.05cm}
\begin{align}
&\mathcal{J}_\text{discr}\!=\!\mathds{E}_{(\vx,\vy)\sim p_{x\times
y}^\text{real}(\vx,\vy)}\max(0, 1\!-\!s(\vx, \vy))+\nonumber\\
&\quad\mathds{E}_{(\vz,\vy_z)\sim p_{z\times y_z}(\vz,\vy_z)}\max(0,
1\!+\!s(G(\vz, \vy_z), \vy_z)). 
\end{align}
%
Our manifold-based pipeline combines loss $\mathcal{J}_\text{prox}$ from 
\eq{prox} with $\mathcal{J}_\text{discr}$ and the original
$\mathcal{J}_\text{gen}$ from \cite{BigGAN}.}

\vspace{0.05cm}
\noindent\textbf{OmniGAN} \cite{zhou2020omni} 
uses a multi-label softmax loss (where a label vector
$\vy\!\in\!\{0,1\}^{C+2}$,  $\|\vy\|_1\!=\!2$ is a concatenation
of the one-hot class label vector and one-hot real/fake vector):
{\setlength{\abovedisplayskip}{0.1cm}
\setlength{\belowdisplayskip}{0.1cm}
\begin{align}
&\mathcal{J}_\text{discr}\!=\!\mathds{E}_{(\vx,\vy)\sim p_{x\times
y}^\text{real}(\vx,\vy)}s(\vx,\vy)+\nonumber\\
&\qquad\quad\mathds{E}_{(\vz,\vy_z)\sim p_{z\times y_z}(\vz,\vy_z)}s(G(\vz,
\vy_z),\vy_z)\quad\text{where}\\
&\qquad\quad s(\vx,\vy)\!=\!\sum\limits_{c=0}^{C+1}\log\left(1\!+\!
e^{-\sign(y_c\!-\!0.5)\phi_c(\vx)}\right). 
\end{align}
We note that $\mX_\text{out}$ represents  output features from $D$, 
$\vphi(\vx)\!=\!\vf'(D(\vx; \vtheta_{D}); \vtheta_{D'})$, where function
$\vf':\mbr{d'}\!\rightarrow\!C\!+\!2$ is  realized by an FC layer with
parameters $\vtheta_{D'}$.} 



\subsection{Feature Coding and Dictionary Learning on Data-driven Manifolds}
\label{sec:code}



Below, we formalize feature encoding and dictionary learning approaches
listed in \S \ref{sec:intro}.  
In experiments, we substitute a chosen coding step into function $\vh(\vx)$ from
\S \ref{sec:pro}.  Moreover,  $\mM\!\equiv\![\vm_1,\cdots,\vm_k]\!\in\!
\mbr{d'\times k}$ is a dictionary whose learning step is detailed at the bottom
of \S \ref{sec:code}, and $\valpha(\vx)$ 
represents the encoding/mapping on the simplex.


\vspace{0.05cm}
\noindent\textbf{Hard Assignment (HA) \cite{steinhaus_ha,csurka04_bovw}.} This encoder assigns each $\vx$ to its
nearest $\vm$ by solving the following optimisation problem:
\vspace{-0.6cm}
\begin{equation}
\begin{array}{l}
\valpha(\vx)=\argmin\limits_{\valpha'\in\left\{0,1\right\}^k}\;\rVert{
\vx-\mM\valpha'}\lVert_2^2,\\
\qquad\quad\,\text{s.t.}\;\;\|\valpha'\|_{1}=1. 
\end{array}\label{eq:ha}
\vspace{-0.2cm}
\end{equation}
If $\mM$ is formed by k-means clustering, HA becomes an equivalent of the
quantizer from FQGAN \cite{FQGAN}.

\vspace{0.05cm}
\noindent\textbf{Sparse Coding (SC) \cite{lee_sparse, yang_sparse} \& Non-negative Sparse Coding
(SC\textsubscript{+}) \cite{nnsc}.} SC encodes $\vx$  as a sparse linear combination of
atoms $\mM$ by optimising the following objective:
\vspace{-0.3cm}
\begin{equation}\label{eq:sp1}
\valpha(\vx)=\argmin\limits_{\valpha'}\;\rVert{ \vx-\mM\valpha'}\lVert_2^2
\;+\;\kappa \rVert{\valpha'}\rVert_1,
\vspace{-0.3cm}
\end{equation}
%
%
whereas SC\textsubscript{+} additionally imposes a constraint that
$\valpha'\!\geq\!0$. 
Both SC and
SC\textsubscript{+} encode $\vx$ on a subset of $\mM$  of size controlled by
the sparsity term. 


\vspace{0.05cm}
\noindent\textbf{Orthogonal Matching Pursuit (OMP) \cite{pati_omp,dmallat_omp}.} This encoder expresses
$\vx$  as a sparse linear combination of atoms $\mM$ by optimising the following
objective:
\vspace{-0.3cm}
\begin{equation}\label{eq:omp}
\begin{array}{l}
\valpha(\vx)=\argmin\limits_{\valpha'}\;\rVert{
\vx-\mM\valpha'}\lVert_2^2,\\
\qquad\quad\,\text{s.t.}\;\;\rVert{\valpha'}\rVert_0\!\leq\!\tau, 
\end{array}
\vspace{-0.3cm}
\end{equation}
where the pseudo-norm $\rVert{\valpha'}\rVert_0$ ensures the count of non-zero
coefficients of $\valpha'$ is at most $\tau$. Unlike SC and
SC\textsubscript{+}, $\rVert{\valpha'}\rVert_0$, the penalty enforces
a strict limit on the number of non-zero elements in $\valpha'$,
but the  problem itself is NP-hard.

\vspace{0.05cm}
\noindent\textbf{Approximate Locality-constrained Linear Coding (LLC) \cite{wang_llc}.} LLC expresses $\vx$  as a linear combination of $k'$ nearest neighbor atoms
of $\vx$ selected from $\mM$,  forming subspaces of size $k'$ on a
piecewise-linear manifold:  
\vspace{-0.2cm}
\begin{equation}
\begin{array}{l}
\SKIP{\valpha(\vx)=\argmin\limits_{\valpha'}\;\rVert{ \vx-\mM_{\text{NN}(\vx; k')}}\valpha'\lVert_2^2,\\
\qquad\quad\,\text{s.t.}\;\;\mathbf{1}^T\valpha'=1,} 
\valpha(\vx)=\argmin\limits_{\valpha'}\;\rVert{ \vx-\mM\valpha'}\lVert_2^2,\\
\qquad\quad\,\text{s.t.}\;\;\mathbf{1}^T\valpha'=1,
\end{array}\label{eq:llc}
\vspace{-0.2cm}
\end{equation}
and $\valpha'$ is further constrained by $\alpha'_i = 0$ unless
$\vm_i$ is one of the $k'$ closest neighbors of $\vx$.

\vspace{0.05cm}
\noindent\textbf{Soft Assignment (SA) \cite{bilmes_gmm,gemert_kernel} \& Locality-constrained Soft Assignment
(LCSA) \cite{liu_sadefense,me_SAO,me_ATN}.} SA expresses $\vx$ as the membership probability (concept known from GMM \cite{bilmes_gmm}) of $\vx$ belonging to
each $\vm_i$ in $\mM$ under equal mixing probability and equal
variance $\sigma$ of GMM. 
SA is given as:
{\setlength{\abovedisplayskip}{0.cm}
\setlength{\belowdisplayskip}{0.cm}
%
%

\vspace{-0.3cm}
\begin{align}
\label{eq:gmm4}
\valpha(\vx;\mM,\sigma) &= S_\sigma(
 \|\vx\!-\!\vm_1\|_2,\cdots, \|\vx\!-\!\vm_k\|_2),
\end{align}

\vspace{0.2cm}
\noindent
where $S_\sigma$ is the softmax function $S_\sigma: \mbr{k} \rightarrow \Delta^{k-1}$, where
$\Delta^{k-1}$ is the probability simplex and:
\vspace{0.1cm}
\begin{equation}
\label{eq:softmax}
S_\sigma(d_1, \ldots, d_k)_j = \frac{\exp(-d_j^2/2\sigma^2)}{\sum_i \exp(-d_i^2/2\sigma^2)} ~.
\vspace{0.2cm}
\end{equation}
This model yields largest values of $\alpha'_i$ for atoms $\vm_i$
in $\mM$ that are close Euclidean neighbors of $\vx$. However, 
$\alpha_i(\vx)>0$ even for $\vm_i$
that are far from $\vx$. For this reason, SA is not strictly locality-constrained.

LCSA differs from SA by setting $\alpha_i(\vx) = 0$
unless $\vm_i$ is among the $k'$ nearest-neighbor atoms
for $\vx$.  The denominator of \eq{softmax} performs normalization, that is the summation runs over the
$k'$ nearest neighbors.  Thus, LCSA maps $\mbr{d}$
onto a set of probability simplices $\Delta^{k'-1}$. 
As LCSA was the best in our 
experiments, we analyze it in \S \ref{sec:lcsa}.}

\vspace{0.05cm}
\noindent\textbf{Dictionary Learning (DL).} For the above coders, we
employ a class-agnostic dictionary learning objective which follows \eq{enc1}.
Let  some 
$\valpha(\mX)\!\equiv\![\valpha_1,
\cdots,\valpha_{N'}]$, then:
\vspace{-0.2cm}
\begin{equation}\label{eq:dl}
\begin{array}{l}
\mM=\argmin\limits_{\mM'}\;\rVert{ \mX-\mM'\valpha}\lVert_F^2,\\
\end{array}
\vspace{-0.3cm}
\end{equation}
where $\mM'$ can be constrained to contain atoms
$\rVert{\vm'_i}\rVert_2\!\leq\!1$ if codes $\valpha$ have non-restricted
$L_2$ norm \eg, for OMP.

\vspace{0.05cm}
\noindent\textbf{Inverse of $\valpha$.} To reproject $\valpha(\mX)$ from the
manifold $\tM$ into the Euclidean space, we simply compute
$\tmX\!=\!\mM\valpha(\mX)$.

\vspace{0.05cm}
\noindent\textbf{Implementation Remarks.} Coding methods, dictionary
learning, and their implementations are detailed in \S \ref{sup:code} of the supplementary material. For dictionary learning, we detach $\mX$ and $\valpha$, and
run 1 iteration of gradient descent per mini-batch \wrt each $\mM$ (no big
gain for $\geq\!2$ iterations). For SC and SC\textsubscript{+}, we detach $\mX$ and
all $\mM$, and let 5 iterations of gradient descent (no gain for $\geq\!6$
iterations). LLC has a closed-form solver \cite{wang_llc}. Our efficient 
OMP solves the system of linear equations (no matrix inversion). SA/LCSA
enjoy a fast closed-form recipe. LLC and LCSA  use the partial sort algorithm
for selecting $k'$ nearest neighbors. We detach $\mM\valpha$ to
compute the proximity loss in \eq{prox}.



\section{Theoretical Analysis of LCSA}
\label{sec:lcsa}

As LCSA is  the best encoder in our experiments, we focus on
its theoretical properties below. All proofs for
theories listed below are  in \S \ref{sup:proofs} of the supplementary material.

\newcommand{\subk}{{2^{\mM}_{k'}}}
\newcommand{\sS}{{U}}
\newcommand{\sK}{{\text{\rm NN}_{k'}}}
\newcommand{\knn}{{\mM_{\text{\rm NN}(\vx; k')}}}

In this section, we use the following notation.
Suppose a dictionary of atoms $\mM\!=\![\vm_i]$ in $\mbr{d}$ is 
given, and $\subk$ denotes the set of all subsets of size $k'$ of 
$\mM$.
We define by
$\sK: \mbr{d} \rightarrow \subk$ to be the set-valued function that takes
a point $\vx$ to its set of $k'$ nearest neighbor atoms.
A maximal subset of $\mbr{d}$ on which $\sK(\vx)$ is constant
is called a Voronoi cell.
For a given set $\sS$ in $\subk$, then, a Voronoi cell $(\sK)^{-1}(\sS)$ is a subset
of $\mbr{d}$ consisting of the all the points for which $\sS$ is the
set of $k'$ nearest neighbors.  The collection of 
all Voronoi cells constitutes a decomposition of
$\mbr{d}$ into disjoint polyhedral regions.  

In the case where the set of $k'$ nearest elements of $\mM$ is not unique,
we leave the set $\sK(x)$ undefined.
Thus, the Voronoi cells are disjoint open polyhedral regions
such that $\sK(\vx)$ is constant on each cell.  The
complement in $\mbr{d}$ of the set of Voronoi cells is a subset
of a finite set of hyperplanes in $\mbr{d}$.

\vspace{0.05cm}
\noindent\textbf{SA and LCSA encoding.} Given a dictionary $\mM \!=\! [\vm_i]$ in $\mbr{d}$ consisting of $k'$ 
elements, with $k'\!\le\!d\!+\!1$, we consider the
function $\mM \valpha(\vx)$ where $\valpha(\vx)\!=\!S_\sigma(\vx)$ is
the softmax mapping \eq{gmm4}. (The dictionary may be 
the dictionary of $k'$ nearest neighbors of some point $\vx$.)
\begin{prop}
If $\sigma\!>\!0$, the mapping $\vx \mapsto \mM \valpha(\vx)$ is a smooth 
fibration from $\mbr{d}$
{\bf onto} the interior of the simplex $\Delta$ with vertices $\vm_i$.  The fibre of this mapping is
equal to the linear subspace of $\mbr{d}$ normal to the affine space spanned by the $\vm_i$.
\end{prop}
A fibre is the set of points that map to the same point in $\Delta^{k'-1}$ under this mapping.

\vspace{-0.3cm}
\begin{prop}
\label{prop:p1}
The following properties of LCSA hold true:
\renewcommand{\labelenumi}{\arabic{enumi}.}
\vspace{-0.6cm}
\hspace{-1.0cm}
\begin{enumerate}[leftmargin=0.6cm]
\item \label{ii:ha} If $\sigma\!\rightarrow\!0$ or $k'\!=\!1$,  the $\valpha$ codes converge to
the HA solution (quantization).
\vspace*{-0.2cm}
\item \label{ii:approx_linear} $\valpha(\vx)$ is an approximately linear coding of $\vx$ in the proximity of
$\vmu(\vx)\!=\!
\frac{1}{k'} \sum_{\vm'\in\sK(\vx)} \vm' ~.
$
\vspace{-0.2cm}
\item \label{ii:rec_err}  For  $\vx$ with  
nearest neighbor atoms $\{\vm_i\}=\sK(\vx)$ with mean $\vmu(\vx)$, and nearest neighbor atom $\mathbf{n}(\vx)\!=\!\text{\rm NN}_1(\vx)$, the  reconstruction error satisfies 
\vspace{-0.3cm}
\[
%
\|\mM\valpha(\vx)\!-\!\vx\|_2 \le\\\max\Big(\|\vx\!-\!\mathbf{n}(\vx)\|_2, \|\vx\!-\!\vmu(\vx)\|_2\Big).
\vspace{-0.3cm}
\]
\item \label{ii:xxp} 
The reconstruction error varies smoothly on each Voronoi cell.
For $\vx$ and $\vx'$ in the same Voronoi cell and
$\Delta$ the simplex with vertices $\sK(\vx)$, we have

\textbullet$\;$Local Lipschitz continuity: if $\sigma> 0$, then
\[
\|\mM\valpha(\vx)\!-\!\mM\valpha(\vx')\|\!\leq\!K\|\vx\!-\!\vx'\|
\] 
where $K\!=\!{D^2/\sigma^2}$ and 
$D$ is diameter of simplex $\Delta$ (the maximum distance between 
its vertices). The Lipschitz  condition holds for $\|\cdot\|_1$ and
$\|\cdot\|_2$ norms.

\textbullet $\;$The biggest change of the reconstruction error  on a
Voronoi cell for HA ($\sigma\!=\!0)$ is less than or equal to $D$.

%


\item 
The LCSA encoding $\mM \valpha(\vx)$ is non-continuous at the boundaries of
the Voronoi regions.

%
\vspace{-0.15cm}


\end{enumerate}
\end{prop}

\begin{prop}
\label{prop:den}
Our design fulfils the principles of GAN with Denoising Auto-Encoder (DAE)
\cite{denoising_JMLR} with  loss:
\vspace{-0.3cm}
\begin{equation}
\mathcal{L}_\text{dae}=\frac{1}{N}\sum\limits_{n=1}^N\Big(\,\Big\lVert\vr(
\vx_n)-\vx_n\Big\rVert_2^2 + \sigma'^2\Big\lVert
\frac{\partial\vr(\vx)}{\partial\vx}\Big|_{\vx=\vx_n}\Big\rVert_2^2\,\Big),
\label{eq:dae}
\vspace{-0.1cm}
\end{equation}
where $\vr(\vx_n)$ is the reconstruction of $\vx_n$ akin to our $\vh(\vx_n)$ in
\eq{prox}. More importantly, $\sigma'^2$ specifies the noise variance. 
%
Specifically, we note the following:
\renewcommand{\labelenumi}{\arabic{enumi}.}
\hspace{-1.0cm}
\begin{enumerate}[leftmargin=0.6cm]
\setcounter{enumi}{5}
\item Not surprisingly, the proximity loss in \eq{prox} fulfils
somewhat similar role to $\lVert\vr(\vx_n)-\vx_n\rVert_2^2$ in
\eq{dae}.
%
\item \label{ii:dae_lcsa} LCSA implicitly fulfils the denoising role (apart from
locality-constrained non-linear coding) of $\sigma'^2\Big\lVert
\frac{\partial\vr(\vx)}{\partial\vx}\Big\rVert_F^2$, with  $\sigma$ of LCSA and
$\sigma'$ of DAE related \ie, $\sigma'^2$ is  proportional to our $\sigma^2$. 
To this end, we notice that DAE penalises the Frobenius norm of the Jacobian matrix $\Big\lVert
\frac{\partial\vr(\vx)}{\partial\vx}\Big\rVert_F$ by increasing $\sigma'^2$. We penalise the spectral norm of Jacobian matrix $\Big\lVert\frac{\partial\mM\boldsymbol{\alpha}(\vx)}{\partial\vx}\Big\rVert_2\!=\!K$ via de facto controlling the Lipschitz constant $K\!=\!{D^2}/{\sigma^2}$.
%
%
\end{enumerate}
\end{prop}

\vspace{0.05cm}
\noindent\textbf{Discussion.} The blue box below explains how properties of LCSA contribute to discriminator training, and how they let LCSA inherit the best properties of other coding methods.
%
\begin{tcolorbox}[width=1.0\linewidth, colframe=blackish, colback=beaublue, boxsep=0mm, arc=3mm, left=1mm, right=1mm, right=1mm, top=1mm, bottom=1mm]
\textbf{LCSA as a trade-off between HA, LLC, SA, DAE.} 
Prop. \ref{prop:p1} and \ref{prop:den} 
show that LCSA 
balances extremes of other coders and inherits their best properties.  Prop. \hyperref[ii:ha]{\ref*{prop:p1}.\ref*{ii:ha}} shows LCSA may act as HA (extreme way to guide $\vf$). HA is a localized coder with big 
reconstruction error. In FQGAN, HA stabilized GAN. 
%
Prop. \hyperref[ii:approx_linear]{\ref*{prop:p1}.\ref*{ii:approx_linear}} shows that LCSA may act as LLC (localized linear coder with very low reconstruction error) in the central parts of Voronoi cells (weak way to guide $\vf$). 
%
Prop. \hyperref[ii:rec_err]{\ref*{prop:p1}.\ref*{ii:rec_err}} shows that large dictionary is bad, making LCSA  a `free learner' as $\vf$ (easy to overfit).

\textbf{LCSA is locally-adaptive denoiser.} 
Prop. \hyperref[ii:xxp]{\ref*{prop:p1}.\ref*{ii:xxp}} shows each Voronoi cell   specialises in how much it denoises  $\vf$ based on the Lipschitz constant $K\!=\!{D^2}/\sigma^2$ (diameter $D$ varies in each cell due to the  dictionary). Denoising limits high frequencies of signal and its complexity. Prop. \hyperref[ii:xxp]{\ref*{prop:den}.\ref*{ii:dae_lcsa}} shows  LCSA denoises by DAE-like mechanism.  
%


 \end{tcolorbox}
 

\section{Experiments}
We evaluate our method  on CIFAR-10
\& CIFAR-100 \cite{CIFAR} \& ImageNet \cite{deng2009imagenet} (conditional GAN) and Oxford-102 Flowers \cite{nilsback_flower102} and FFHQ \cite{StyleGAN} (unconditional setting). We show our LCSA harmonizes with BigGAN \cite{BigGAN}, OmniGAN \cite{zhou2020omni}, MSG-StyleGAN
\cite{karnewar2020msg} and StyleGAN2 \cite{StyleGANV2}.

\vspace{0.05cm}
\noindent\textbf{Datasets.}
CIFAR-10 has $50K$ and $10K$ training and testing images ($32\!\times\!32$) from 10 classes, whereas CIFAR-100 
has 100 categories. 
ImageNet has $1.2M$ and $50K$ training and validation images with $1K$ classes. We center-crop and downscale its images to $64\!\times\!64$ and $128\!\times\!128$ pixels. Oxford-102
Flowers contains $8K$ images of 102 fine-grained
flower species. We  center-crop its images and
resize to $256\!\times\!256$. FFHQ dataset provides $70K$ human face images at multiple resolutions (we opt for $256\!\times\!256$). Following \cite{ADA}, we augmented the $70K$ dataset to $140K$ 
with $x$-flips.

\vspace{0.05cm}
\noindent\textbf{Evaluation Metrics.} 
 We generate $50K$ images per dataset to compute the commonly used Inception
Score \cite{Improved_GAN} and Fr\'{e}chet Inception Distance (FID) \cite{FID}. %
 Mean/standard dev. are computed over 5 runs, where both reported. %
 We report tFID, computed between $50K$ generated images and all training
images.  For CIFAR-10/CIFAR-100/ImageNet, we also compute vFID between
$10K$/$10K$/$50K$ generated images and  $10K$/$10K$/$50K$ real testing (val. on
ImageNet) images. For Oxford-102 Flowers/FFHQ, we calculate FID between $10K$/$50K$ fake
images and the entire training set.

\subsection{Network Architecture and Hyper-parameters}
We build on OmniGAN/BigGAN/StyleGAN2 for CIFAR-10. For CIFAR-100/ImageNet ($64\!\times\!64$), we experiment with OmniGAN/BigGAN. For ImageNet ($128\!\times\!128$), we build upon OmniGAN given OmniGAN consistently outperforms BigGAN. We employ MSG-StyleGAN as our baseline for Oxford-102 Flowers. For FFHQ, we build upon StyleGAN2 (see \S \ref{sup:hyper} of the supplementary material).

\subsection{Results of Image Generation}
The generated images for each dataset are given in \S \ref{sup:exim} of the supplementary material.
\vspace{0.05cm}

\noindent\textbf{CIFAR-10.} 
Table \ref{table:comparison_C10} shows results on OminGAN+LCSA,  which
outperforms the baseline OmniGAN by 0.25 and 1.43 on the IS and tFID
metrics 
($d'\!=\!256$). 
With $d'\!=\!512$, we outperform OmniGAN by $0.32$ and 3.52. 
We obtain further improvements with $d'\!=\!1024$ while baselines struggle to
converge. Comparisons of different models with $d'\!=\!1024$ are in the \S \ref{sup:cifar} of the
supplementary material. 

\begin{table}[t]
\begin{center}
\footnotesize
\newcommand{\cs}{\hspace{0.09cm}}
\begin{tabular}{|@{\cs}l@{\cs}|@{\cs}c@{\cs}|@{\cs}c@{\cs}|@{\cs}c@{\cs}|@{\cs}c@{\cs}|}
\hline
Model & $d'$  & IS $\uparrow$ & tFID$\downarrow$ & vFID $\downarrow$\\
\hline\hline
BigGAN$^\dagger$ & \multirow{7}{*}{256}& 9.14 & 7.05 & $-$ \\
FQGAN$^\dagger$ & & 9.16 & 6.16 & $-$ \\
OmniGAN$^\dagger$ & & 9.63 & 5.52 & $-$ \\
CR-GAN & & $-$ & $-$ & 11.48 \\
ContraGAN & & $-$ & $-$ & 10.32 \\
ICR-GAN & & $-$ & $-$ & 9.21 \\
OmniGAN+LCSA & & \textbf{9.88}$\pm$0.02 & \textbf{4.09}$\pm$0.10 &
\textbf{8.16}$\pm$0.07\\
\hline
BigGAN & \multirow{6}{*}{512} & 9.36 & 8.16 & 12.16 \\
FQGAN & & 9.38 & 7.65 & 11.72 \\
OmniGAN & & 9.70 & 6.88 & 10.65 \\
OmniGAN+LCSA & & 10.02$\pm$0.05 & 3.36$\pm$0.06 & 7.40$\pm$0.06\\
StyleGAN2+ADA & & 10.14$\pm$0.09 & 2.42$\pm$0.04
&6.54$\pm$0.06\\
StyleGAN2+ADA+LCSA & & \textbf{10.18}$\pm$0.06 & \textbf{2.32}$\pm$0.05 & \textbf{6.36}$\pm$0.10\\
\hline
OmniGAN+LCSA & 1024 & \textbf{10.21}$\pm$0.03 & \textbf{2.94}$\pm$0.02 &
\textbf{6.98}$\pm$0.04\\
\hline
\end{tabular}
\end{center}
\vspace{-0.65cm}
\caption{Results on CIFAR-10. We combine OmniGAN and StyleGAN2+ADA with LCSA.
$^\dagger$ are results collected from \cite{zhou2020omni}.}
\label{table:comparison_C10}
\vspace{-0.15cm}
\end{table}

\vspace{0.05cm}
\noindent\textbf{CIFAR-100.} Table \ref{table:comparison_C100} shows results on
OmniGAN+LCSA against the state of the art. For $d'\!=\!256$, our method gains $0.09$ and 1.9 on the IS and tFID metrics over the baseline OmniGAN. For $d'\!=\!512$, we outperform OmniGAN by  0.93 and 3.91 (IS and tFID). Further gains are achieved for $d'\!=\!1024$ while other methods struggle to converge. We include TAC-GAN \cite{TAC-GAN} for comparison.
\begin{table}[t]
\begin{center}
\footnotesize
\newcommand{\cs}{\hspace{0.15cm}}
\begin{tabular}{|@{\cs}l@{\cs}|@{\cs}c@{\cs}|@{\cs}c@{\cs}|@{\cs}c@{\cs}|@{\cs}c@{\cs}|}
\hline
Model & $d'$  & IS $\uparrow$ & tFID $\downarrow$ & vFID $\downarrow$\\
\hline\hline
BigGAN$^\dagger$ & \multirow{5}{*}{256} & 10.89 & 10.18 & $-$ \\
FQGAN$^\dagger$ & & 10.62 & 8.23 & $-$ \\
OmniGAN$^\dagger$ & & 13.51 & 8.14 & $-$ \\
TAC-GAN & & 9.34 & 7.22 & $-$ \\
OmniGAN+LCSA & & \textbf{13.60}$\pm$0.11 & \textbf{6.24}$\pm$0.09 &
\textbf{11.02}$\pm$0.13\\
\hline
BigGAN & \multirow{4}{*}{512} & 11.44 & 10.16 & 15.24\\
FQGAN & & 11.05 & 7.76 & 12.70 \\
OmniGAN & & 12.78 & 9.13 & 13.82 \\
OmniGAN+LCSA & & \textbf{13.71}$\pm$0.03 & \textbf{5.22}$\pm$0.10 &
\textbf{9.98}$\pm$0.08 \\
\hline
OmniGAN+LCSA & 1024 & \textbf{13.88}$\pm$0.12 & \textbf{4.97}$\pm$0.09 &
\textbf{9.72}$\pm$0.08 \\
\hline
\end{tabular}
\end{center}
\vspace{-0.65cm}
\caption{Comparison of OmniGAN+LCSA with others on CIFAR-100. $^\dagger$
are results collected from \cite{zhou2020omni}.}
\label{table:comparison_C100}
\vspace{-0.55cm}
\end{table}

\vspace{0.05cm}
\noindent\textbf{ImageNet ($\mathbf {64\times64}$).} Table
\ref{table:comparison_I64} shows that BiGAN+LCSA outperforms BigGAN by 2.97 and 3.01 (tFID \& vFID). OmniGAN+LCSA improves OmniGAN by 6.86 and 2.72 (IS \& tFID), which is the state of the art in GANs. 

\noindent\textbf{ImageNet ($\mathbf {128\times128}$).} Table
\ref{table:comparison_I128} shows OmniGAN+LCSA improves OmniGAN by 23.32 and 2.28 (IS \& tFID).

\begin{figure}[t]
    \centering
    \begin{subfigure}{0.45\linewidth}
    \begin{footnotesize}
    \newcommand{\cs}{\hspace{0.05cm}}
    \begin{tabular}[b]{|@{\cs}l@{\cs}|@{\cs}c@{\cs}|@{\cs}c@{\cs}|}
    \hline
    Model & Iters & FID $\downarrow$\\
    \hline\hline
    MSG-ProGAN$^\dagger$ & 53$K$ & 28.27\\
    MSG-StyleGAN$^\dagger$ & 50$K$ & 19.60 \\
    MSG-StyleGAN & 125$K$ & 18.59 \\
    \hline
    Ours & 125$K$ & \textbf{13.13} \\
    \hline
    \multicolumn{3}{}{}\\
    \end{tabular}
    \end{footnotesize}
    \vspace{-0.4cm}
    \caption{FID of different models.}
    \end{subfigure}
    \hspace{0.01\linewidth}
    \begin{subfigure}{0.5\linewidth}
    \includegraphics[width=\linewidth]{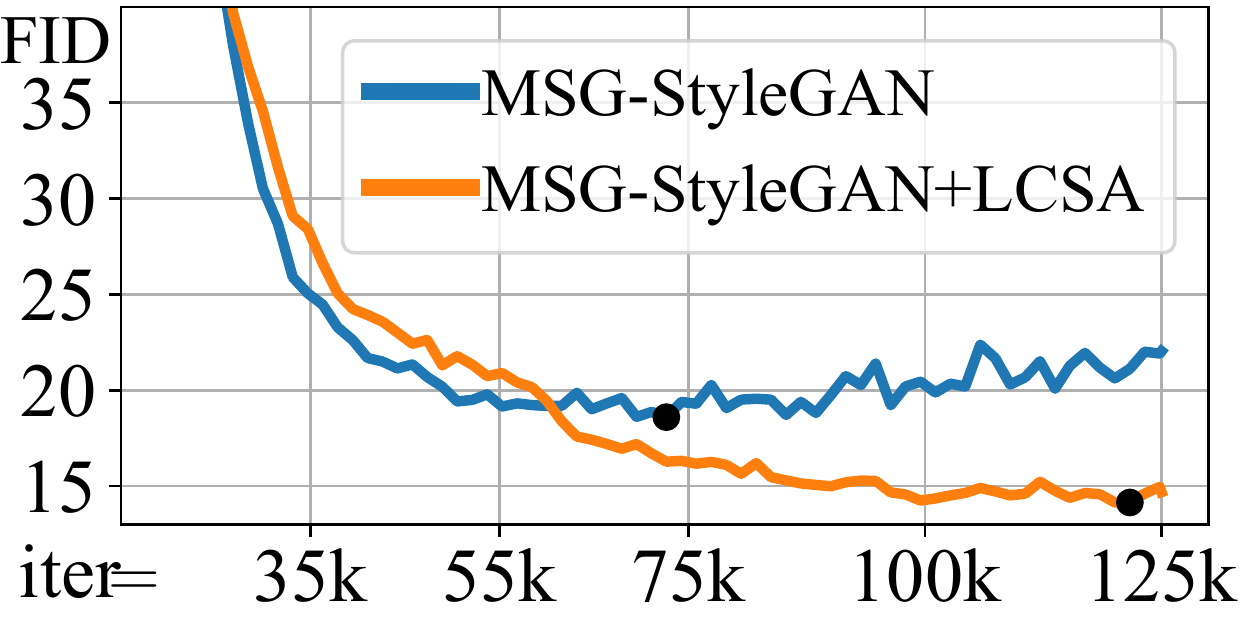}
    \caption{Training progress.}
    \end{subfigure}
    \vspace{-0.3cm}
    \caption{Results on Oxford-102 Flowers. (a) FID of different models ($^\dagger$ are results
collected from \cite{karnewar2020msg}). (b) FID \wrt the iteration number for
MSG-StyleGAN and MSG-StyleGAN+LCSA on Oxford-102 Flowers. Black dots
indicate the minimum FID.}
    \label{fig:comparison_oxford}
\end{figure}

\vspace{0.05cm}
\noindent\textbf{Oxford-102 Flowers.} Figure \ref{fig:comparison_oxford}(a)
shows that MSG-StyleGAN+LCSA improves MSG-StyleGAN by 5.46 on FID. Moreover, Figure \ref{fig:comparison_oxford}(b) shows that at around iteration number $75K$, MSG-StyleGAN starts diverging while MSG-StyleGAN+LCSA (Ours) continues to decrease FID.

\vspace{0.05cm}
\noindent\textbf{FFHQ ($\mathbf {256\times256}$).} Table \ref{table:comparison_FFHQ} shows that StyleGAN2 with LCSA improves the FID of StyleGAN2 by 1.55/0.39 on $70K/140K$ dataset. Moreover, our LCSA can also be combined with bCR \cite{ICRGAN} to improve the performance.

\vspace{0.05cm}
\noindent\textbf{Data-limited Generation on CIFAR-10/100.} We provide comprehensive experiments (10\% and 20\% data) in \S \ref{sup:limited_data} of the supplementary material. We summarize our findings as: 1) the augmentation-based  ADA \cite{ADA} and DA \cite{DiffAug} leak augmentation artifacts to the generator, while ADA+LCSA and DA+LCSA alleviate this issue (see Figure \ref{fig:C10_lim10} in the supplementary material). 2) LCSA harmonizes with ADA, DA and LeCam loss \cite{lecamgan}.
3) we achieve the state of the art on this limited data setting.

\begin{table}[t]
\vspace{0.14cm}
\begin{center}
\footnotesize
\begin{tabular}{|l|c|c|c|}
\hline
Model & IS $\uparrow$ & tFID $\downarrow$ & vFID $\downarrow$\\
\hline\hline
Inst. Sel. GAN \cite{devries2020instance}& \textbf{43.30} & 9.07 & $-$ \\
BigGAN & 34.50 & 8.96 & 8.80 \\
FQGAN & 33.14 & 8.27 & 8.15 \\
BigGAN+LCSA & 33.29 & \textbf{5.99} & \textbf{5.79}\\
\hline
OmniGAN & 70.59 & 7.09 & 7.66 \\
OmniGAN+LCSA & \textbf{77.45} & \textbf{4.26} & \textbf{4.94} \\
\hline
\end{tabular}
\end{center}
\vspace{-0.65cm}
\caption{Results on ImageNet ($64\times64$). We combine OmniGAN and BigGAN with
LCSA. We set $d'\!=\!384$.}
\label{table:comparison_I64}
\vspace{-0.1cm}
\end{table}

\begin{table}[t]
\footnotesize
\centering
\begin{tabular}{|l|c|c|c|}
\hline
Model & IS $\uparrow$ & tFID $\downarrow$ & vFID $\downarrow$\\
\hline\hline
BigGAN$^\dagger$ & 104.57 & 9.19 & 9.18 \\
OmniGAN$^\dagger$ & 190.94 & 8.30 & 8.93 \\
OmniGAN$^{\ddagger}$ & 169.13 & 7.11 & 7.30 \\
OmniGAN+LCSA & \textbf{192.45} & \textbf{4.83} & \textbf{5.24} \\
\hline
\end{tabular}
\vspace{-0.3cm}
\caption{Results on ImageNet ($128\!\times\!128$) with $d'\!=\!384$. $^\dagger$ stands for quoting from \cite{zhou2020omni}, $^{\ddagger}$ are
results reproduced by us.}
\vspace{-0.1cm}
\label{table:comparison_I128}
\end{table}

\begin{table}[t]
\vspace{-0.2cm}
\begin{center}
\footnotesize
\newcommand{\cs}{\hspace{0.23cm}}
\newcommand{\halfcs}{\hspace{0.05cm}}
\begin{tabular}{|@{\halfcs}l@{\halfcs}|@{\halfcs}c@{\halfcs}|@{\halfcs}c@{\cs}c@{\cs}c@{\cs}c@{\cs}c@{\halfcs}|}
\hline
Data & StyleGAN2 & +ADA & +LeCam & +bCR & +LCSA & +LCSA+bCR \\
\hline
70k & 5.28 & 4.30 & - & 3.79 & 3.83 & \textbf{3.47}\\
140k & 3.71 & 3.81 & 3.66 & 3.53 & 3.32 & \textbf{3.20} \\
\hline
\end{tabular}
\end{center}
\vspace{-0.65cm}
\caption{The FID $\downarrow$ results on FFHQ (256$\times$256) dataset.}
\label{table:comparison_FFHQ}
\vspace{-0.5cm}
\end{table}

\subsection{Impact of Hyperparameters}

\begin{figure*}[h]
\begin{center}
\begin{subfigure}{0.24\linewidth}
\includegraphics[width=\linewidth]{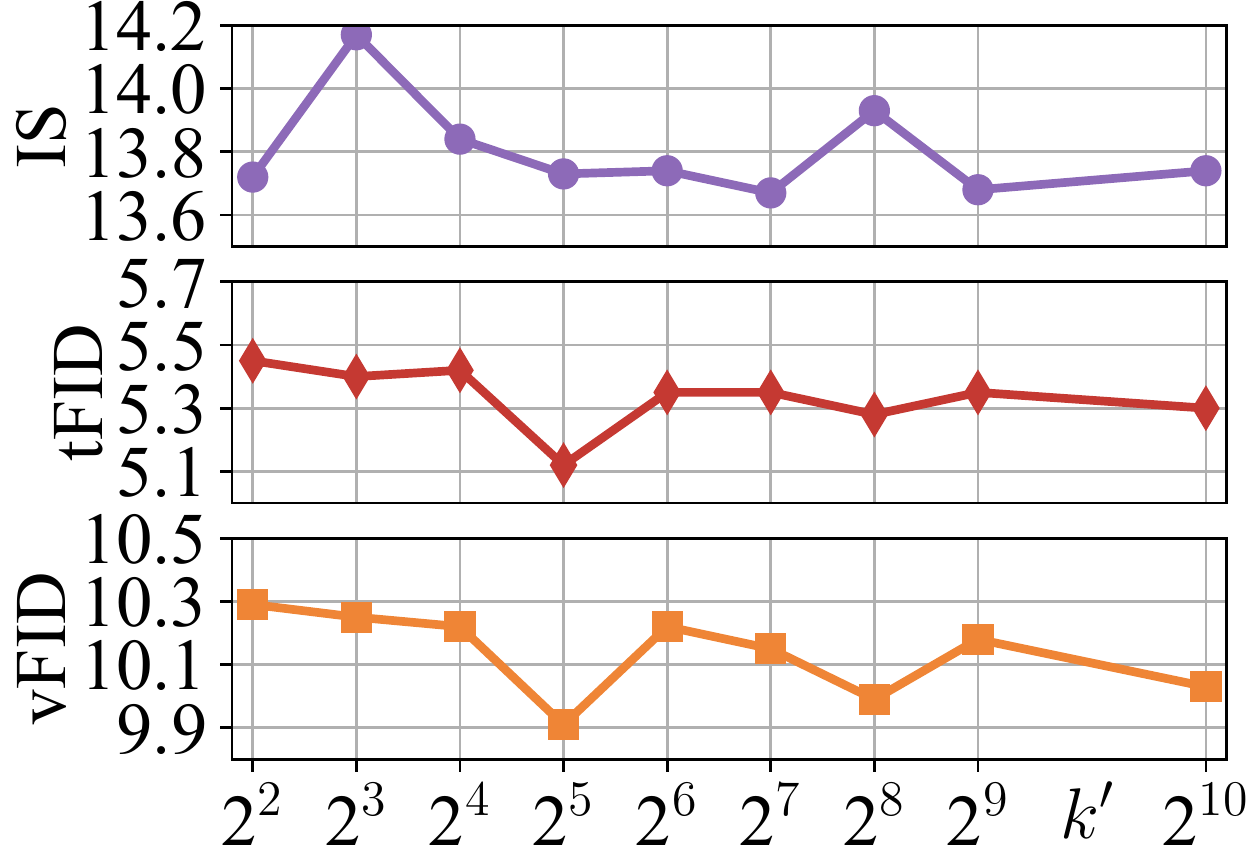}
\caption{$k'$}
\end{subfigure}
\begin{subfigure}{0.24\linewidth}
\includegraphics[width=\linewidth]{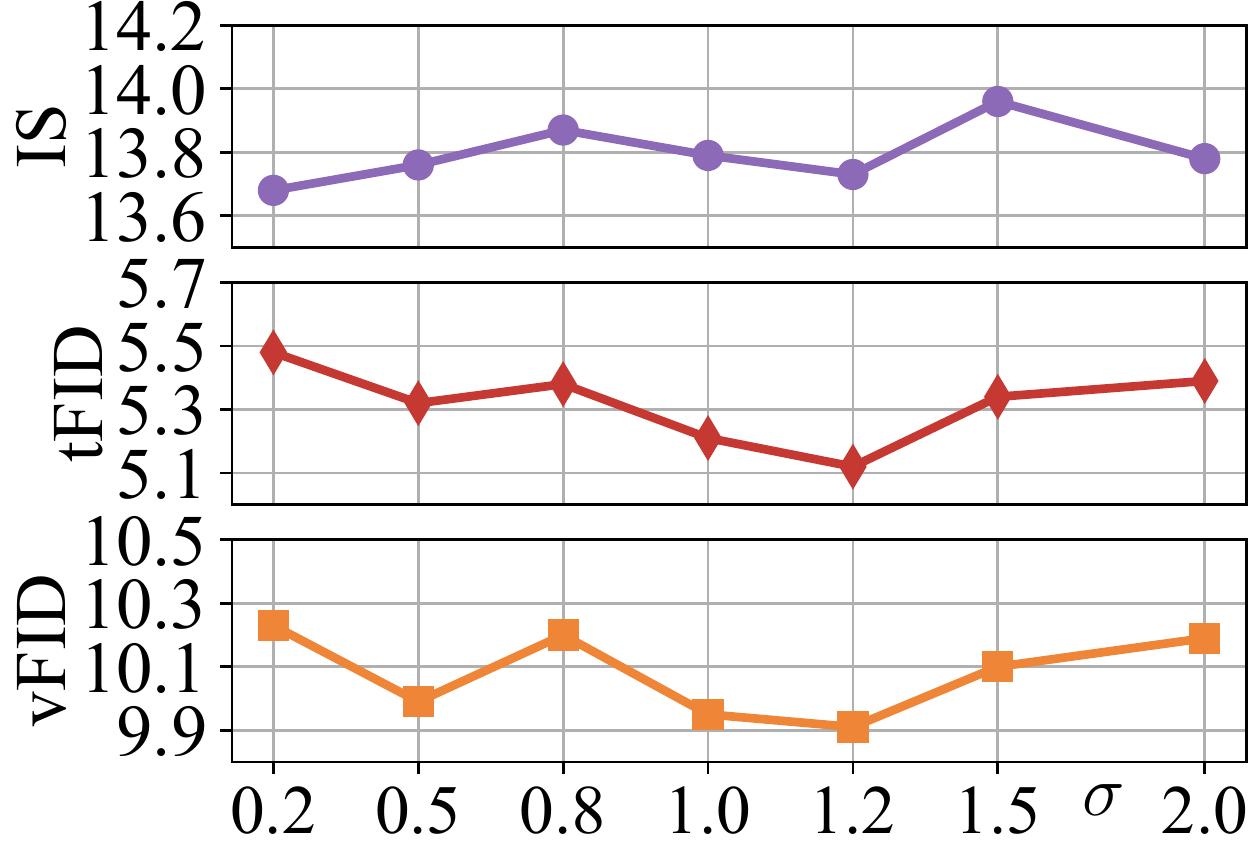}
\caption{$\sigma$}
\end{subfigure}
\begin{subfigure}{0.24\linewidth}
\includegraphics[width=\linewidth]{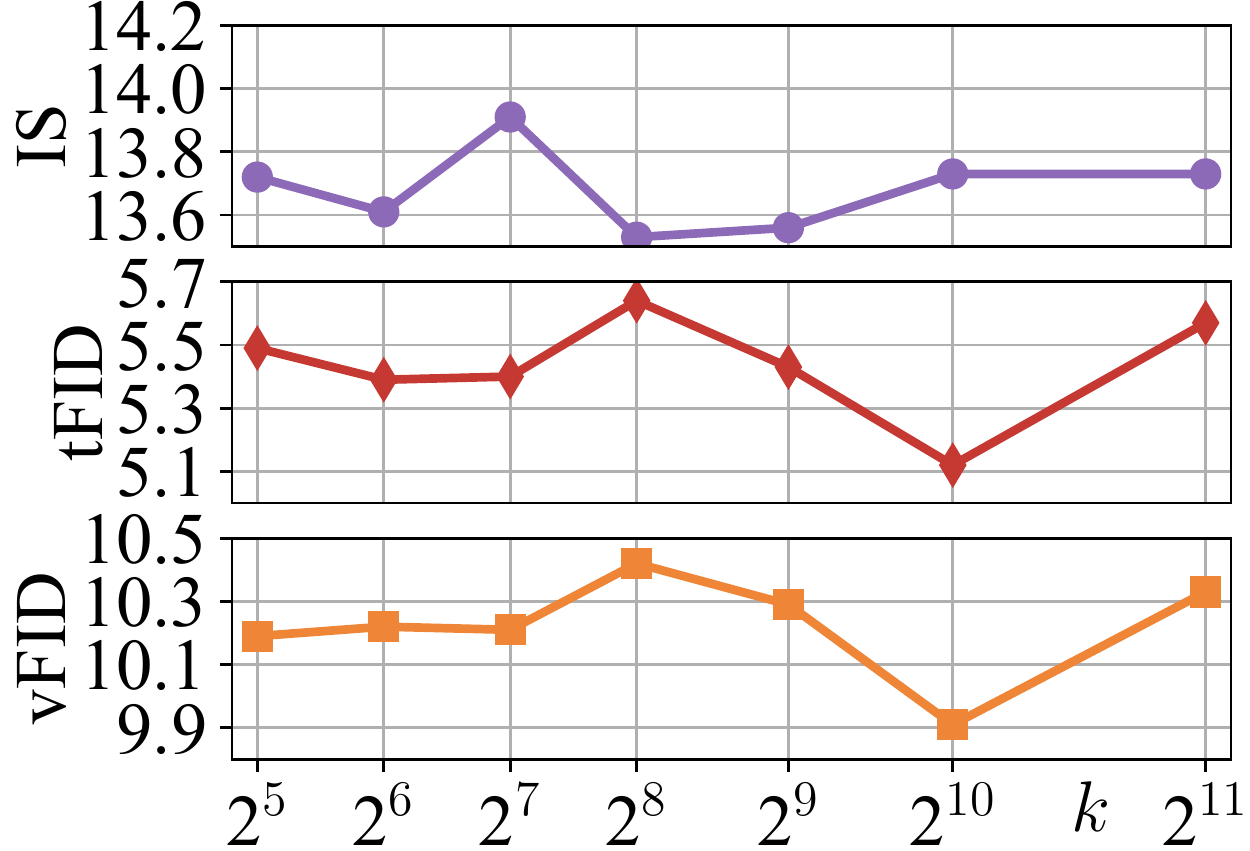}
\caption{$k$}
\end{subfigure}
\begin{subfigure}{0.24\linewidth}
\includegraphics[width=\linewidth]{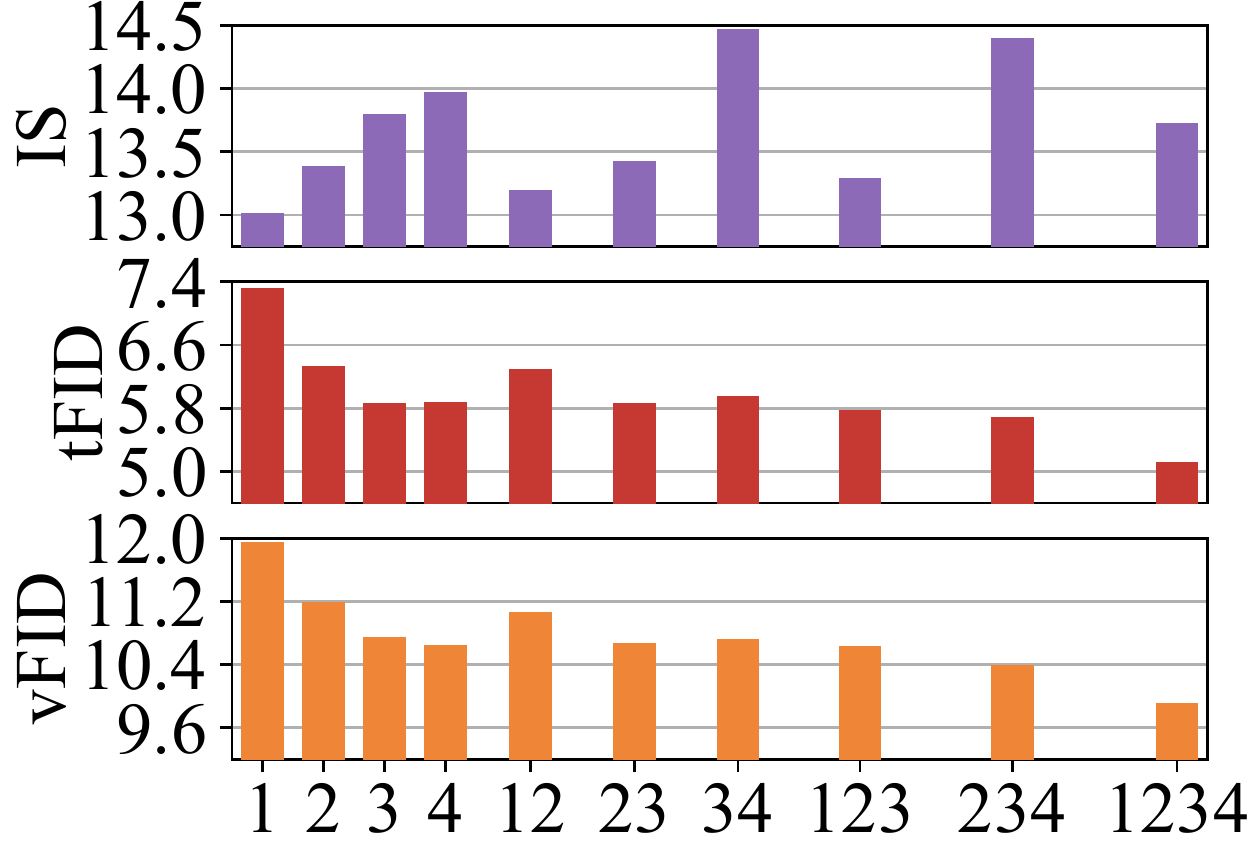}
\caption{$l\!\in\!\{1,\cdots,L\}$}
\end{subfigure}
\vspace{-0.3cm}
\caption{Ablation studies on CIFAR-100 \wrt $k'$ (nearest neighbors), $\sigma$
(controls the Lipschitz constant of LCSA), $k$ (dictionary size), and blocks $l$ which
use LCSA. We indicate metrics such as the IS $\uparrow$, tFID $\downarrow$
and vFID $\downarrow$.}
\label{different_hyper_parameters}
\end{center}
\vspace{-0.2cm}
\end{figure*}

Below, we conduct  ablations on CIFAR-100 for $d'\!=\!512$. 

\vspace{0.05cm}
\noindent\textbf{Nearest neighbors $\boldsymbol{k'}$, $\boldsymbol{\sigma}$ and
$\boldsymbol{k}$.}
Figure \ref{different_hyper_parameters}(a) 
shows results on OmniGAN+LCSA \wrt $k'$ on CIFAR-100, which verifies that the
locality-constrained manifold can be constructed with
$8\!\leq\!k'\!\leq\!32\!\ll\!k$ with  a sufficiently small
reconstruction error 
below the worst case
(Prop. \hyperref[ii:rec_err]{\ref*{prop:p1}.\ref*{ii:rec_err}}). For ImageNet,  $k'\!=\!8$ also led to best results. Figure \ref{different_hyper_parameters}(b) verifies that $1\!\leq\!\sigma\!\leq\!1.5$
provides the best trade-off in terms of smoothness (quantization \vs  linearization)
(Prop. \hyperref[ii:ha]{\ref*{prop:p1}.\ref*{ii:ha}}, \hyperref[ii:approx_linear]{\ref*{prop:p1}.\ref*{ii:approx_linear}} and \hyperref[ii:xxp]{\ref*{prop:p1}.\ref*{ii:xxp}}).
Figure \ref{different_hyper_parameters}(c) verifies the benefit of dictionary overcompleteness 
$d'\!\ll\!k$ as $k\!=\!1024$.

\vspace{0.05cm}
\noindent\textbf{Blocks $\boldsymbol{l\!\in\!\{1,\cdots,L\}}$ and LCSA. } Figure
\ref{different_hyper_parameters}(d) shows the results for various combinations
of injection of LCSA into blocks of discriminator. It appears that injecting
LCSA into all blocks yields the lowest tFID/vFID, which verifies that
constructing multiple manifolds at coarse-to-fine semantic level  controls
the complexity of the discriminator.

\begin{figure}[t]
\vspace{-0.6cm}
\begin{center}
\begin{subfigure}{0.493\linewidth}
\includegraphics[width=\linewidth]{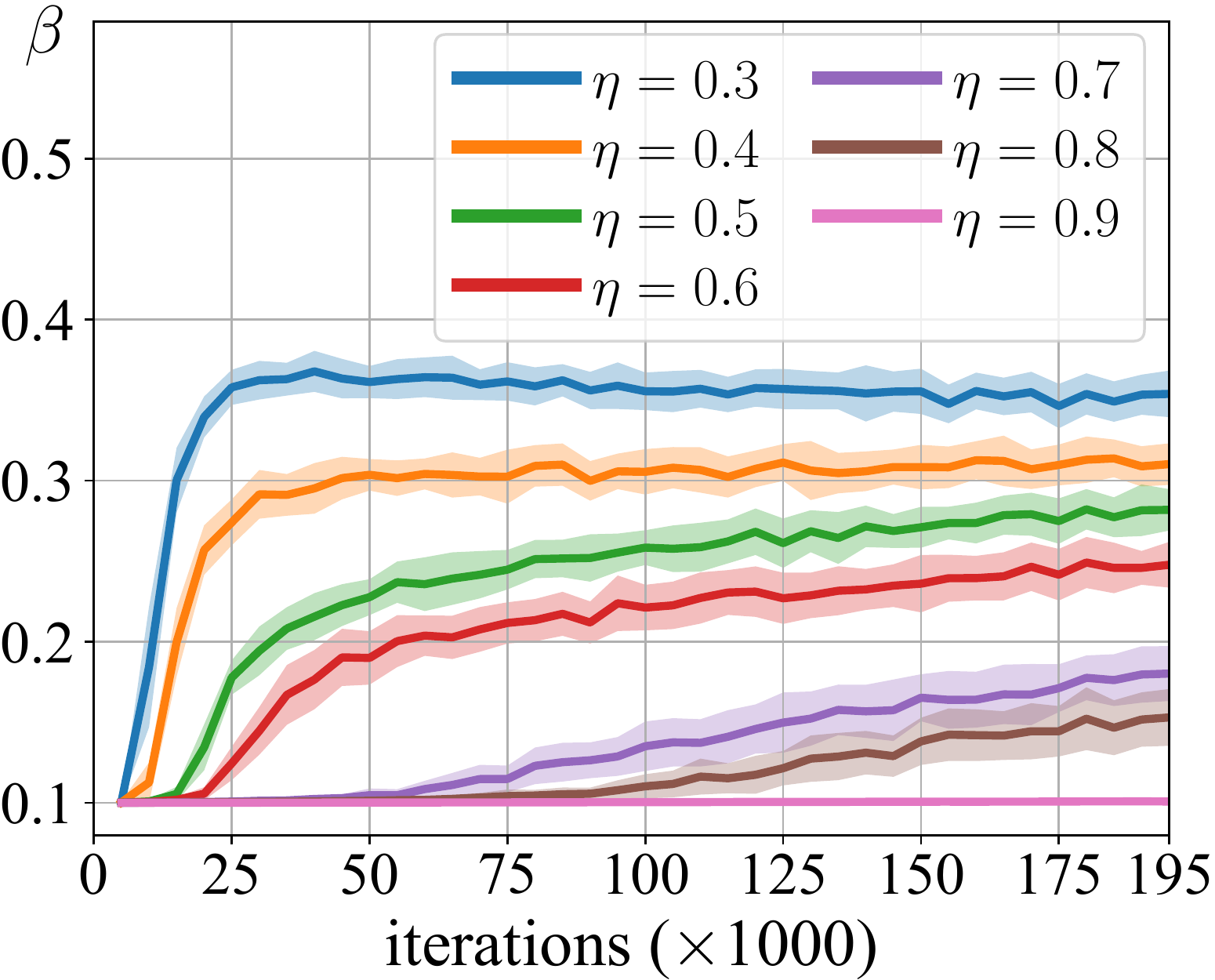}
\caption{$\beta$ \wrt iteration.}
\end{subfigure}
\begin{subfigure}{0.493\linewidth}
\includegraphics[width=\linewidth]{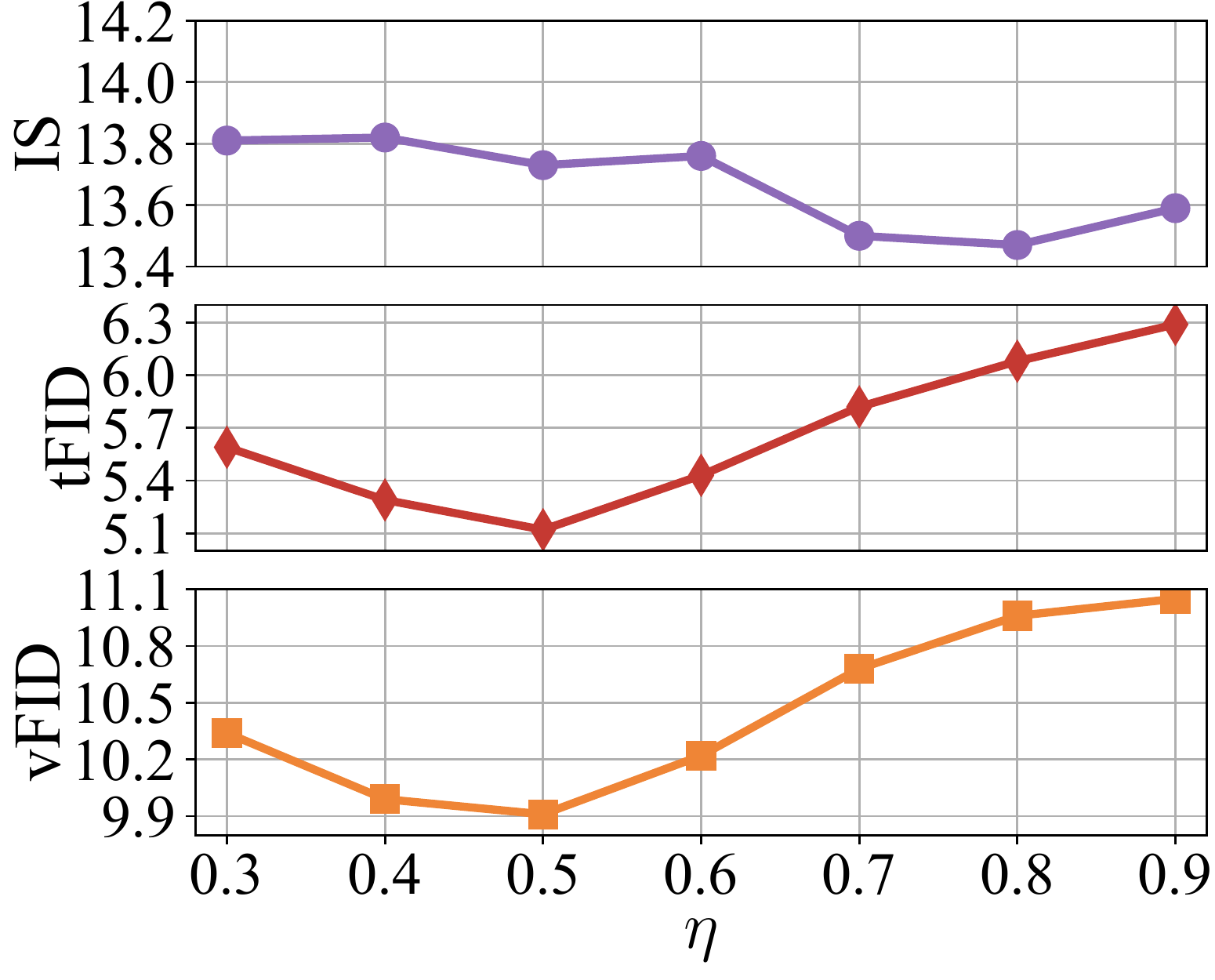}
\caption{$\eta$ \wrt iteration.}
\end{subfigure}
\vspace{-0.3cm}
\caption{Evolution of metaparameter $\beta$ and the impact of $\eta$ which
controls the behavior of the detector of overfitting.}
\label{figure:eta_beta}
\end{center}
\end{figure}

\vspace{0.05cm}
\noindent\textbf{Metaparameter $\boldsymbol{\beta}$ and impact of $\boldsymbol{\eta}$.} 
 $\beta$ in \eq{beta} controls the mixing of conv.
features obtained by $\vf$ and their view $\vh$ recovered from the manifold, as in
\eq{enc2}. 
%
%
Figure ~\ref{figure:eta_beta}(a) shows that in early iterations, blocks of
discriminator learn conv. features. Around $25K$ iterations, the model starts
oscillating between prevention of overfitting and refining conv. features, as
indicated by the green curve that continues to gradually grow for $\eta\!=\!0.5$
(\cf flat curve for $\eta\!=\!0.3$). Figure \ref{figure:eta_beta}(b) also shows
that $\eta\!=\!0.5$ is a universally good threshold. \SKIP{(we set $\eta\!=\!0.5$ on
all datasets).} 
%
%

\subsection{Analysis and Ablation studies}
We analyse our method on CIFAR-10/100 ($d'\!=\!512$).

\begin{figure}[t]
\vspace{-0.6cm}
\begin{center}
\includegraphics[width=1\linewidth]{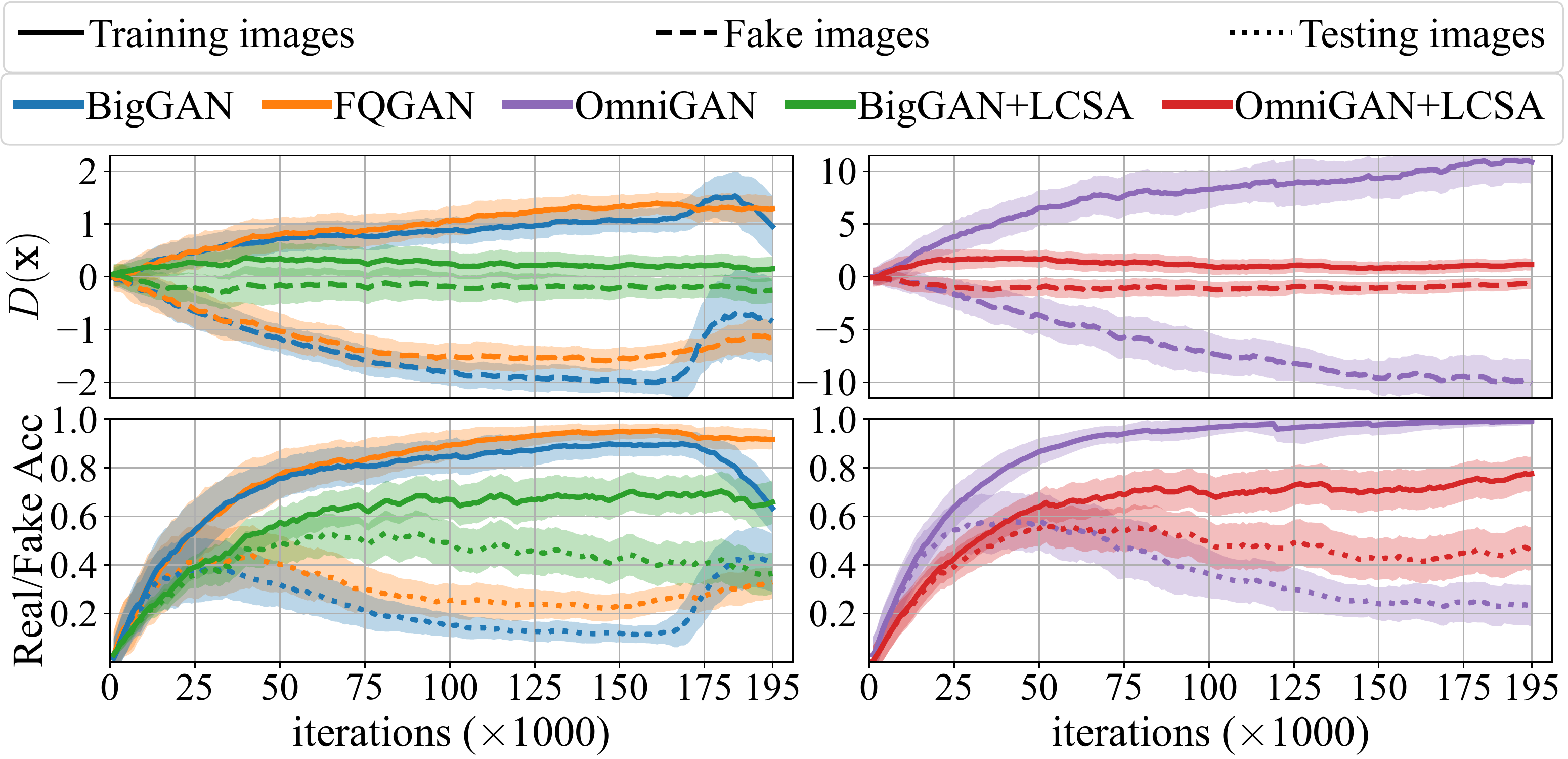}
\vspace{-0.6cm}
\caption{Discriminator predictions on CIFAR-10. We plot the output of discriminator on training images and generated fake images. We test Acc (real/fake accuracy predicted by the discriminator) on the training images and testing images. We use different colors to represent different models. Solid/dash/dot lines indicate training/fake/testing images.}
\label{figure:Dout_acc_C10}
\end{center}
\vspace{-0.5cm}
\end{figure}

\begin{figure}[t]
\vspace{-0.1cm}
\begin{center}
\includegraphics[width=1\linewidth]{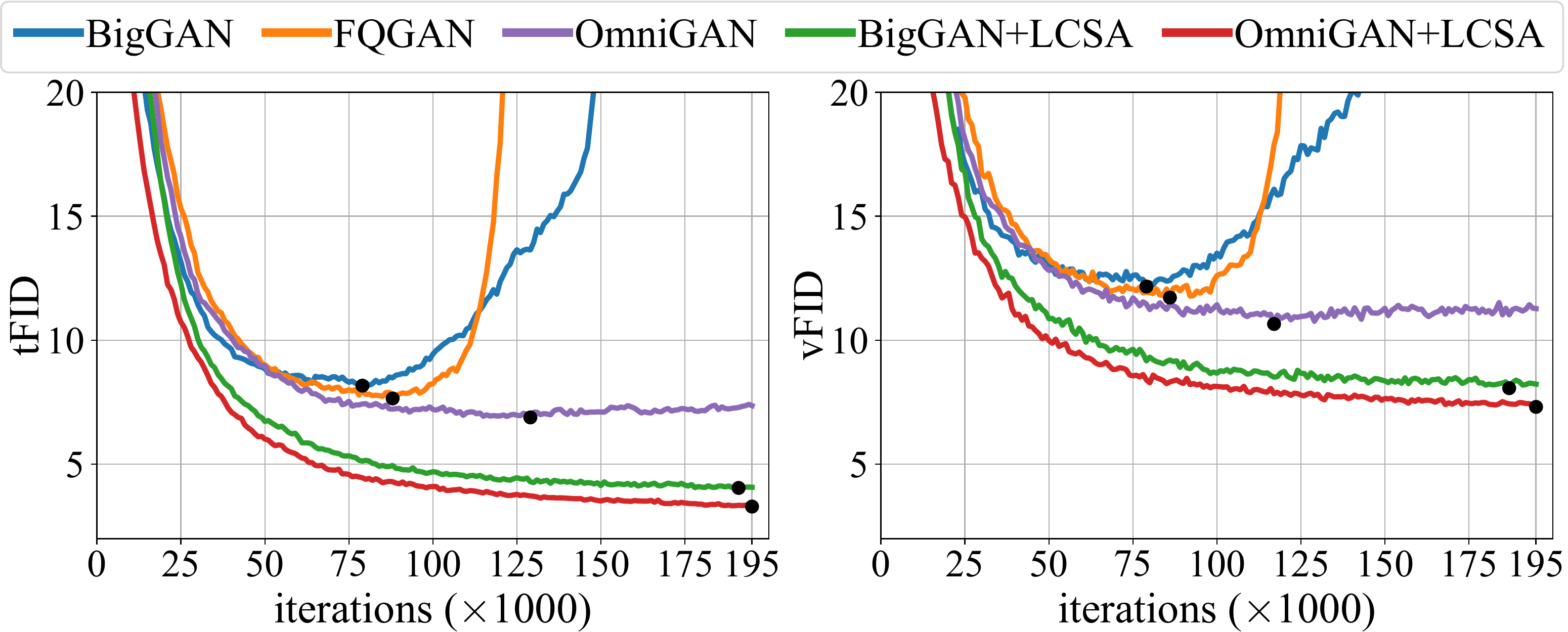}
\vspace{-0.6cm}
\caption{Evolution of tFID and vFID for different models on CIFAR-10.
Black dots indicate the minimum FID. 
}
\label{figure:fid_C10}
\end{center}
\vspace{-0.9cm}
\end{figure}

\noindent\textbf{Preventing Discriminator Overfitting.} Figure \ref{figure:Dout_acc_C10} verifies that the discriminator of standard methods yields high real/fake accuracy on training images but low accuracy on testing images. Thus, they overfit to the training set (see the large discrepancy between real and fake images). Thus, standard methods diverge early (see FID) but our method continues learning, as shown in Figure \ref{figure:fid_C10} and Figure \ref{fig:comparison_oxford}(b).


%

\noindent\textbf{Impact of Different Components.} In Table \ref{table:coding_methods}, we ablate (i) dictionary learning (Eq. (\ref{eq:dl})), (ii) the adaptive mixing input Eq. (\ref{eq:enc2}) and (iii) proximity loss Eq. (\ref{eq:prox}). We conduct experiments with settings: (1) ACM: removing our mainfold learner and the proximity loss, and changing Eq. (\ref{eq:enc2}) to be $\mX_{l+1}\!=\!(1\!-\!\beta)\tmX_l$ to adaptively control the magnitude of $\tmX_l$; (2) LCSA($\gamma\!=\!0$): removing the proximity loss but keeping the adaptive mixing input; (3) LCSA($\beta\!=\!0$): removing the adaptive mixing input but keeping the adaptive proximity loss; (4) LCSA(EMA): replacing our dictionary learning with the exponential moving average; (5) LCSA(fixed ($\beta$, $\gamma$)): fixed meta-controller ($\beta$, $\gamma$) (different combination of ($\beta$ and $\gamma$) is given in \S \ref{sup:fixx} of the supplementary material).  Table \ref{table:coding_methods} shows that (i) LCSA($\gamma\!=\!0$) is better than ACM and OmniGAN which verifies the benefit of adaptive mixing input.  (ii) LCSA($\beta\!=\!0$) is better than 
OmniGAN which verifies the importance of controlling the complexity of learning. (iii) EMA performs worse than our dictionary learning. (iv) As fixed ($\beta$, $\gamma$) scores lower thus the adaptive meta-controller is useful. In \S \ref{sup:residual} (supplementary material),  LCSA  yields larger errors at foregrounds and smaller at backgrounds, thus intertwining step (mixing input) helps refine/denoise features before input to next layer.

\vspace{0.05cm}
\noindent\textbf{Different Encoders.}
Table \ref{table:coding_methods} compares  coding methods, each applied to all $l\!=\!1,\cdots,4$ blocks of discriminator.
LCSA is the best performer followed by SA and LLC or SC, which validates that
using locality-constrained soft assignment whose continuity is controlled via the Lipschitz constant (Prop. \hyperref[ii:xxp]{\ref*{prop:p1}.\ref*{ii:xxp}}) inverse-proportional to $\sigma^2$ is more robust than
locally-linear coding, due to the quantization \vs linear reconstruction trade-off and the role of $\sigma^2$ in denoising (Prop.  \ref{prop:den}). 
%
We also tried replacing the LCSA coder with Denoising
Auto-Encoder (DAE) \cite{denoising_JMLR} (the setup is in \S \ref{sup:dae} of the supplementary material). LCSA achieves better tFID and  vFID than locality-constrained linear coding,  subspace learning, and DAE. 
Figure \ref{fig:var1} of the supplementary material shows the variance computed over LCSA codes for indivisual images. While visually mundane regions have low variance (\ie, single code of $\boldsymbol{\alpha}$ encodes it), the visually diverse regions have variance that is somewhat higher for LCSA than other coders, preventing overfitting where it matters. 
%

\begin{table}[t]
\vspace{-0.6cm}
\begin{center}
\footnotesize
\newcommand{\ics}{\hspace{0.1cm}} 
\newcommand{\ocs}{\hspace{0.1cm}} 
\begin{tabular}{|@{\ocs}l@{\ocs}|c@{\ics}c@{\ics}c|c@{\ics}c@{\ics}c@{\ocs}|}
\hline
\multirow{2}{*}{Method}& \multicolumn{3}{|c|}{CIFAR-10} & \multicolumn{3}{|c|}{CIFAR-100}\\
\cline{2-7}
& IS $\uparrow$ & tFID $\downarrow$ &
vFID $\downarrow$ & IS $\uparrow$ & tFID $\downarrow$ &
vFID $\downarrow$ \\
\hline\hline
OmniGAN & 9.70 & 6.88 & 10.65 & 12.78 & 9.13 & 13.82\\
\hline
+ACM & 9.79 & 6.03 & 9.99 & 12.82 & 10.61 & 15.51 \\
+LCSA($\gamma=0$) & 9.80 & 5.15 & 8.35 & 13.42 & 8.12 & 12.78 \\
+LCSA($\beta=0$) & 9.73 & 4.77 & 8.77 & 13.64 & 5.54 & 10.37 \\
+LCSA(EMA) & 9.91 & 3.83 & 7.89 & 13.71 & 5.63 & 10.40 \\
+LCSA(fixed($\beta$, $\gamma$)) & 10.01 & 4.06 & 8.03 & 13.64 & 5.46 & 10.29 \\
\hline
+LCSA & 10.09 & \textbf{3.29} & \textbf{7.31} & 13.73 & \textbf{5.12} & \textbf{9.91}\\
+HA & 9.88 & 4.52 & 8.49 & 13.68 & 5.52 & 10.33\\
+SC$_+$ & 10.09 & 3.53 & 7.65 & 13.82 & 5.49 & 10.31\\
+SC & 10.14 & 3.48 & 7.46 & 13.68 & 5.40 & 10.20\\
+OMP & 10.02 & 3.83 & 7.83 & 13.58 & 5.52 & 10.40\\
+LLC & 10.04 & 3.77 & 7.75 & 13.61 & 5.35 & 10.14\\
+SA & 10.00 & 3.46 & 7.45 & 13.76 & 5.31 & 10.19\\
\hline
+DAE & 10.08 & 3.89 & 7.91 & 13.65 &5.41 & 10.22 \\
\hline
\end{tabular}
\end{center}
\vspace{-0.5cm}
\caption{Results for ablation studies on CIFAR-10 \& CIFAR-100.}
\label{table:coding_methods}
\vspace{-0.5cm}
\end{table}

\vspace{-0.1cm}
\section{Conclusions}
We have applied data-manifold learning (LCSA) in the coarse-to-fine manner to conv. features of
discriminator to prevent its overfitting by adaptively balancing the trade-off between denoising on the mainfold and refining the mainfold (intertwining  and dictionary learning steps), controlling the complexity of learning (the proximity loss). Locality-constrained soft assignment is controlled via the Lipschitz constant, resulting in a  trade-off between quantization and linear coding, yielding  state-of-the-art results.



\noindent\textbf{Acknowledgements.} $\!$We thank CSIRO’s Machine Learning and Artificial Intelligence Future Science Platform (MLAI FSP) {and  China Scholarship Council (CSC)} for the support.

{\small
\bibliographystyle{ieee_fullname}
\bibliography{arxiv-ready}
}


\newpage
\appendix
\title{Manifold Learning Benefits GANs (Supplementary Material)\vspace{-0.3cm}}

\author{%
  Yao Ni\textsuperscript{\textasteriskcentered}$^{\!, \dagger}$, \quad Piotr Koniusz\textsuperscript{\textasteriskcentered}$^{, \S,\dagger}$, \quad Richard Hartley$^{\dagger,\vardiamond}$, \quad Richard Nock$^{\vardiamond\!,\clubsuit,\dagger}$\\\vspace{0.3cm}
  $^{\dagger}$The Australian National University \quad 
   $^\S$Data61/CSIRO  \quad $^{\vardiamond}$Google Research \\
	\vspace{-0.5cm}
  firstname.lastname@anu.edu.au \\
	\vspace{-0.3cm}
}

\maketitle

Below is the contents of our supplementary material:

\begin{itemize}
\item \S \ref{sup:notations} explains our mathematical notation.

\item \S \ref{sup:cifar} provides further evaluations on CIFAR-10 and CIFAR-100
(32$\times$32) datasets \cite{CIFAR} given the large feature size of
$d'\!=\!1024$. We believe that it is valuable to demonstrate that our method can
benefit from such a large $d'$ while other methods cannot benefit much when
applying $d'\!=\!1024$ instead of the usual $d'\!=\!512$.

\item \S \ref{sup:limited_data} provides results for the limited data setting on CIFAR-10 and CIFAR-100.

\item \S \ref{sup:imagenet32} provides results on the ImageNet (32$\times$32)
dataset \cite{deng2009imagenet} in order to demonstrate that our LCSA-based
pipeline benefits experiments on GAN for lower resolution images, which are
favoured over high resolution images for instance when enriching training
datasets of few-shot learning approaches. Note that results on ImageNet (64$\times$64) and ImageNet (128$\times$128) are given in our main submission.




\item \S \ref{sup:note_fqgan} explains why FQGAN suffers from discriminator
overfitting.


\item Section \ref{sup:dae} investigates the use of Denoising Auto-Encoder (DAE)
\cite{denoising_JMLR} in place of LCSA, and concludes that  LCSA is more
beneficial. 

\item \S \ref{sup:fixx} investigates the use of fixed $\beta$ and $\gamma$ rather
than metaoptimized $\beta$ and $\gamma$. The conclusion is that with mixing and
proximity operators controlled by these parameters, we can reduce overfititng in
the discriminator while training the manifold dictionary.

\item \S \ref{sup:residual} illustrates the residual error between $\tmX$ and
$h(\tmX)$. The results support our belief that conv. encoder refines LCSA on
difficult foreground objects while generating coarse-to-fine cov. features close
to the manifold.

\item \S \ref{sup:hyper} details hyperparameters used in our experiments. We
point out that the majority of parameters remain unchanged throughout many
experiments.

\item \S \ref{sup:code} provides details of the feature encoding methods and the
dictionary learning step.

\item \S \ref{sup:exim} illustrates examples of images generated with or without the
LCSA coder.

\item \S \ref{sup:proofs} offers  proofs for claims of \S
\ref{sec:lcsa}. 

\end{itemize}

\section{Notations}
\label{sup:notations}
Capitalized boldface symbols are matrices (\eg, $\mX$), lowercase boldface
symbols  are vectors (\eg, $\vx$), and regular fonts denote scalars \eg, $n$,
$N$, $l$, $L$, $X_{i,j}$ is a $(i,j)$-th coefficient  of $\mX$, $x_{i}$ is an
$i$-th coefficient of $\vx$. We define a vector of all-ones as
$\mathbf{1}\!=\!\left[1,\cdots,1\right]^T$ and concatenation of $\alpha_i$ as
$[\alpha_i]_{i=1,\cdots,I}$.

\section{Experiments on CIFAR-10/100 ($\boldsymbol{d'\!=\!1024}$)}
\label{sup:cifar}
\begin{figure}[h]
\vspace{-0.1cm}
\centering
\includegraphics[width=1\linewidth]{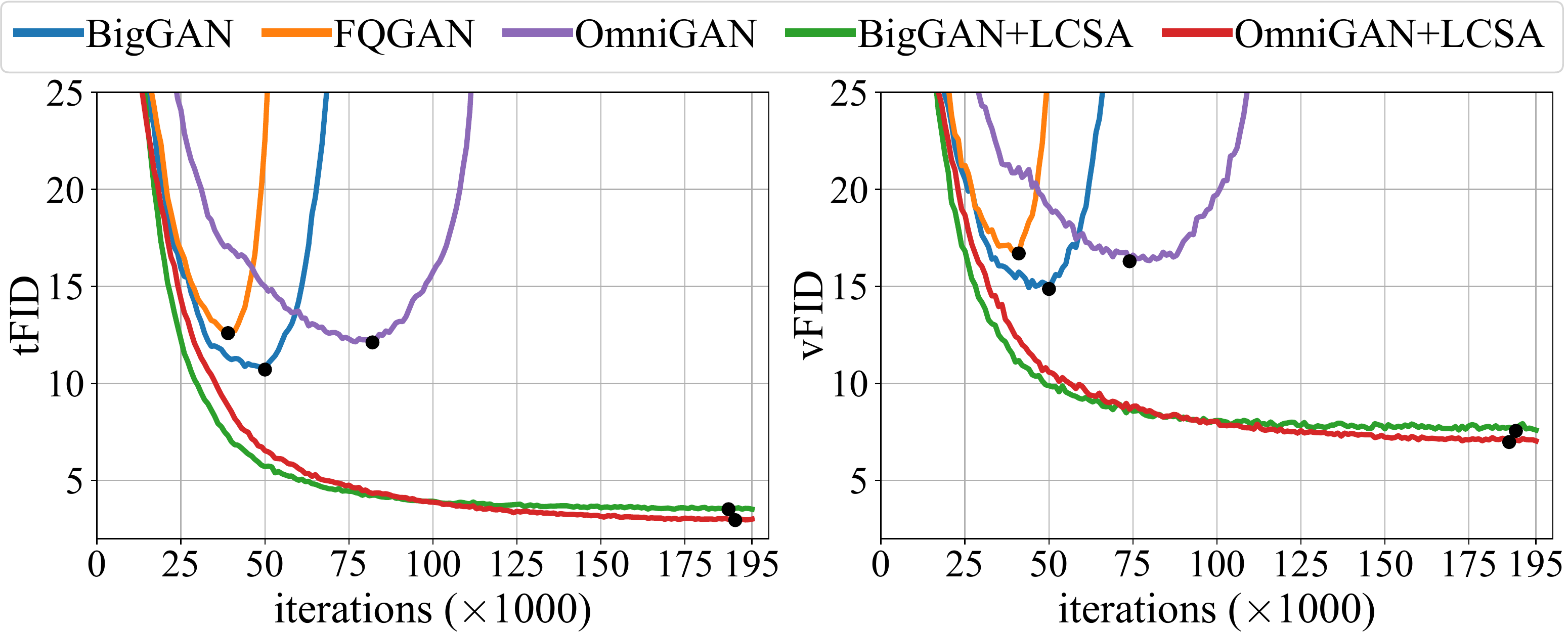}
\vspace{-0.6cm}
\caption{Evolution of tFID and vFID for different models on CIFAR-10
with $d'\!=\!1024$. Black dots indicate the minimum FID.
}
\label{figure:fid_1024_C10}
\end{figure}

\begin{figure}[h]
\vspace{-0.1cm}
\centering
\includegraphics[width=1\linewidth]{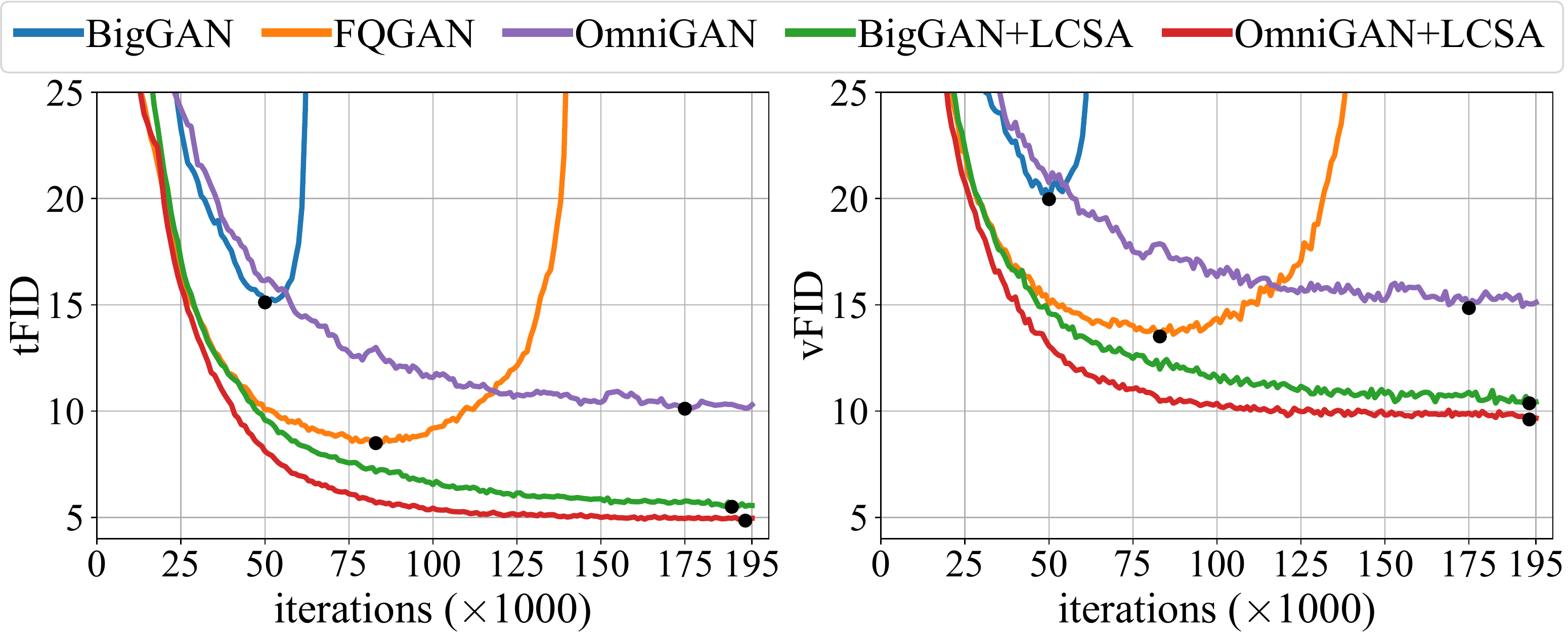}
\vspace{-0.6cm}
\caption{Evolution of tFID and vFID for different models on CIFAR-100
with $d'\!\!=\!\!1024$. Black dots indicate the minimum FID.
}
\label{figure:fid_1024_C100}
\end{figure}

Below we demonstrate that utilizing LCSA in BigGAN and OmniGAN further enables
them to utilize large feature sizes ($d'\!=\!1024$), improving results further.
Original BigGAN and OmniGAN struggle to attain good results for such a large
feature size. Tables \ref{table:CIFAR_1024} verifies our
pipelines on CIFAR-10 and CIFAR-100. Figures \ref{figure:fid_1024_C10} and
\ref{figure:fid_1024_C100} show that our BigGAN+LCSA and OmniGAN+LCSA continue
to learn while other models struggle to converge. 
\begin{table}[h]
\vspace{-0.3cm}
\centering
\footnotesize
\newcommand{\cs}{\hspace{0.1cm}}
\begin{tabular}{|@{\cs}l@{\cs}|c@{\cs}|@{\cs}c@{\cs}|@{\cs}c@{\cs}|c@{\cs}@{\cs}|c@{\cs}|@{\cs}c@{\cs}|}
\hline
\multirow{2}{*}{Model} & \multicolumn{3}{c|}{CIFAR-10} & \multicolumn{3}{c|}{CIFAR-100}\\
\cline{2-7}
& IS $\uparrow$ & tFID $\downarrow$ & vFID $\downarrow$ & IS $\uparrow$ & tFID $\downarrow$ & vFID $\downarrow$\\
\hline\hline
BigGAN & 9.64 & 10.71 & 14.86& 11.54 & 15.11 & 19.97\\
FQGAN & 9.44 & 12.59 & 16.69& \textbf{11.74} & 8.49 & 13.51\\
BigGAN+LCSA & \textbf{9.81} & \textbf{3.51} & \textbf{7.55} & 11.60 & \textbf{5.49} & \textbf{10.37} \\
\hline
OmniGAN & 9.92 & 12.11 & 16.29 & 12.42 & 10.11 & 14.85\\
OmniGAN+LCSA & \textbf{10.21} & \textbf{2.94} & \textbf{6.98} & \textbf{13.88} & \textbf{4.97} & \textbf{9.72}\\
\hline
\end{tabular}
\vspace{-0.3cm}
\caption{Results for different models on CIFAR-10 and CIFAR-100 with $d'\!=\!1024$.}
\label{table:CIFAR_1024}

\end{table}

\begin{table*}[ht!]
\begin{center}
\footnotesize
\newcommand{\cs}{\hspace{0.05cm}}
\setlength{\tabcolsep}{0.12cm}
\begin{tabular}{|@{\cs}l@{\cs}|c@{\cs}|c@{\cs}|c@{\cs}|c@{\cs}|@{\cs}c@{\cs}|@{\cs}c@{\cs}|}
\hline
 \multirow{3}{*}{Model} & \multicolumn{3}{c|}{CIFAR-10} & \multicolumn{3}{c|}{CIFAR-100}\\
\cline{2-7}
& 100\% & 20\% & 10\%& 100\% & 20\% & 10\%\\
\cline{2-7}
& IS$\uparrow$ / tFID$\downarrow$ / vFID$\downarrow$ & IS$\uparrow$
/ tFID$\downarrow$ / vFID$\downarrow$& IS$\uparrow$
/ tFID$\downarrow$ / vFID$\downarrow$& IS$\uparrow$
/ tFID$\downarrow$ / vFID$\downarrow$& IS$\uparrow$
/ tFID$\downarrow$ / vFID$\downarrow$& IS$\uparrow$
/ tFID$\downarrow$ / vFID$\downarrow$\\
\hline\hline
BigGAN($d'\!=\!256$) & 9.26 / \ \ 5.46 / \ \ 9.38 & 8.70 / 16.25 / 20.37  & 8.20 / 31.42 / 35.58 & 10.99 / 7.89 / 12.74 & \ \ 9.94 / 25.96 / 30.89 & \ \ 7.54 / 50.89   / 55.15 \\
\cline{1-7}
+DA & 9.39 / \ \ 4.47 / \ \ 8.58 & 8.95 / \ \ 9.38 / 13.26 &  8.65 / 18.35  / 22.04 & 10.91 / 7.30 / 11.99 & \ \ 9.73 / 16.32 / 20.88 & \ \ 9.33 / 27.01 / 31.32 \\
+LeCam & 9.48 / \ \ 4.25 / \ \ 8.27 & 8.96 / 11.32 / 15.25  & 8.50 / 26.30 / 30.55  & \textbf{11.44} / 6.53 / 11.23 & 10.00 / 20.82 / 25.77 & \ \ 8.10 / 39.33 / 44.23 \\
+LCSA &  \textbf{9.51} / \ \ \textbf{4.12} / \ \ \textbf{8.20} & \textbf{8.99} / \ \ \textbf{8.72}  / \textbf{12.73}  & \textbf{8.74} / \textbf{12.36} / \textbf{16.46} & 11.02 / \textbf{6.38} / \textbf{11.13} & \textbf{10.38} / \textbf{13.22} / \textbf{18.06}  &  \textbf{10.08} / \textbf{21.13} / \textbf{25.87} \\
\cline{1-7}

+LeCam+DA & 9.47 / \ \ 4.29 / \ \ 8.29 & 9.05 / \ \ 7.53 / 11.46 &  8.84 / 12.15 / 15.92   & 11.15 / 6.55 / 11.34 & 10.42 / 13.01 / 17.65 & \ \ 9.92 / 21.73 / 26.19 \\
+LCSA+DA & \textbf{9.50} / \ \ \textbf{3.75} / \ \ \textbf{7.83} & \textbf{9.08} / \ \ 7.29 / 11.31 & 8.86 / 10.86 / 14.78  & \textbf{11.21} / \textbf{5.76} / \textbf{10.52} & 10.55 / 12.03 / 16.92 &  \textbf{10.68} / 19.38 / 24.21 \\
+LCSA+LeCam+DA & 9.47 / \ \ 3.80 / \ \ 7.89 & 9.04 / \ \ \textbf{6.95} / \textbf{10.95} & \textbf{8.96} / \textbf{10.05} / \textbf{13.88}  & 11.17 / 5.85 / 10.64 & \textbf{10.67} / \textbf{10.16} / \textbf{15.00}  &  10.28 / \textbf{18.24} / \textbf{23.12}\\

\hline\hline


 OmniGAN($d'\!=\!1024$) & 9.98 / \ \ 6.89 / 10.76 & 8.62 / 37.86 / 42.18 & 6.59 / 54.04 / 58.71 & 12.61 / 8.43 / 13.21  & 10.11 / 40.57 / 44.86 & \ \ 6.87 / 63.41  / 67.46 \\
\cline{1-7}
+DA  & \!\!\!10.10 / \ \ 4.26 / \ \ 8.15  & 9.47 / 13.56 / 17.34 & 8.96 / 19.59 / 23.60 & 12.96 / 7.48 / 12.11 & 11.42 / 17.72 / 22.49 & 10.21 / 32.61 / 36.86 \\
+ADA & \!\!\!\textbf{10.27} / \ \ 5.03 / \ \ 9.15 & 9.44 / 27.20 / 31.05 & 7.72 / 41.82 / 45.36 &  13.47 / 6.13 / 10.88 & 12.18 / 13.66 / 18.34 & \ \ 8.81 / 46.74 / 51.03 \\
+LCSA & \!\!\!10.20 / \ \ \textbf{2.58} / \ \  \textbf{6.65} & \textbf{9.97} / \ \ \textbf{5.15} / \ \ \textbf{9.01} & \textbf{9.74} / \ \ \textbf{7.66} / \textbf{11.47} & \textbf{13.74} / \textbf{4.94} / \ \ \textbf{9.67} & \textbf{12.86} / \textbf{11.15} / \textbf{15.90} & \textbf{11.85} / \textbf{13.82} / \textbf{18.56}\\
\cline{1-7}
+LCSA+DA & \!\!\!10.22 / \ \ 2.52 / \ \ 6.60  & 9.93 / \ \ 5.01 / \ \ 8.89 & 9.75 / \ \ 7.46 / 11.35 & 13.79 / 4.58 / \ \ 9.46 & 12.92 / 11.06 / 15.73 & 11.88 / 13.68 / 18.44 \\
+LCSA+ADA & \!\!\!\textbf{10.38} / \ \  \textbf{2.35} / \ \ \textbf{6.38} & \!\!\!\textbf{10.12} / \ \ \textbf{4.48} / \ \ \textbf{8.41} & \!\!\!\textbf{10.04} / \ \ \textbf{6.45} / \textbf{10.39} &  \textbf{13.80} / \textbf{4.07} / \ \ \textbf{8.90} & \textbf{13.78} / \ \ \textbf{7.45} / \textbf{12.11} & \textbf{12.67} / \textbf{10.18} / \textbf{14.87} \\ \hline


\end{tabular}
\end{center}
\vspace{-0.5cm}
\caption{Comparison of our LCSA with DA, ADA and LeCam on CIFAR-10 and
CIFAR-100 given different percentage of training data. }
\label{tab:comparison_aug}
\vspace{-0.3cm}
\end{table*}

\begin{figure}[h]
\vspace{-0.08cm}
\centering
\begin{subfigure}{0.485\linewidth}
\includegraphics[width=\linewidth]{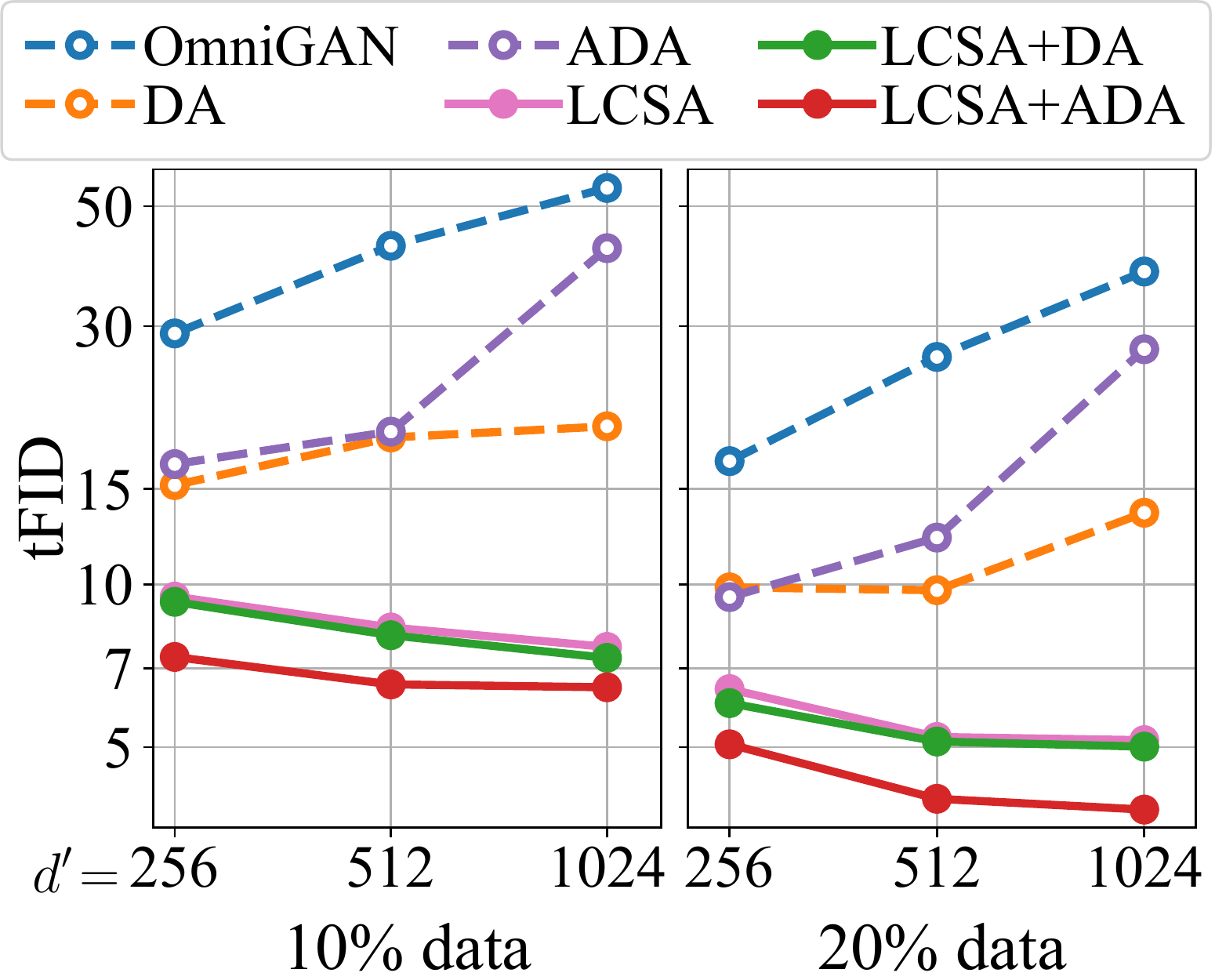}
\caption{CIFAR-10}
\end{subfigure}\hspace{0.015\linewidth}
\begin{subfigure}{0.485\linewidth}
\includegraphics[width=\linewidth]{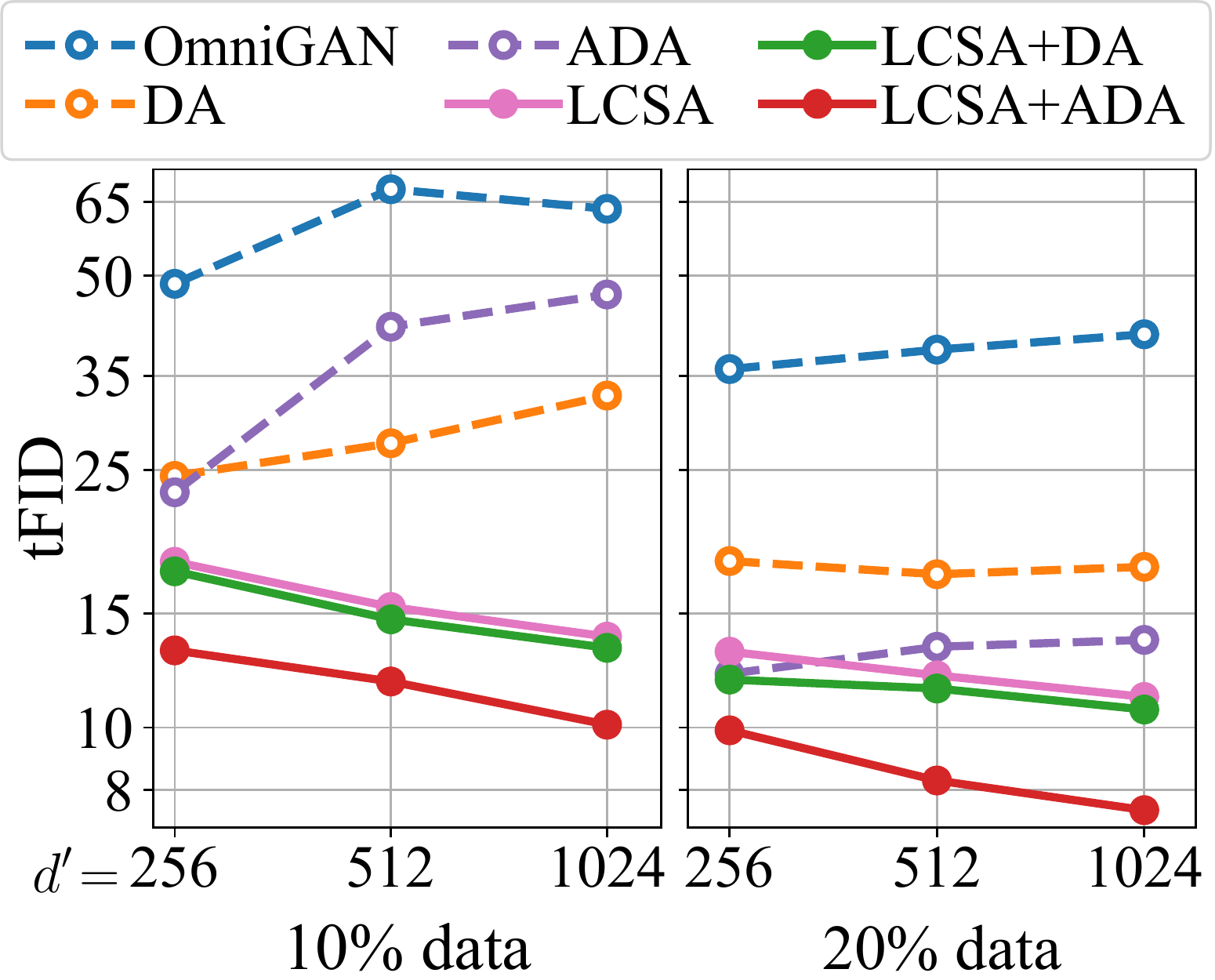}
\caption{CIFAR-100}
\end{subfigure}
\vspace{-0.3cm}
\caption{tFID \wrt different $d'$ on CIFAR-10 and CIFAR-100 under the limited data setting.}
\label{fig:limited_feature_size}
\vspace{-0.4cm}
\end{figure}

\begin{figure}[h]
\vspace{-0.1cm}
\centering
\begin{subfigure}{0.485\linewidth}
\includegraphics[width=\linewidth]{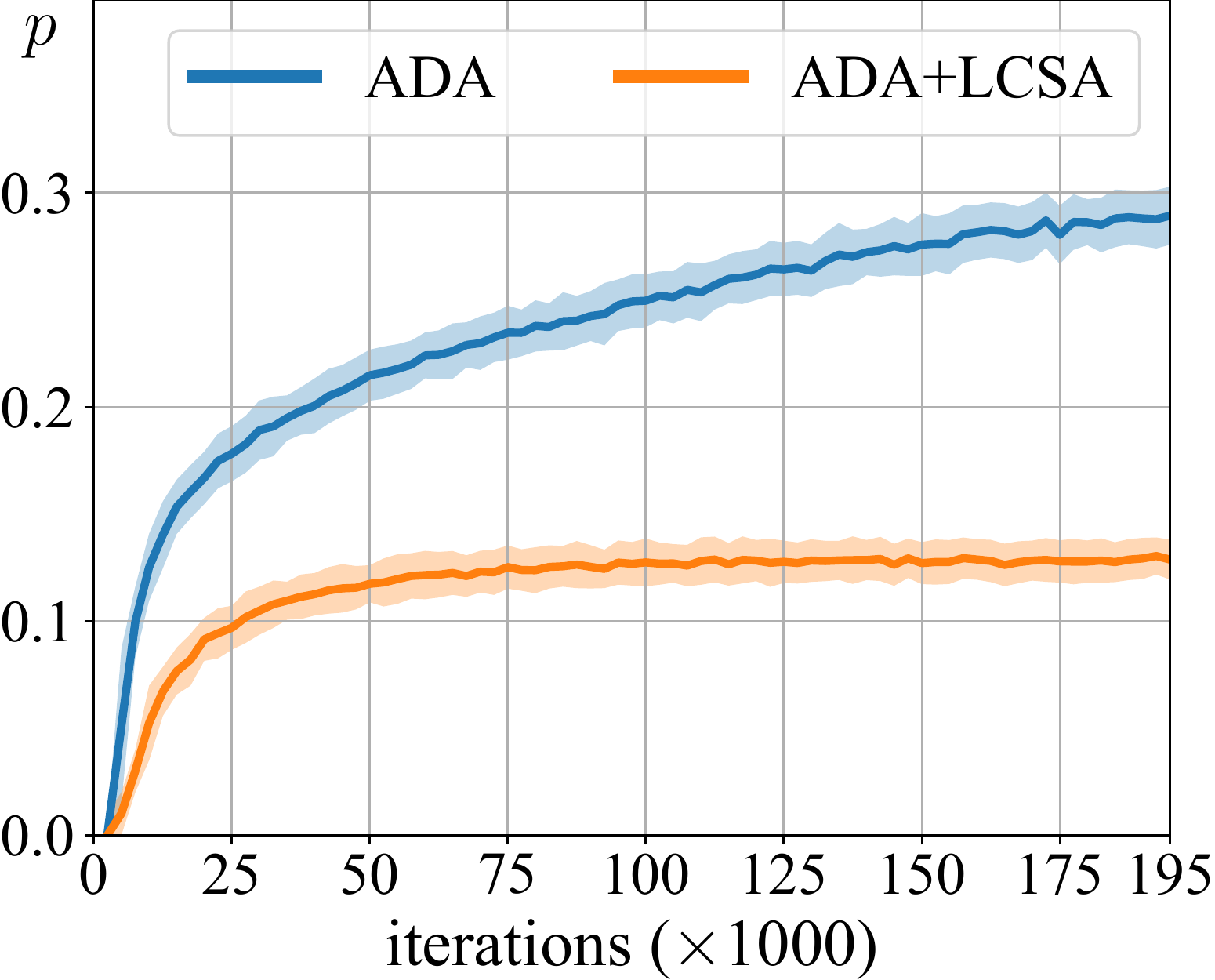}
\caption{10\% CIFAR-10}
\end{subfigure}\hspace{0.015\linewidth}
\begin{subfigure}{0.485\linewidth}
\includegraphics[width=\linewidth]{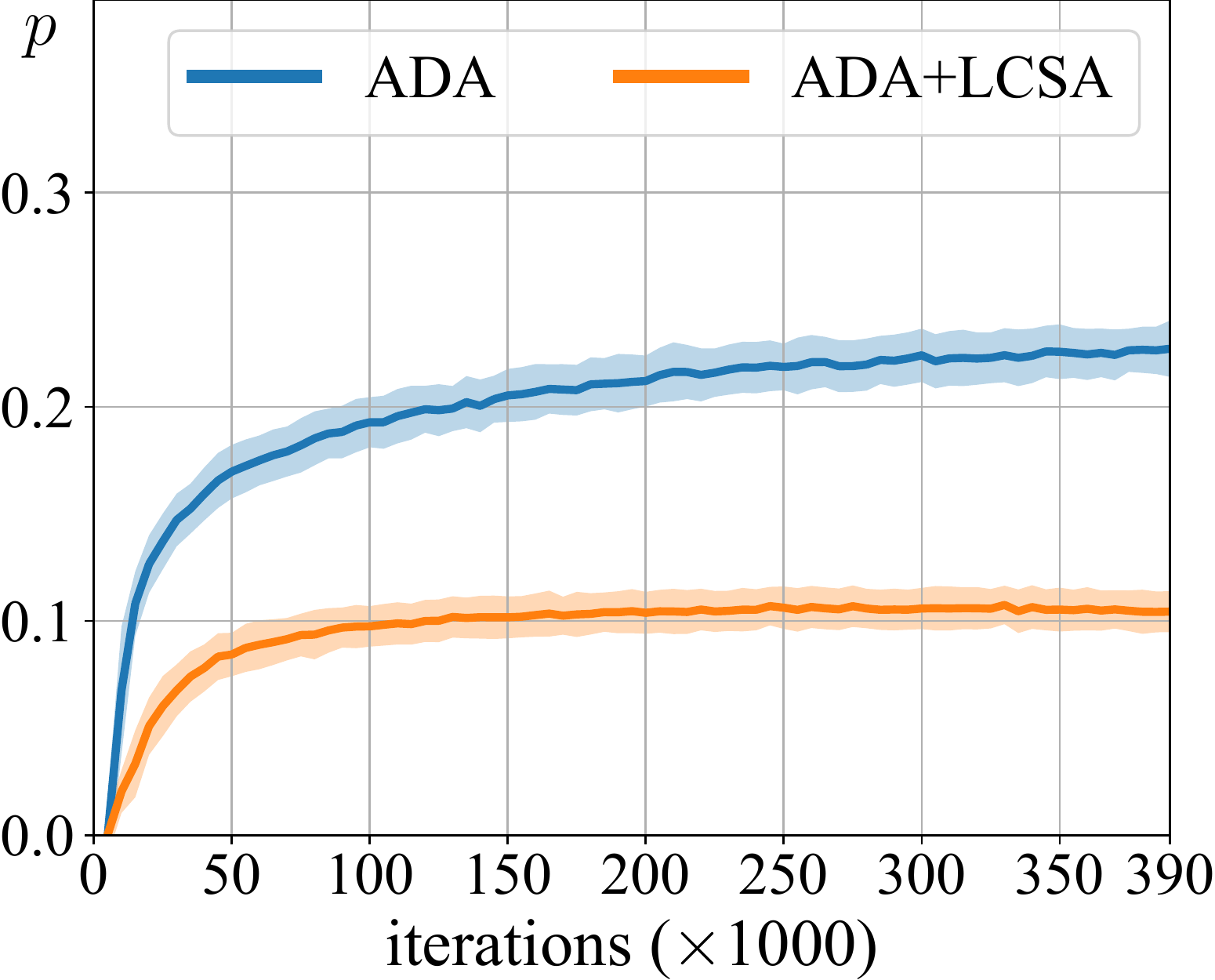}
\caption{20\% CIFAR-10}
\end{subfigure}
\vspace{-0.4cm}
\caption{Augmentation strength ($p$) of ADA \vs ADA+LCSA during training on 10\%/20\% CIFAR-10 data (OmniGAN $d'\!\!=\!\!256$).}
\label{fig:aug_p}
\vspace{-0.4cm}
\end{figure}

\section{Data-limited Generation on CIFAR-10/100}
\label{sup:limited_data}
Both DA \cite{DiffAug} and ADA \cite{ADA} limit discriminator overfitting by augmenting the real and generated images that are passed to the discriminator. However, in case of DA and ADA, augmentation artifacts leak into the generated images as shown in Figure \ref{fig:C10_lim10}. The LeCamGAN \cite{lecamgan} restricts discriminator overfitting by reducing the real/fake loss discrepancy at the output of the discriminator. However, it is not enough to limit overfitting by just operating at the  output of the discriminator (otherwise a well-designed loss would deal with all overfitting issues which is not the case in the literature). Therefore, we prevent the discriminator overfitting by modifying its blocks by encoding both the features of real and fake images into a common manifold and recovering their view from it, which has the ability to control the complexity of our feature space.

Below, we compare our LCSA with DA \cite{DiffAug}, ADA \cite{ADA} on OmniGAN (and/or BigGAN) to variants without LCSA. We conduct experiments based on BigGAN to compare our LCSA with LeCamGAN \cite{lecamgan} due to the issue of combining LeCam loss \cite{lecamgan} with the multi-output loss of OmniGAN. Following \cite{DiffAug}, we augmented the real images with $x$-flip, and trained the OmniGAN and BigGAN for $1K$ epochs on the full data and $5K$ epochs on 10\%/20\% data setting. 
The DiffAug used  translation and cutout augmentations. The ADA used 18 augmentations, including rotations and scaling. Details can be founded in \cite{ADA}.

Table \ref{tab:comparison_aug} shows that our LCSA outperforms DA, ADA and LeCamGAN. LCSA   can be always combined with them to improve their performance. Specifically, we achieve state-of-the-art results on CIFAR-10 (100\%, 20\% and 10\%) and CIFAR-100 (20\% and 10\%) based on OmniGAN ($d'\!=\!1024$).  In addition, in Figure \ref{fig:limited_feature_size}, OmniGAN, DA and ADA become worse when increasing the $d'$, while combining them with our LCSA  improves the tFID given the larger model size, on which the discriminator can memorize the training data easier, demonstrating that our LCSA has stronger performance while preventing overfitting better than ADA and DA alone. Moreover, we find that the DA and ADA leak the augmentation cues to the generated images, as shown in Figure \ref{fig:C10_lim10}, indicating that their discriminators overfit to the augmentation cues instead of learning meaningful discrimination boundary. ADA strives to control the strength of augmentations by a controller which observes the discriminator output and adjusts the desired augmentation strength. Thus, in Figure \ref{fig:aug_p} we  plot the augmentation strength of ADA and LCSA+ADA. We notice that \textbf{\color{red}LCSA+ADA is able to limit the augmentation strength of ADA, thus reducing the leakage of augmentation artifacts.} This means LCSA can deal with the discriminator overfitting well. Concluding, LCSA enjoys better performance than ADA, DA, and LeCam for data-limited generation. LCSA gains further improvements with larger model size $d'$, and LCSA lowers the risk of leaking augmentation artifacts. Thus, LCSA appears to have the superior capability of preventing discriminator overfitting than other strategies.

\section{Experiments on ImageNet (32$\times$32)}
\label{sup:imagenet32}
We preprocess images by center-cropping and downscaling them to $32\times32$
pixels. We train the network for 100 epochs. We set mini-batch size to 64,
$\beta_0\!=\!0.$, $\gamma_0\!=\!0.1$, $\eta\!=\!0.5$, $\sigma\!=\!1.2$, whereas
the dictionary learning rate is set to 1e-3. We equip each residual block of the
discriminator with our LCSA module. We set $\Delta_{\gamma}\!=\!1.2$ for 
$d'\!=\!512$. For OmniGAN and OmniGAN+LCSA, we set weight decay of discriminator
to be 1e-4 given $d'\!=\!512$. Table \ref{tab:my_label} presents our results in
the above setting.
\begin{table}[!h]
\vspace{-0.2cm}
    \footnotesize
    \centering
    \begin{tabular}{|l|c|c|c|}
    \hline
    Model & IS $\uparrow$ & tFID $\downarrow$ & vFID $\downarrow$ \\
    \hline\hline
    BigGAN & 13.25 & 5.44 & 5.60 \\
    BigGAN+LCSA & \textbf{13.61} & \textbf{4.82} & \textbf{5.01}\\
    \hline
    OmniGAN & 29.55 & 3.74 & 4.56\\
    OmniGAN+LCSA & \textbf{30.53} & \textbf{3.53} & \textbf{4.37} \\
    \hline
    \end{tabular}
    \vspace{-0.2cm}
    \caption{Results for ImageNet ($32\!\times\!32$).}
    \label{tab:my_label}
    \vspace{-0.5cm}
\end{table}

\section{A difference between FQGAN and Hard
Assignment}\label{sup:note_fqgan}
As we note in \S \ref{sec:code}, if $\mM$ is formed by k-means clustering,
Hard Assignment becomes an equivalent of the quantizer from FQGAN. However, the
FQGAN suffers from  discriminator overfitting, as we shown in Figures
\ref{figure:fid_C10}, \ref{figure:fid_1024_C10} and \ref{figure:fid_1024_C100}.
The FQGAN  uses the proximity loss Eq(\ref{eq:prox}), however, it has no
intertwining step between the manifold learning step with blocks of
discriminator. That is, for FQGAN, the feature input to the next layer from our
pipeline Eq(\ref{eq:enc2}) is changed to:
\vspace{-0.2cm}
\begin{equation}
\mX_{l+1}\!=\tmX_l
\vspace{-0.2cm}
\end{equation}

Without the interwining step, without the adaptive mechanism, and without LCSA,
the FQGAN cannot prevent blocks in discriminator from over-expressing the
learned features, which may lie outside of the manifold, resulting in
discriminator overfitting.

\section{Replacing LCSA with Denoising Auto-Encoder (DAE)}
\label{sup:dae}

\vspace{-0.3cm}
\begin{figure}[!h]
\centering
\includegraphics[width=8.0cm]{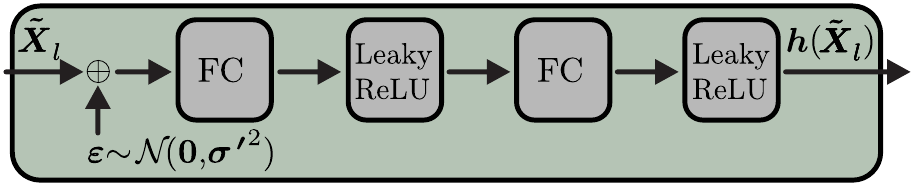}
%
\vspace{-0.3cm}
\caption{Architecture of DAE. We plug DAE in place  of LCSA in Figure \ref{fig:res_blcoks}: we replace $\vh(\cdot)$ based on LCSA with $\vh(\cdot)$ based on DAE.}
\label{fig:dae}
\vspace{-0.3cm}
\end{figure}

We conduct experiments in which we replace the LCSA coder with Denoising
Auto-Encoder(DAE) \cite{denoising_JMLR}. As illustrated in Figure \ref{fig:dae},
two FC layers, each intertwined with LeakyReLU, form our DAE module. Moreover,
we apply the mixing mechanism from  \eq{enc2} and the proximity
operator from \eq{prox} which is an equivalent of the reconstruction
loss of DAE given in \eq{dae}. As Figure \ref{fig:dae} indicates, DAE
is equipped with a noise injector $\boldsymbol{\varepsilon}\!\sim\!\mathcal{N}(
\boldsymbol{0},\boldsymbol{\sigma'}^2)$ with variance $\boldsymbol{\sigma'}^2$
(equivalent in  \eq{dae}). We use the multivariate normal distribution
which produces noise samples $\boldsymbol{\varepsilon}$ that are added to
$\mX'_l$, while $\boldsymbol{\sigma'}^2$ is a vector of on-diagonals equal
$\sigma'^2$ (off-diagonal coefficients equal 0).

Instead of $k$ (dictionary size), we vary the size of FC layers
$(d'\!\times\!k_*)$ and $(k_*\!\times\!d')$, by varying  the size of hidden
representation $k_*$. We investigate $k_*\!\in\!\{8, 16, 32, 64, 128, 256\}$ and
$\sigma'^2\!\in\!\{0.01, 0.04, 0.25, 0.64, 1.0\}$. After searching for the best
set of parameters, we obtain $k_*\!=\!64$, $\sigma'^2\!=\!0.04$ for CIFAR-10 and $k_*\!=\!32$, $\sigma'^2\!=\!0.04$ for CIFAR-100. 
The DAE module was updated once at each mini-batch.


Table \ref{table:dae} shows that OmniGAN+LCSA outperforms OmniGAN+DAE, which
validates our assumptions on locality-constrained non-linear manifold learning
and its ability to deal with the noise and regularization of discriminator.
\begin{table}[!htbp]
\vspace{-0.3cm}
    \centering
    \footnotesize
    \newcommand{\cs}{\hspace{0.1cm}}
    \begin{tabular}{|@{\cs}l@{\cs}|c@{\cs}|@{\cs}c@{\cs}|@{\cs}c@{\cs}|c@{\cs}|@{\cs}c@{\cs}|@{\cs}c@{\cs}|}
    \hline
    \multirow{2}{*}{Model} & \multicolumn{3}{c|}{CIFAR-10} & \multicolumn{3}{c|}{CIFAR-100} \\
    \cline{2-7}
    & IS $\uparrow$ & tFID $\downarrow$ & vFID $\downarrow$  & IS $\uparrow$ & tFID $\downarrow$ & vFID $\downarrow$\\
    \hline\hline
    OmniGAN & 9.70 & 6.88 & 10.65 & 12.78 & 9.13 & 13.82\\
    OmniGAN+DAE & 10.08 & 3.89 & 7.91 & 13.65 & 5.52 & 10.33\\
    OmniGAN+LCSA & \textbf{10.09} & \textbf{3.29} & \textbf{7.31} & \textbf{13.73} & \textbf{5.12} & \textbf{9.91}\\
    \hline
    \end{tabular}
    \vspace{-0.1cm}
    \caption{OmniGAN+DAE \vs OmniGAN+LCSA ($d'\!=\!512$).}
    \label{table:dae}
    \vspace{-0.3cm}
\end{table}

\vspace{-0.3cm}
\section{Fixed $\beta$ and $\gamma$}
\label{sup:fixx}
\begin{figure}[h]
\centering
\begin{subfigure}{0.485\linewidth}
\includegraphics[width=\linewidth]{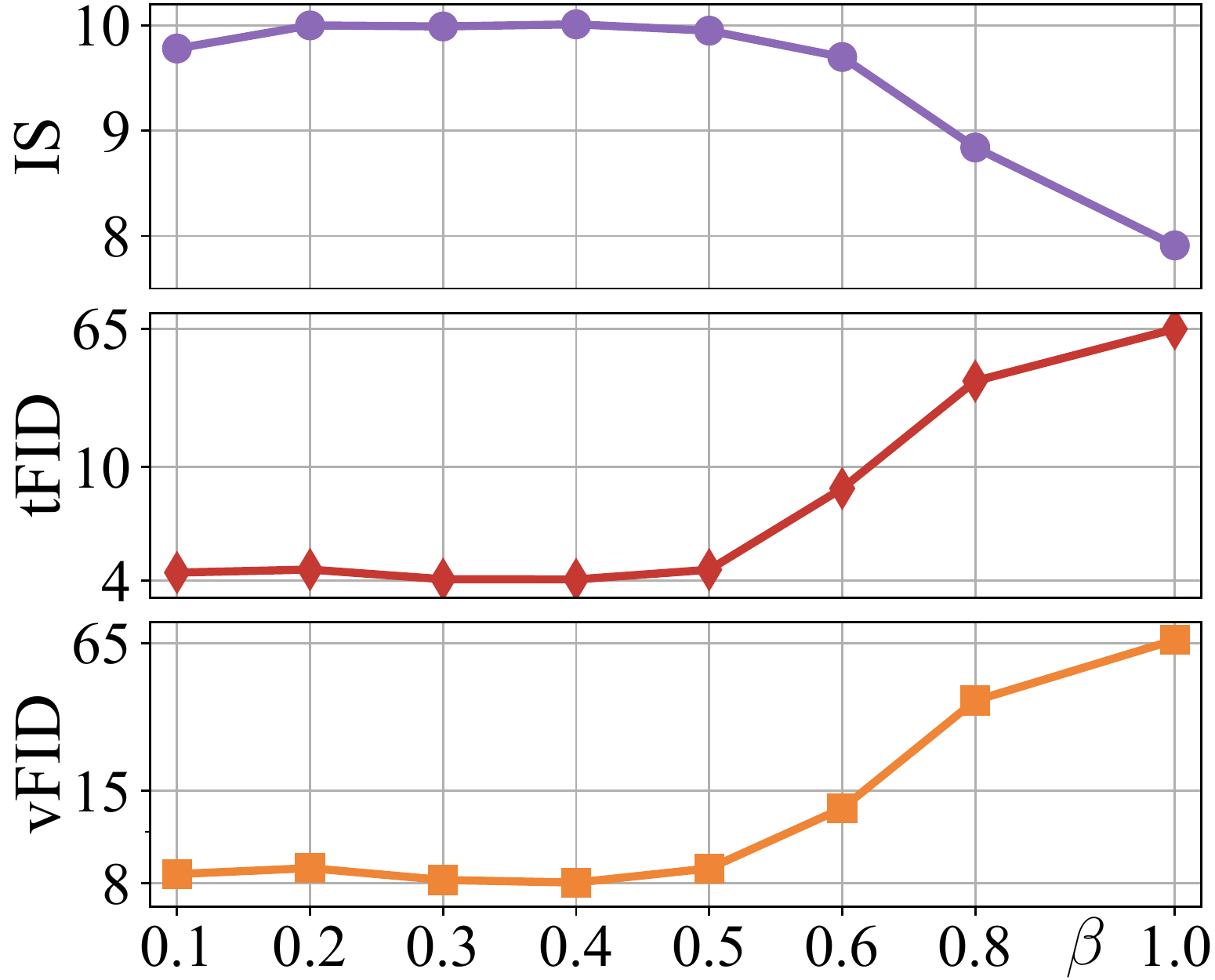}
\caption{CIFAR-10}
\end{subfigure}\hspace{0.1cm}
\begin{subfigure}{0.485\linewidth}
\includegraphics[width=\linewidth]{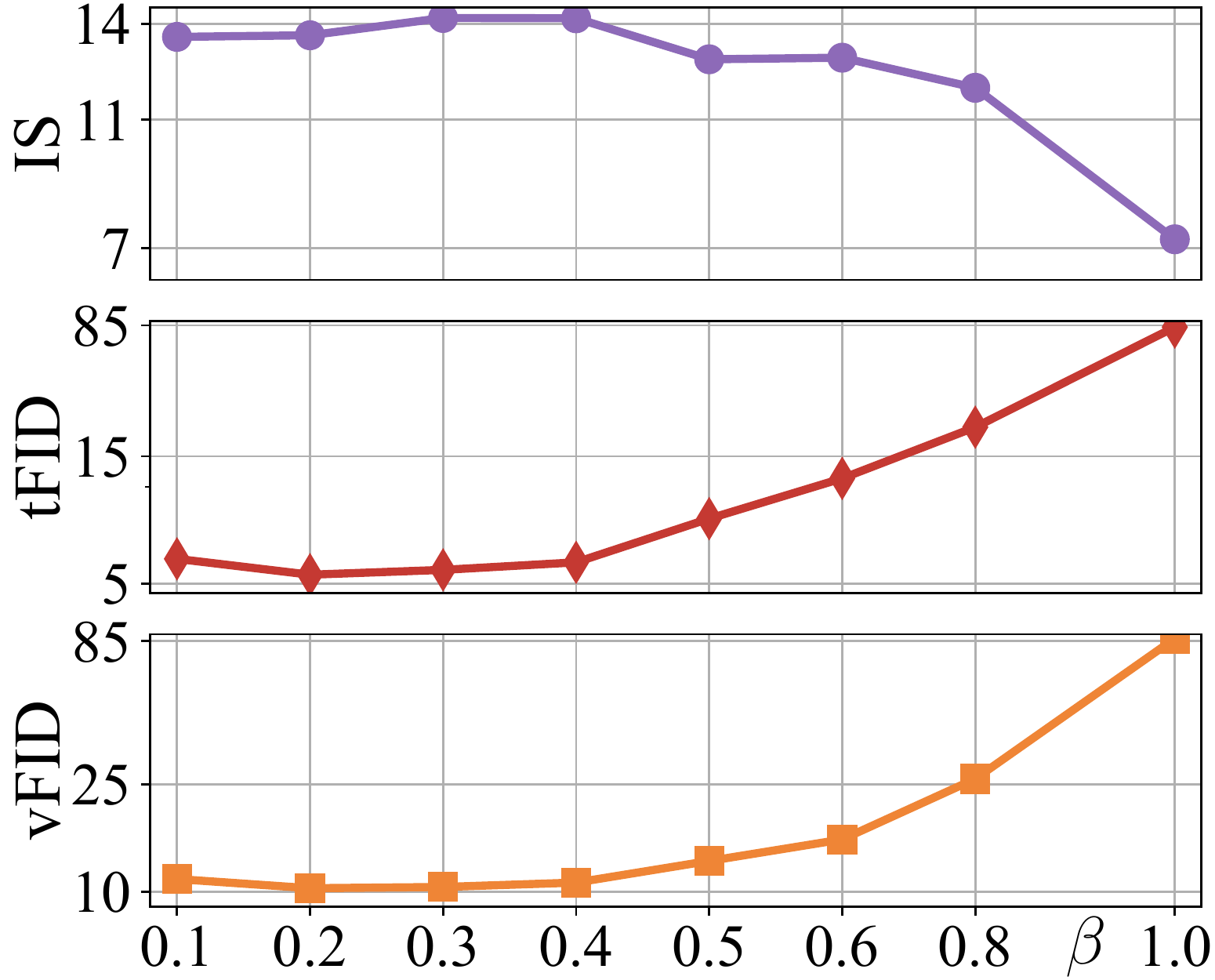}
\caption{CIFAR-100}
\end{subfigure}
\vspace{-0.3cm}
\caption{Fixed $\beta$ and $\gamma$, $\gamma=\Delta_{\gamma}\beta$ where
$\Delta_{\gamma}=1.2$ (as in the adaptive setting). We show results for several
fixed values of $\beta$.}
\label{fig:fixed_beta_gamma}
\vspace{-0.4cm}
\end{figure}

To verify that metaparameter learning of $\beta$ and $\gamma$ works better than setting fixed $\beta$ and $\gamma$, Table \ref{table:fixed_betaa} shows for the best possible fixed $\beta$. For each experiment, $\gamma\!=\!\Delta_{\gamma}\beta$ and $\Delta_{\gamma}\!=\!1.2$ (same as for learnable experiment). For fixed $\beta$ and $\gamma$, the best results are obtained with ($\beta\!=\!0.4$, $\gamma\!=\!0.48$) for CIFAR-10 and ($\beta\!=\!0.2$, $\gamma\!=\!0.24$) for CIFAR-100. Figure \ref{fig:fixed_beta_gamma} provides further ablations.

\begin{table}[h]
\vspace{0.2cm}
    \centering
    \footnotesize
    \newcommand{\cs}{\hspace{0.13cm}}
    \begin{tabular}{|@{\cs}l@{\cs}|c@{\cs}|@{\cs}c@{\cs}|@{\cs}c@{\cs}|c@{\cs}|@{\cs}c@{\cs}|@{\cs}c@{\cs}|}
    \hline
    \multirow{2}{*}{Method} & \multicolumn{3}{c|}{CIFAR-10} & \multicolumn{3}{c|}{CIFARR-100} \\
    \cline{2-7}
    & IS $\uparrow$ & tFID $\downarrow$ & vFID $\downarrow$ & IS $\uparrow$ & tFID $\downarrow$ & vFID $\downarrow$\\
    \hline\hline
    Fixed & 10.01 & 4.06 & 8.03 & 13.64 & 5.46 & 10.29 \\
    Learnable & \textbf{10.09} & \textbf{3.29} & \textbf{7.31} & \textbf{13.73} & \textbf{5.12} & \textbf{9.91}\\
    \hline
    \end{tabular}
    \vspace{-0.2cm}
    \caption{Best fixed $\beta$ and $\gamma$ \vs best metalearnable $\beta$ and
$\gamma$.}
    \label{table:fixed_betaa}
    \vspace{-0.1cm}
\end{table}

\section{Illustration of the residual error 
}
\label{sup:residual}
\begin{figure*}[!t]
    \centering
    \begin{subfigure}{0.24\linewidth}
    \includegraphics[width=\linewidth]{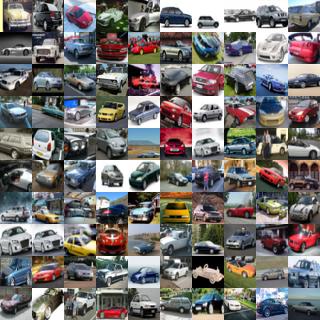}
    \caption{car}
    \end{subfigure}
    \begin{subfigure}{0.24\linewidth}
    \includegraphics[width=\linewidth]{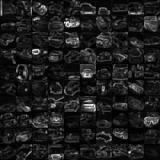}
    \caption{$\|\tmX_1-h(\tmX_1)\|_F^2$ of car}
    \end{subfigure}
    \begin{subfigure}{0.24\linewidth}
    \includegraphics[width=\linewidth]{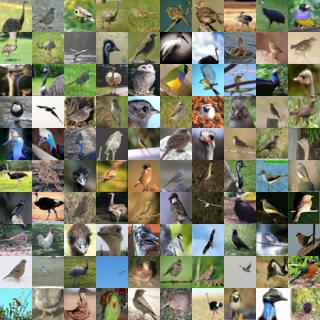}
    \caption{bird}
    \end{subfigure}
    \begin{subfigure}{0.24\linewidth}
    \includegraphics[width=\linewidth]{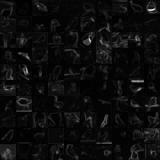}
    \caption{$\|\tmX_1-h(\tmX_1)\|_F^2$ of bird}
    \end{subfigure}
    \begin{subfigure}{0.24\linewidth}
    \includegraphics[width=\linewidth]{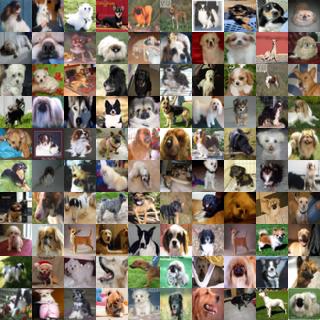}
    \caption{dog}
    \end{subfigure}
    \begin{subfigure}{0.24\linewidth}
    \includegraphics[width=\linewidth]{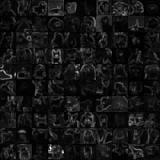}
    \caption{$\|\tmX_1-h(\tmX_1)\|_F^2$ of dog}
    \end{subfigure}
    \begin{subfigure}{0.24\linewidth}
    \includegraphics[width=\linewidth]{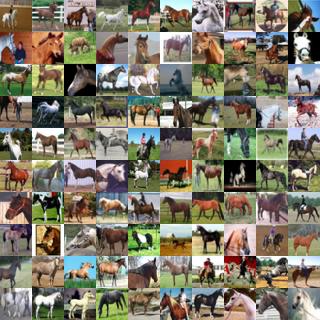}
    \caption{horse}
    \end{subfigure}
    \begin{subfigure}{0.24\linewidth}
    \includegraphics[width=\linewidth]{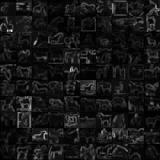}
    \caption{$\|\tmX_1-h(\tmX_1)\|_F^2$ of horse}
    \end{subfigure}
    \vspace{-0.2cm}
    \caption{Real samples and the corresponding $\|\tmX_1-h(\tmX_1)\|_F^2$ on
CIFAR-10. Whiter pixels correspond to larger $\|\tmX_1-h(\tmX_1)\|_F^2$. }
    \label{fig:visualization}
\end{figure*}
We randomly sample images from real dataset on CIFAR-10, and input them into the
discriminator. We obtain $\|\mX_1\!-\!\tmX_1\|_F^2$ at the first residual block
and normalize for visualization. 
Figure \ref{fig:visualization} shows LCSA coder perpetrates larger errors at
locations of difficult objects and smaller ones at locations of backgrounds. As
\eq{enc2}.

\section{Network architecture and Hyperparameters}
\label{sup:hyper}

Unless specified otherwise, we set $k\!=\!1024$, $\sigma=1.2$, $\gamma_0\!=\!0$, $\beta_0\!=0.1$, and $\eta=0.5$.
\vspace{0.05cm}

\noindent\textbf{CIFAR-10.} Below, we experiment with OmniGAN
\cite{zhou2020omni} which adopts the  architecture of BigGAN \cite{BigGAN}. 
We follow OmniGAN mini-batch size of 32 and train our networks for 500 epochs.
To this end, we equip each residual block of the discriminator with the LCSA
module. For our experiments on CIFAR-10, we apply weight decays of 1e-4, 1e-5
and 0 on generator given  $d'\!=\!256, 512, 1024$, weight decay 0 for the
discriminator. To update the dictionary, we use the Adam optimizer
\cite{kingma2014adam} with learning rate of 0.002. We set 
We use $\Delta_{\gamma}$ equal 1.0, 1.2 and 2.5  given  $d'\!=\!256, 512, 1024$. For StyleGAN2+ADA+LCSA, we follow \cite{ADA} and set $\eta=0.6$. We set $\Delta_\gamma=0.05$ due to the architecture differences between StyleGAN2 and BigGAN.

\vspace{0.05cm}
\noindent\textbf{CIFAR-100.} We use CIFAR-10 settings but we set the learning rate of DL to 0.0015, $k'\!=\!32$, and weight decays of 5e-4, 2e-4, 5e-5 for discriminators with $d'\!=\!256, 512, 1024$. 

\vspace{0.05cm}
\noindent\textbf{ImageNet ($\mathbf {64\!\times\!64}$).} We equip BigGAN and OmniGAN with LCSA (the last 3 blocks $l\!\in\!\{3,4,5\}$, $d'\!=\!384$). We use the architecture of BigGAN \cite{BigGAN}. For OmniGAN and OmniGAN+LCSA, we set mini-batch size to 256, learning rates of generator/discriminator to 4e-4, and take 4 discriminator steps per generator step to prevent OmniGAN from diverging. Weight decays are 0 and 2e-5 for generator and discriminator. For BigGAN, FQGAN and BigGAN+LCSA, we apply settings of FQGAN \cite{FQGAN} to set mini-batch size to 512, learning rates of generator and discriminator to 1e-4 and 4e-4. We set dictionary learning rate to 1e-3, 
$\Delta_{\gamma}\!=\!1.5$, $k'\!=\!8$. We set $\sigma\!=\!0.8$, $\beta_0\!=\!0$, $\gamma_0\!=\!0.1$ given the discriminator does not overfit on this diverse dataset at the early training stage.
All models are trained for 200 epochs.

\noindent\textbf{ImageNet ($\mathbf {128\!\times\!128}$)}. We only equip OmniGAN with LCSA as OmniGAN consistently outperforms BigGAN. We use setting from ImageNet ($64\!\times\!64$) but set dict. learning rate to 2e-4 and weight decay of discriminator to 5e-6. The two models are
trained for 300 epochs.

\vspace{0.05cm}
\noindent\textbf{Oxford-102 Flowers ($\mathbf {256\!\times\!256}$)}. We employ the state-of-the-art model for this dataset,  MSG-StyleGAN \cite{karnewar2020msg}, as our baseline. Following \cite{karnewar2020msg}, we set the size of mini-batch to 32 while $d'\!=\!512$. We set the number of iteration to $125K$, $k'\!=\!32$, $\Delta_{\gamma}\!=\!1$, $\eta=0.7$, dictionary learning rate is 5e-4,
and LCSA is applied to blocks 4 and 5 of discriminator.

\noindent\textbf{FFHQ ($\mathbf {256\!\times\!256}$)}. Following \cite{ADA}, we set the mini-batch size to 64, $d'\!\!=\!\!512$ and train the networks until the discriminator had seen $25M$ real images. We equip StyleGAN2 with LCSA (the last 4 blocks $l\!\in\!\{4,5,6,7\}$). We set $k'\!\!=\!\!32$, $\Delta_{\gamma}\!\!=\!\!0.05$, $\eta\!\!=\!\!0.9$, dictionary learning rate is 5e-4.

Table \ref{table:hyperparams} summarizes hyperparameters used in our
experiments. We used a very limited range of these parameters (most par. do not
change between experiments) \eg, dict. learning rate equals 1e-3, 1.5e-3, 2e-3, 5e-4 or 2e-4, $\beta_0$ is  set to 0 or 0.1, $\gamma_0$ is set  to 0 or 0.1,
$\Delta_\gamma$ is set to  0.05, 1, 1.2, 1.5 or 2.5, $\sigma$ is set to  0.8 or 1.2,
$\eta$ is set to  0.5, 0.6, 0.7 or 0.9, and $k'$ to 8, 32, or 128. 

Table \ref{table:time_cost} provides number of parameters, FLOPs and the training time for models with/without the LCSA coder.

\begin{table*}[h]
\centering
\footnotesize
\begin{tabular}{ccccccccccccc}
\toprule
dataset & $d'$ & Model  & dict LR & $\beta_0$ & $\gamma_0$ & $\Delta_{\gamma}$ &
$k'$ & $\sigma$ & $\eta$ & $L$ & GWD & DWD \\
\midrule
\multirow{9}{*}{CIFAR-10} & 256 & OmniGAN+LCSA & 2e-3 & 0.1 & 0 & 1.0 & 128 &
1.2 & 0.5 & \{1,2,3,4\} & 1e-4 & 0\\
\cline{2-13}
& \multirow{5}{*}{512} & BigGAN & $-$ & $-$ & $-$ & $-$ & $-$ & $-$ & $-$ & $-$
& 0 & 0\\
& & BigGAN+LCSA & 2e-3 & 0.1 & 0 & 1.2 & 128 & 1.2 & 0.5 & \{1,2,3,4\} & 0 &
0\\
& & OmniGAN & $-$ & $-$ & $-$ & $-$ & $-$ & $-$ & $-$ & $-$ & 1e-5 & 0\\
& & OmniGAN+LCSA & 2e-3 & 0.1 & 0 & 1.2 & 128 & 1.2 & 0.5 & \{1,2,3,4\} & 1e-5 &
0\\
& & StyleGAN2+ADA+LCSA & 2e-3 & 0.1 & 0 & 0.05 & 128 & 1.2 & 0.6 & \{1,2,3,4\} & 0 & 0\\
\cline{2-13}
& \multirow{4}{*}{1024} & BigGAN & $-$ & $-$ & $-$ & $-$ & $-$ & $-$ & $-$ & $-$
& 0 & 0\\
& & BigGAN+LCSA & 2e-3 & 0.1 & 0 & 2.5 & 128 & 1.2 & 0.5 & \{1,2,3,4\} & 0 &
0\\
& & OmniGAN & $-$ & $-$ & $-$ & $-$ & $-$ & $-$ & $-$ & $-$ & 0 & 0\\
& & OmniGAN+LCSA & 2e-3 & 0.1 & 0 & 2.5 & 128 & 1.2 & 0.5 & \{1,2,3,4\} & 0 &
0\\
\midrule
\multirow{9}{*}{CIFAR-100} & 256 & OmniGAN+LCSA & 1.5e-3 & 0.1 & 0 & 1.0 & 32 &
1.2 & 0.5 & \{1,2,3,4\} & 1e-4 & 5e-4\\
\cline{2-13}
& \multirow{4}{*}{512} & BigGAN & $-$ & $-$ & $-$ & $-$ & $-$ & $-$ & $-$ & $-$
& 0 & 0\\
& & BigGAN+LCSA & 1.5e-3 & 0.1 & 0 & 1.2 & 32 & 1.2 & 0.5 & \{1,2,3,4\} & 0 &
0\\
& & OmniGAN & $-$ & $-$ & $-$ & $-$ & $-$ & $-$ & $-$ & $-$ & 1e-5 & 2e-4\\
& & OmniGAN+LCSA & 1.5e-3 & 0.1 & 0 & 1.2 & 32 & 1.2 & 0.5 & \{1,2,3,4\} & 1e-5
& 2e-4\\
\cline{2-13}
& \multirow{4}{*}{1024} & BigGAN & $-$ & $-$ & $-$ & $-$ & $-$ & $-$ & $-$ & $-$
& 0 & 0\\
& & BigGAN+LCSA & 1.5e-3 & 0.1 & 0 & 2.5 & 32 & 1.2 & 0.5 & \{1,2,3,4\} & 0 &
0\\
& & OmniGAN & $-$ & $-$ & $-$ & $-$ & $-$ & $-$ & $-$ & $-$ & 0 & 5e-5\\
& & OmniGAN+LCSA & 1.5e-3 & 0.1 & 0 & 2.5 & 32 & 1.2 & 0.5 & \{1,2,3,4\} & 0 &
5e-5\\
\midrule
& \multirow{4}{*}{384} & BigGAN & $-$ & $-$ & $-$ & $-$ & $-$ & $-$ & $-$ & $-$
& 0 & 0\\
ImageNet & & BigGAN+LCSA & 1e-3 & 0 & 0.1 & 1.5 & 8 & 0.8 & 0.5 & \{3,4,5\} & 0
& 0\\
$64\times64$& & OmniGAN & $-$ & $-$ & $-$ & $-$ & $-$ & $-$ & $-$ & $-$ & 0 &
2e-5\\
& & OmniGAN+LCSA & 1e-3 & 0 & 0.1 & 1.5 & 8 & 0.8 & 0.5 & \{3,4,5\} & 0 &
2e-5\\
\midrule
ImageNet & \multirow{2}{*}{384} & OmniGAN & $-$ & $-$ & $-$ & $-$ & $-$ & $-$ &
$-$ & $-$ & 0 & 5e-6 \\
$128\times128$ & & OmniGAN+LCSA & 2e-4 & 0 & 0.1 & 1.5 & 8 & 0.8 & 0.5 &
\{4,5,6\} & 0 & 5e-6\\
\midrule
Oxford-102 & \multirow{2}{*}{512} & MSG-StyleGAN & $-$ & $-$ & $-$ & $-$ & $-$ &
$-$ & $-$ & $-$ & 0 & 0\\
Flowers& & MSG-StyleGAN+LCSA & 5e-4 & 0.1 & 0 & 1 & 32 & 1.2 & 0.7 & \{4,5\} &
0 & 0\\
\midrule
FFHQ & 512 & StyleGAN2+LCSA & 5e-4 & 0.1 & 0 & 0.05 & 32 & 1.2 & 0.9 & \{4,5,6,7\} & 0 & 0\\
\bottomrule
\end{tabular}
\vspace{-0.2cm}
\caption{Hyperparameter used in our experiments. Abbreviation `dict LR' denotes
dictionary learning rate, `GWD' and `DWD' are weight decays of generator and
discriminator, respectively.}
\label{table:hyperparams}
\vspace{-0.3cm}
\end{table*}

\begin{table}[t]
\begin{center}
\footnotesize
\newcommand{\cs}{\hspace{0.06cm}}
\begin{tabular}{|@{\cs}c@{\cs}@{\cs}c@{\cs}@{\cs}c@{\cs}|c@{\cs}|@{\cs}c@{\cs}|}
\hline
\multirow{3}{*}{Dataset} & \multirow{3}{*}{Baseline} & \multirow{3}{*}{GPUs} & Baseline & +LCSA\\\cline{4-5}
& & & \#Par. / FLOPs & \#Par. / FLOPs \\
& & & Training Time & Training Time \\
\hline
\multirow{2}{*}{CIFAR-10} & OmniGAN & \multirow{2}{*}{1} & 32.88M / 11.12G  & 34.97M / 12.04G  \\
& $(d'\!\!=\!\!512)$ & & 27h11m & 30h54m\\\hline
\multirow{2}{*}{CIFAR-100} & OmniGAN &  \multirow{2}{*}{1} & 33.48M / 11.12G & 35.57M / 12.04G \\
& $(d'\!\!=\!\!512)$ & & 27h29m & 30h32m \\\hline
ImageNet & \multirow{2}{*}{OmniGAN} & \multirow{2}{*}{4} & 115.69M / 18.84G  & 118.44M / 19.10G \\
$(64\!\!\times\!\!64)$ & & & 3d15h & 4d01h\\\hline
ImageNet & \multirow{2}{*}{OmniGAN} & \multirow{2}{*}{4} & 158.36M / 31.45G & 162.29M / 31.86G\\
$(128\!\!\times\!\!128)$ & & & 20d03h & 22d10h \\\hline
Oxford-102 & MSG- & \multirow{2}{*}{4} &  50.28M / 85.11G & 51.33M / 86.29G\\
Flowers & StyleGAN & & 22h46m & 23h09m\\\hline
\multirow{2}{*}{FFHQ} & \multirow{2}{*}{StyleGAN2} & \multirow{2}{*}{4} & 48.76M / 51.64G & 50.86M / 53.14G\\
& & & 58h42m & 59h51m\\\hline
\end{tabular}
\vspace{-0.2cm}
\caption{{Number of parameters/FLOPs (for both generator and discriminator) and training time for models with \vs without the LCSA encoder. Experiments were performed on NVIDA Tesla V100 GPUs.}}
\label{table:time_cost}
\end{center}
\end{table}


\section{Coding Methods}
\label{sup:code}

Algorithms \ref{code:nnsc} and \ref{code:sc} detail our implementations of
Non-Negative Sparse Coding (SC$_+$) and Sparse Coding (SC).

\algblock{while}{endwhile}
\algblock{for}{endfor}
\algtext*{endfor}

\algblock[TryCatchFinally]{try}{endtry}
\algcblockdefx[TryCatchFinally]{TryCatchFinally}{catch}{endtry}
	[1]{\textbf{except}#1}{}
\algcblockdefx[TryCatchFinally]{TryCatchFinally}{elsee}{endtry}
	[1]{\textbf{else}#1}{}
\algtext*{endwhile}
\algtext*{endtry}

\algblockdefx{ifff}{endifff}
	[1]{\textbf{if}#1}{}
\algtext*{endifff}
	
\begin{algorithm}[!h]
\caption{Non-negative Sparse Coding (SC$_+$).}
\label{code:nnsc}
{\bf Input:} $\mX, \mM$ for a given forward pass, 
$\kappa$: the $L_1$ norm penalty, $\iota$: the number of iterations,
$\omega$: the learning rate.
\begin{algorithmic}[1]
\State $\{\alpha_{in}\!\sim\!\mathcal{U}(1e\!-\!6,1)\}_{i\!=\!1,\cdots,k\text{
and } n\!=\!1,\cdots,N'}$ 
\State $\valpha\!\gets\!\left[\frac{\valpha_n}{\|\valpha_n\|_1+1e\!-\!6}
\right]_{n\!=\!1}^{_{N'}}$(concatenate $\valpha_n$ into matrix $\valpha$)
\for{ $i\!=\!1,\cdots,\iota$}
	\State $\Delta\valpha\!=\!2\mM^T\!(\mX\!-\!\mM\valpha)+\kappa$ (gradient computation)
	\State $\valpha\!\gets\!\valpha\!-\!\frac{\omega}{(1+i)^{0.3}}\Delta\valpha$
(stochastic gradient descent)
	\State $\valpha\!\gets\!\text{ReLU}(\valpha)$ (reprojection into the feasible
set)
\endfor
\end{algorithmic}
{\bf Output:} $\valpha$: sparse codes
\end{algorithm}

\begin{algorithm}[!h]
\caption{Sparse Coding (SC).}
\label{code:sc}
{\bf Input:} $\mX, \mM$ for a given forward pass, 
$\kappa$: the $L_1$ norm penalty, $\iota$: the number of iterations,
$\omega$: the learning rate.
\begin{algorithmic}[1]
\State $\{\alpha_{in}\!\sim\!\mathcal{U}(-1,1)\}_{i\!=\!1,\cdots,k\text{ and }
n\!=\!1,\cdots,N'}$ 
\State $\valpha_+\!=\!\text{ReLU}(\valpha)\!+\!1e\!-\!6$ and 
$\valpha_-\!=\!\text{ReLU}(-\valpha)\!+\!1e\!-\!6$
\State $\valpha_+\!\gets\!\left[\frac{\valpha_{+,n}}{\|\valpha_{+,n}\|_1+1e\!-
\!6}\right]_{n\!=\!1}^{_{N'}}$,  $\valpha_-\!\gets\!\left[\frac{
\valpha_{-,n}}{\|\valpha_{-,n}\|_1+1e\!-\!6}\right]_{n\!=\!1}^{_{N'}}\!\!\!\!
\!\!\!\!\!\!\!\!\!\!\!\!$
\for{ $i\!=\!1,\cdots,\iota$}
	\State $\Delta\valpha\!=\!2\mM^T\!(\mX\!-\!\mM(\valpha_+\!-\!\valpha_-))$
(compute gradient)
	\State $\Delta\valpha_+\!=\!\Delta\valpha+\kappa$ and
$\Delta\valpha_-\!=\!-\Delta\valpha+\kappa$
	\State $\valpha_+\!\gets\!\valpha_+\!-\!\frac{\omega}{(1+i)^{0.3}}\Delta
\valpha_+$ and $\valpha_-\!\gets\!\valpha_-\!-\!\frac{\omega}{(1+i)^{0.3}}\Delta
\valpha_-\!\!\!\!\!\!\!\!$
	\State $\valpha_+\!\gets\!\text{ReLU}(\valpha_+)$ and
$\valpha_-\!\gets\!\text{ReLU}(\valpha_-)$
\endfor
\State $\valpha\!\gets\!\valpha_+\!-\!\valpha_-$
\end{algorithmic}
{\bf Output:} $\valpha$: sparse codes
\end{algorithm}

\begin{algorithm}[!h]
\caption{Orthogonal Matching Pursuit (OMP).}
\label{code:omp}
{\bf Input:} $\mX, \mM$ for a given forward pass, 
$\tau$: the number of non-zero elements (iterations).
\begin{algorithmic}[1]
\State $\boldsymbol{E}\!=\!\mX$ (initialize matrix of residuals)
\State $\mM^{*}\!=\!\emptyset$ (active dictionary tensor
$\mM^{*}\!\in\!\mbr{d'\times 0 \times N'}$)
\for{ $i\!=\!1,\cdots,\tau$}
	\State $\boldsymbol{P}\!=\!\mM^T\!\boldsymbol{E}$ (projection of $\mX$ onto
residual $\boldsymbol{E}$)
	\State $\boldsymbol{\varphi^i}\!=\!\text{idx\_max}(|\boldsymbol{P}|)$ (index of
maximum coefficient 
	\State $\qquad\qquad\qquad\quad\;\;\;$per column)
	\State $\mM^*\!=\text{cat}(\mM^*, [\vm_{\varphi_1^i},\cdots,
\vm_{\varphi_{N'}^i}]; 1)$ (active atoms$\!\!\!\!$ 
	\State $\qquad$are added to tensor $\mM^{*}$ so that
$\mM^{*}\!\in\!\mbr{d'\times i \times N'})\!\!\!\!\!\!\!\!$
	\for{ $n\!=\!1,\cdots,N'$ (this loop is batch solved) }
	    \State $\bmM\!=\!\text{squeeze}(\mM_{:,:,n})$ (dict.
$\bmM\!\in\!\mbr{d'\times i}$ for $\vx_n$)$\!\!\!\!$
	    \State $\valpha_n\!=\!\text{solve}(\bmM^T\!\vx_n, \bmM^T\!\bmM)$ (system of
linear  
	    \State $\quad$equations solved by batch solver torch.solve($\cdot$)$\!\!\!\!$
	    \State $\boldsymbol{e}_n\!=\!\vx_n\!-\!\bmM\valpha_n$ (residual solved by
batch ope-$\!\!\!\!\!\!\!\!$
	    \State $\qquad\;\;\;$rations at once, that is, 
$\boldsymbol{E}\!\equiv\![\boldsymbol{e}_1,\cdots,\boldsymbol{e}_{N'}]$)$\!\!\!\!\!\!\!\!$
	\endfor
\endfor
\State $\valpha\!=\!\text{one\_hot}([\boldsymbol{\varphi^1,\cdots,\boldsymbol{
\varphi^\tau}}])\!\odot\![\valpha_1,\cdots,\valpha_{N'}]$
\end{algorithmic}
{\bf Output:} $\valpha$: sparse codes
\end{algorithm}

Algorithm \ref{code:omp} is our efficient implementation of the Orthogonal
Matching Pursuit (OMP) which is based on the batch support of PyTorch.
Operations {\em cat}, {\em squeeze}, {\em solve} and {\em one\_hot}  are
equivalent to PyTorch methods with the same names while  $\odot$ is the
element-wise multiplication.

\begin{algorithm}[!h]
\caption{Locality-constrained Linear Coding (LLC).}
\label{code:llc}
{\bf Input:} $\mX, \mM$ for a given forward pass, 
$k'$: the $k'$ nearest neighbors, $\rho$: a small reg. constant equal
$1e\!-\!6$.
\begin{algorithmic}[1]
\for{ $n\!=\!1,\cdots,N'$ (this loop is batch solved)}
    \State $\boldsymbol{C}_n\!=\!(\mM\!-\!\boldsymbol{1}\vx_n^T)(\mM\!-\!
\boldsymbol{1}\vx_n^T)^T$ (so-called data cov.)
	\State $\valpha_n\!=\!\text{solve}(\boldsymbol{1},
\boldsymbol{C}_n+\text{diag}(\boldsymbol{1})\!\cdot\!\rho)$ (matrix left div. by
$\boldsymbol{1}$)$\!\!\!\!\!\!\!\!$
	\State $\valpha_n\!\gets\!\valpha_n/\sum_j\alpha_{jn}$ (normalization)
\endfor
\State $\valpha\!=\![\valpha_1,\cdots,\valpha_{N'}]$ (concatenation of vectors
$\valpha_n$ into
\State $\qquad\qquad\qquad\qquad\!$matrix $\valpha$)
\end{algorithmic}
{\bf Output:} $\valpha$: locality-constrained linear codes
\end{algorithm}

\begin{algorithm}[!h]
\caption{Locality-constrained Soft Assignment (LCSA).}
\label{code:lcsa}
{\bf Input:} $\mX, \mM$ for a given forward pass, 
$k'$: the $k'$ nearest neighbors, $\sigma$: the bandwidth controlling the slope
rate.
\begin{algorithmic}[1]
\for{ $n\!=\!1,\cdots,N'$ (this loop is batch solved)}
    \State $\boldsymbol{\varphi}\!=\!\text{idx\_nn}(\vx_n; \mM, k')$ ($k'$
indexes  of $k'$ nearest $\vm_j$$\!\!\!\!$
    \State $\qquad\qquad\qquad\qquad\qquad\!\!\!$of $\vx_n$)
	\State $\valpha'_n\!=\!\valpha'(\vx_n;[\vm_{\varphi_1},\cdots,\vm_{
\varphi_{k'}}])$ (SA computed by
	\State $\qquad\qquad\qquad\qquad$\eq{gmm4} on local dictionary)
	\State $\valpha_n\!=\!\pi(\valpha'_n; \boldsymbol{\varphi},k)$ (deploy coeff.
of $\valpha'_n$ to locations
	\State $\qquad\qquad\qquad\qquad\quad\;$$\boldsymbol{\varphi}$ of $k$ dim.
zeroed vector)
\endfor
\State $\valpha\!=\![\valpha_1,\cdots,\valpha_{N'}]$ (concat. vectors
$\valpha_n$ into matrix $\valpha$)
\end{algorithmic}
{\bf Output:} $\valpha$: locality-constrained soft-assigned codes
\end{algorithm}

\vspace{0.5cm}
Algorithm \ref{code:llc} illustrates that Locality-constrained Linear Coding
(LLC) is computed with operations which enjoy the batch support of PyTorch. Once
more, operation {\em solve} is equivalent to the batch-wise linear solver of
PyTorch.

Algorithm \ref{code:lcsa} is a sketch implementation of Locality-constrained
Soft Assignment (LCSA). Function {\em idx\_nn} firstly computes distances
$\boldsymbol{D}$ of $\mX$ to $\mM$ as:
\begin{align}
&\boldsymbol{D}\!=\![\|\vx_1\|_2^2,\cdots,\|\vx_{N'}\|_2^2]\!\cdot\!
\boldsymbol{1}^T+\boldsymbol{1}\!\cdot\![\|\vm_1\|_2^2,\cdots,\|\vm_{k}\|_2^2]^T
\nonumber\\
&\qquad-2\mM^T\!\mX.
\end{align}
Subsequently, function {\em ktop} implemented in PyTorch selects $k'$ indexes of
$k'$ smallest distances from $\boldsymbol{D}$ for each $\vx_n$.

\begin{algorithm}[!h]
\caption{Dictionary Learning (DL).}
\label{code:dl}
{\bf Input:} $\mX, \mM, \valpha$ for a given forward pass, 
 $\omega$: the learning rate.
\begin{algorithmic}[1]
\State $\{m_{ij}\!\sim\!\mathcal{U}(-1,1)\}_{i\!=\!1,\cdots,d'\text{ and }
j\!=\!1,\cdots,k}$ 
\State $\mM\!\gets\!\left[\frac{\vm_j}{\|\vm_j\|_1+1e\!-\!6}\right]_{j\!=\!1}^{k}$(concat. of $\vm_j$ into matrix $\mM$)$\!\!\!\!\!\!\!\!$
\for{ $i\!=\!1,\cdots,\iota$}
	\State $\Delta\mM\!=\!2(\mX\!-\!\mM\valpha)\valpha^T$ (gradient computation)
	\State $\mM\!\gets\!\mM\!-\!\frac{\omega}{(1+i)^{0.3}}\Delta\mM$ (gradient descent)
\endfor
\end{algorithmic}
{\bf Output:} $\mM$: dictionary atoms
\end{algorithm}

Algorithm \ref{code:dl} illustrates how we perform the Dictionary Learning (DL)
step. While we could solve the Least Squares Problem in the closed-form, we
noted that a simple one step of SGD per mini-batch is sufficient to obtain good
results. Of course, where needed and/or requested by reviewers, we will refine
our claims and proofs further.

\begin{figure*}[t]
\newcommand{\picspace}{\hspace{0.01\linewidth}}
\centering
\begin{subfigure}{0.24\linewidth}
\includegraphics[width=\linewidth]{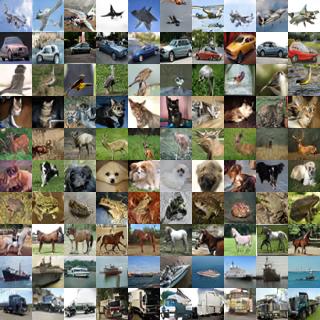}
\caption{OmniGAN, $d'\!\!=\!\!256$}
\end{subfigure}\picspace
\begin{subfigure}{0.24\linewidth}
\includegraphics[width=\linewidth]{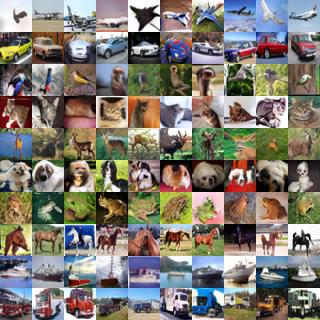}
\caption{OmniGAN+LCSA, $d'\!\!=\!\!256$}
\end{subfigure}\picspace
\begin{subfigure}{0.24\linewidth}
\includegraphics[width=\linewidth]{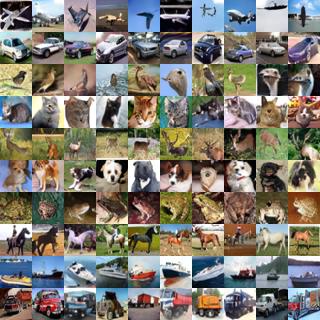}
\caption{OmniGAN+LCSA, $d'\!\!=\!\!512$}
\end{subfigure}\picspace
\begin{subfigure}{0.24\linewidth}
\includegraphics[width=\linewidth]{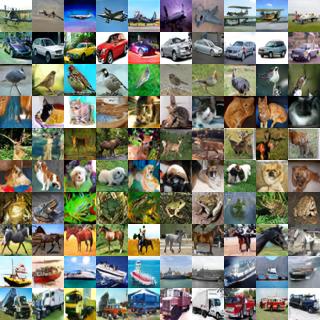}
\caption{OmniGAN+LCSA, $d'\!\!=\!\!1024$}
\end{subfigure}
\caption{Examples of images generated after training on CIFAR-10 ($32\times32$).}
\label{fig:gen_C10}
\end{figure*}

\section{Generated Images}
\label{sup:exim}
\begin{figure*}[h]
\newcommand{\picspace}{\hspace{0.01\linewidth}}
\centering
\begin{subfigure}{0.24\linewidth}
\includegraphics[width=\linewidth]{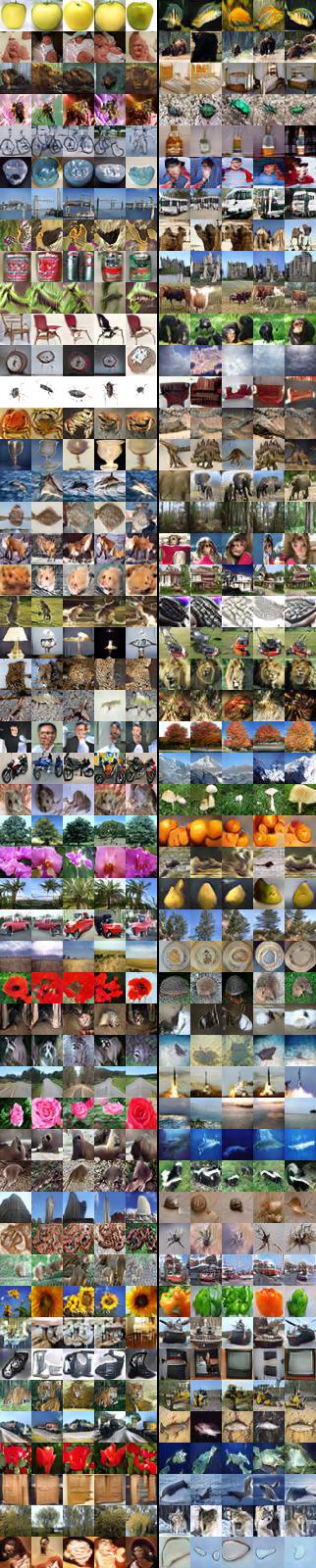}
\caption{OmniGAN, $d'\!\!=\!\!256$}
\end{subfigure}\picspace
\begin{subfigure}{0.24\linewidth}
\includegraphics[width=\linewidth]{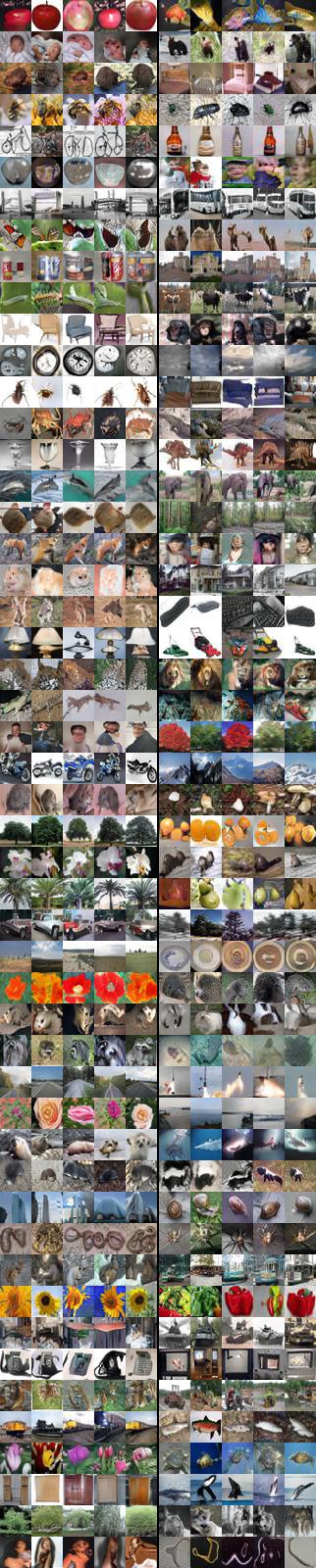}
\caption{OmniGAN+LCSA, $d'\!\!=\!\!256$}
\end{subfigure}\picspace
\begin{subfigure}{0.24\linewidth}
\includegraphics[width=\linewidth]{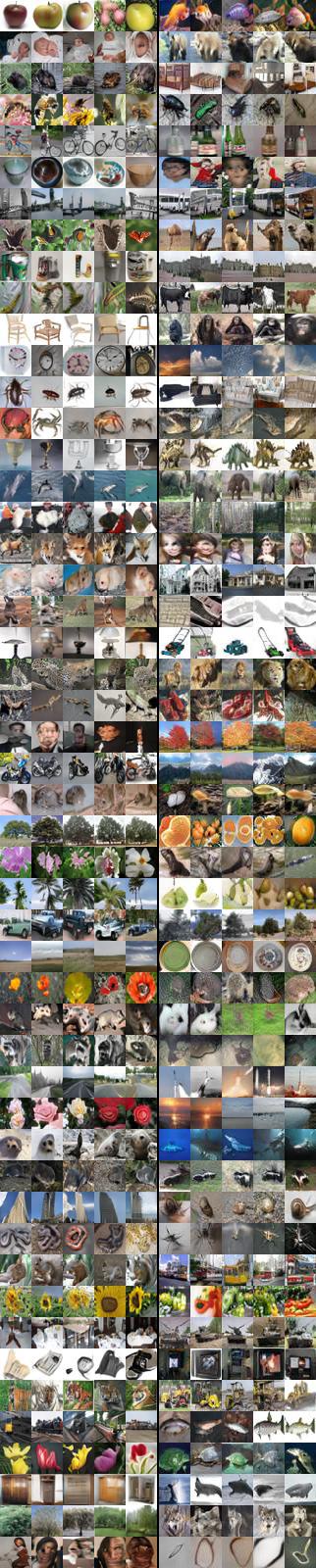}
\caption{OmniGAN+LCSA, $d'\!\!=\!\!512$}
\end{subfigure}\picspace
\begin{subfigure}{0.24\linewidth}
\includegraphics[width=\linewidth]{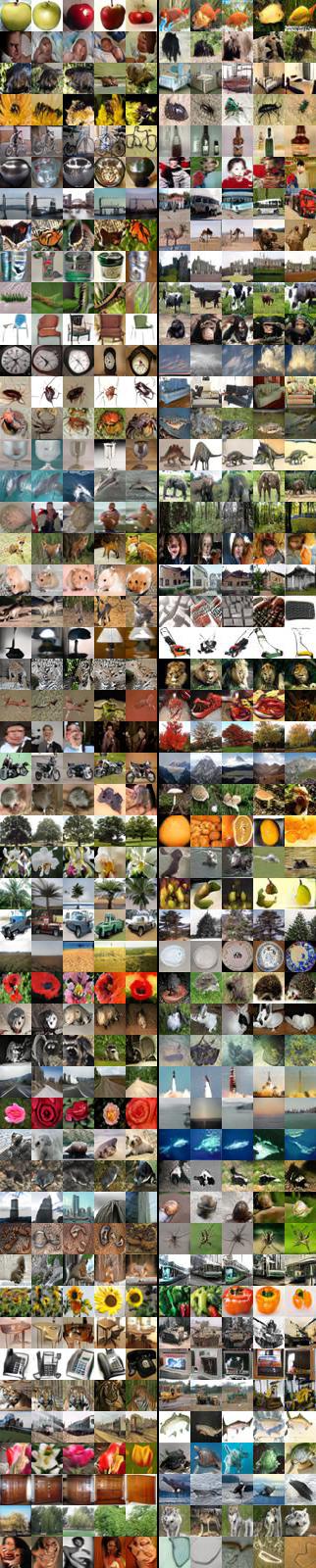}
\caption{OmniGAN+LCSA, $d'\!\!=\!\!1024$}
\end{subfigure}
    \caption{Examples of images generated after training on CIFAR-100
($32\times32$). Each row contains samples from two classes. We note that
OmniGAN+LCSA  produces more diverse images than OmniGAN.}
    \label{fig:gen_C100}   
\end{figure*}

\begin{figure*}[h]
\newcommand{\picspace}{\hspace{0.015\linewidth}}
\centering
\begin{subfigure}{0.485\linewidth}
\includegraphics[width=\linewidth]{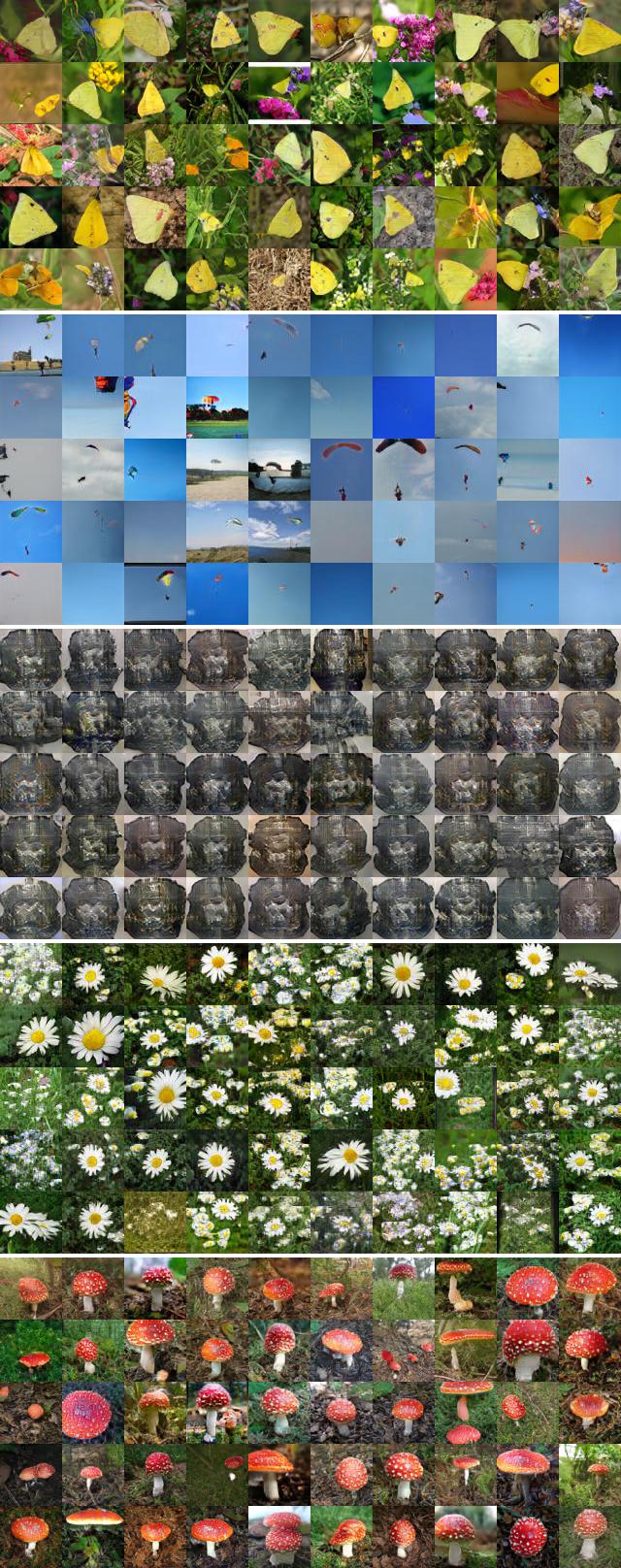}
\caption{BigGAN}
\end{subfigure}\picspace
\begin{subfigure}{0.485\linewidth}
\includegraphics[width=\linewidth]{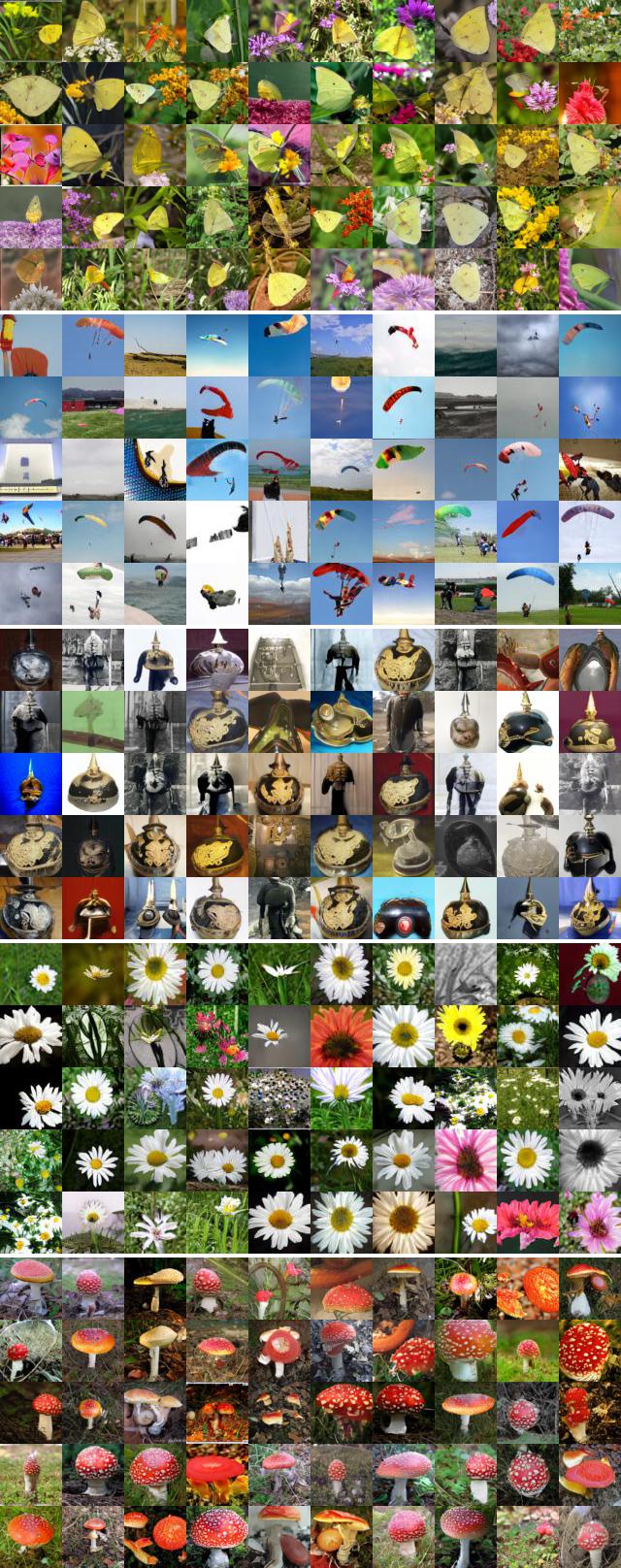}
\caption{BigGAN+LCSA}
\end{subfigure}
\vspace{-0.3cm}
\caption{Images generated by BigGAN and BigGAN+LCSA on ImageNet
(64$\times$64). Our BigGAN+LCSA can generate more diverse and realistic images
than BigGAN. We also note that the BigGAN has completely failed on class
\textit{pickelhaube} (the $3^{rd}$ row).}
\label{fig:imagenet64_proj}
\end{figure*}

\begin{figure*}[h]
\newcommand{\picspace}{\hspace{0.015\linewidth}}
\centering
\begin{subfigure}{0.485\linewidth}
\includegraphics[width=\linewidth]{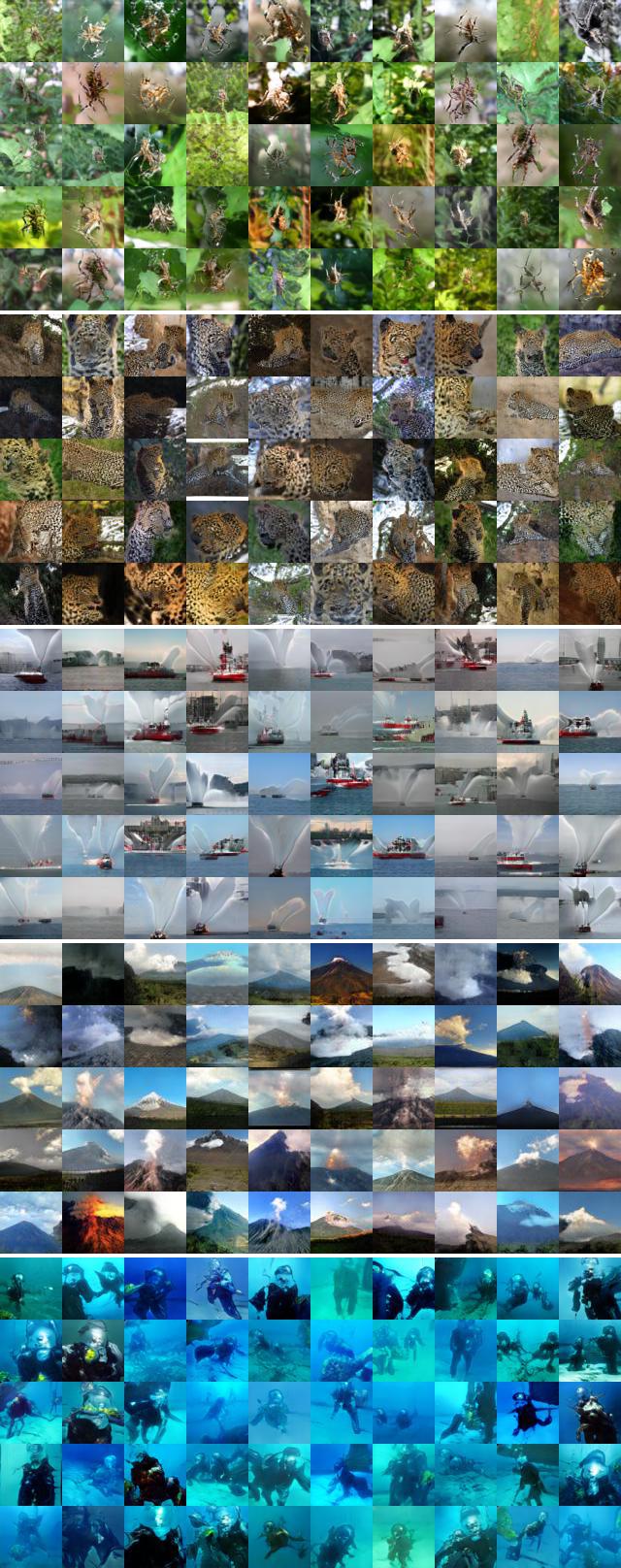}
\caption{OmniGAN}
\end{subfigure}\picspace
\begin{subfigure}{0.485\linewidth}
\includegraphics[width=\linewidth]{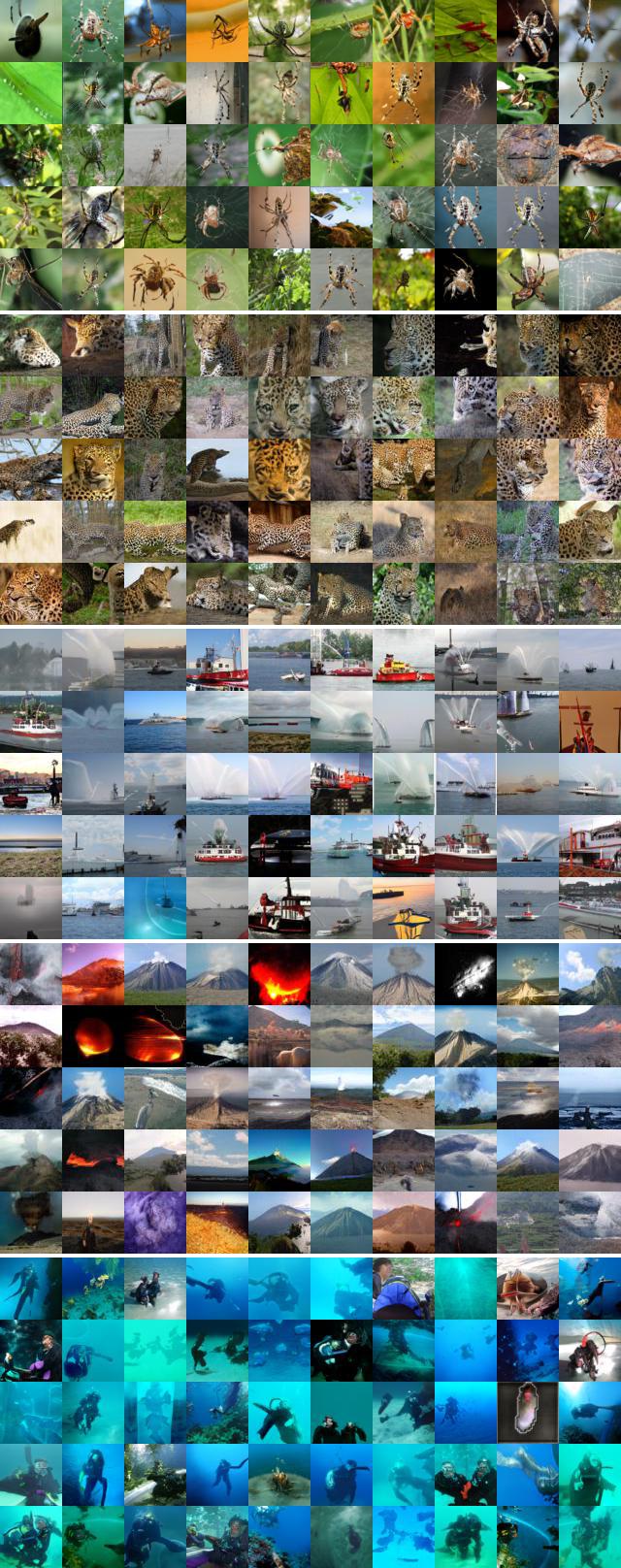}
\caption{OmniGAN+LCSA}
\end{subfigure}
\vspace{-0.3cm}
\caption{Images generated by OmniGAN and OmniGAN+LCSA on ImageNet
(64$\times$64). Our OmniGAN+LCSA produces more diverse and accurate
images than OmniGAN alone.}
\label{fig:imagenet64_omni}
\end{figure*}

\begin{figure*}[h]
\newcommand{\picspace}{\hspace{0.015\linewidth}}
\centering
\begin{subfigure}{0.485\linewidth}
\includegraphics[width=\linewidth]{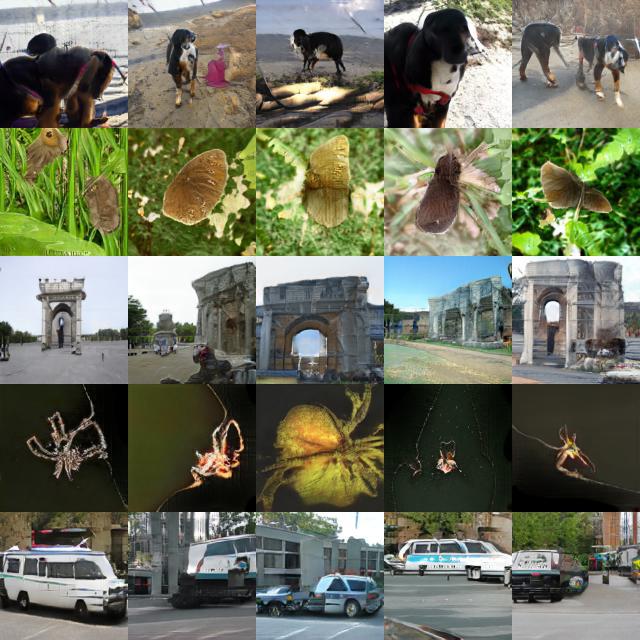}
\caption{OmniGAN}
\end{subfigure}\picspace
\begin{subfigure}{0.485\linewidth}
\includegraphics[width=\linewidth]{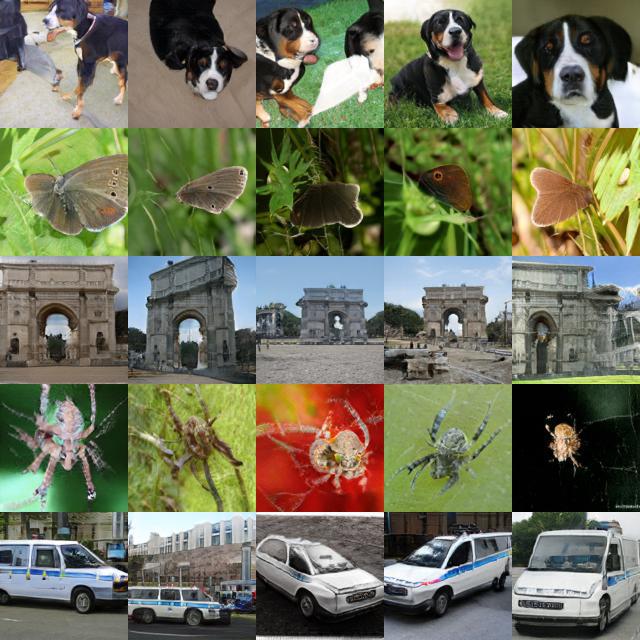}
\caption{OmniGAN+LCSA}
\end{subfigure}
\caption{Generated images on ImageNet (128$\times$128) dataset. Our
OmniGAN+LCSA can generate more diverse and realistic images than OmniGAN.}
\label{fig:imagenet128}
\end{figure*}

\begin{figure*}[h]
\newcommand{\picspace}{\hspace{0.015\linewidth}}
\centering
\begin{subfigure}{0.485\linewidth}
\includegraphics[width=\linewidth]{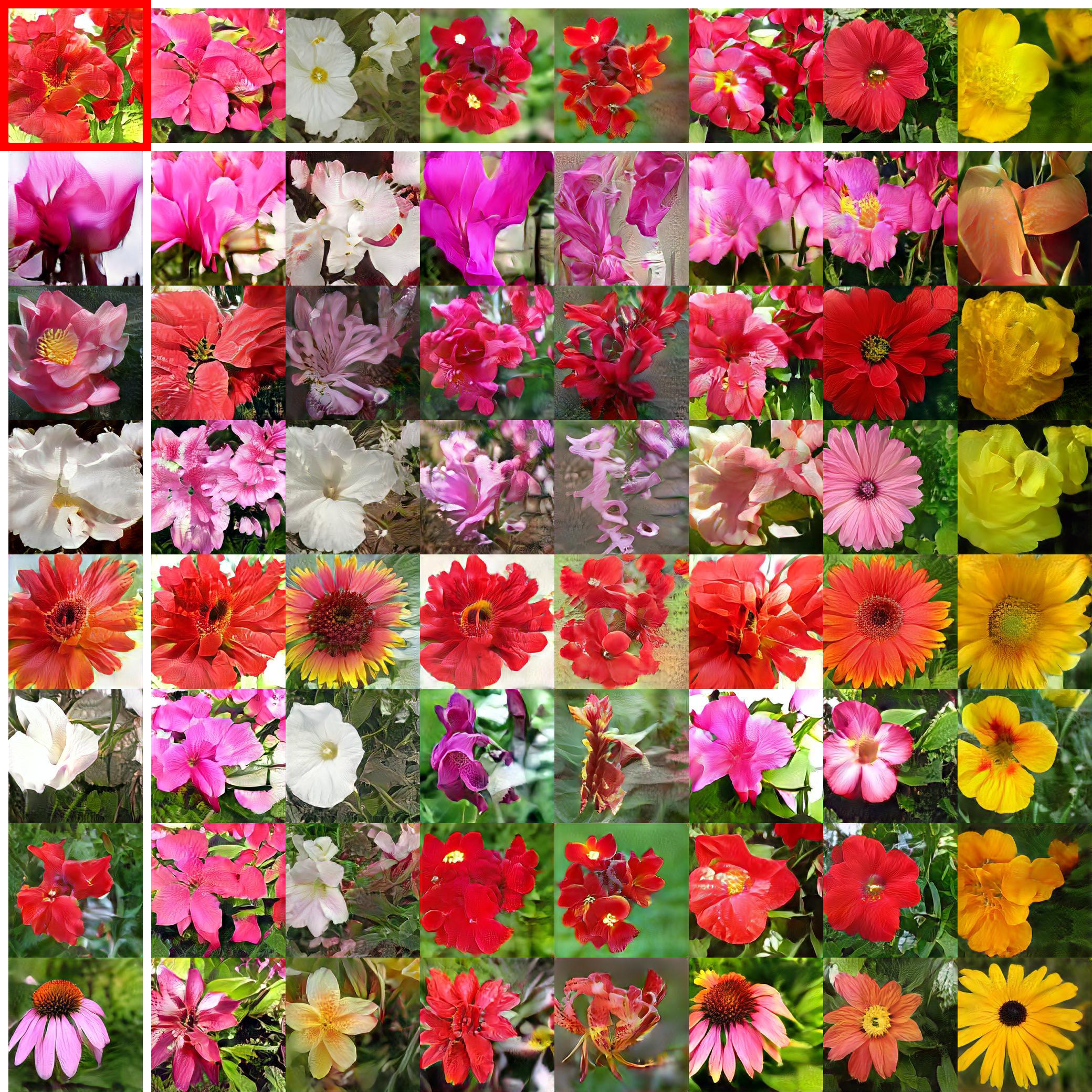}
\caption{MSG-StyleGAN}
\end{subfigure}\picspace
\begin{subfigure}{0.485\linewidth}
\includegraphics[width=\linewidth]{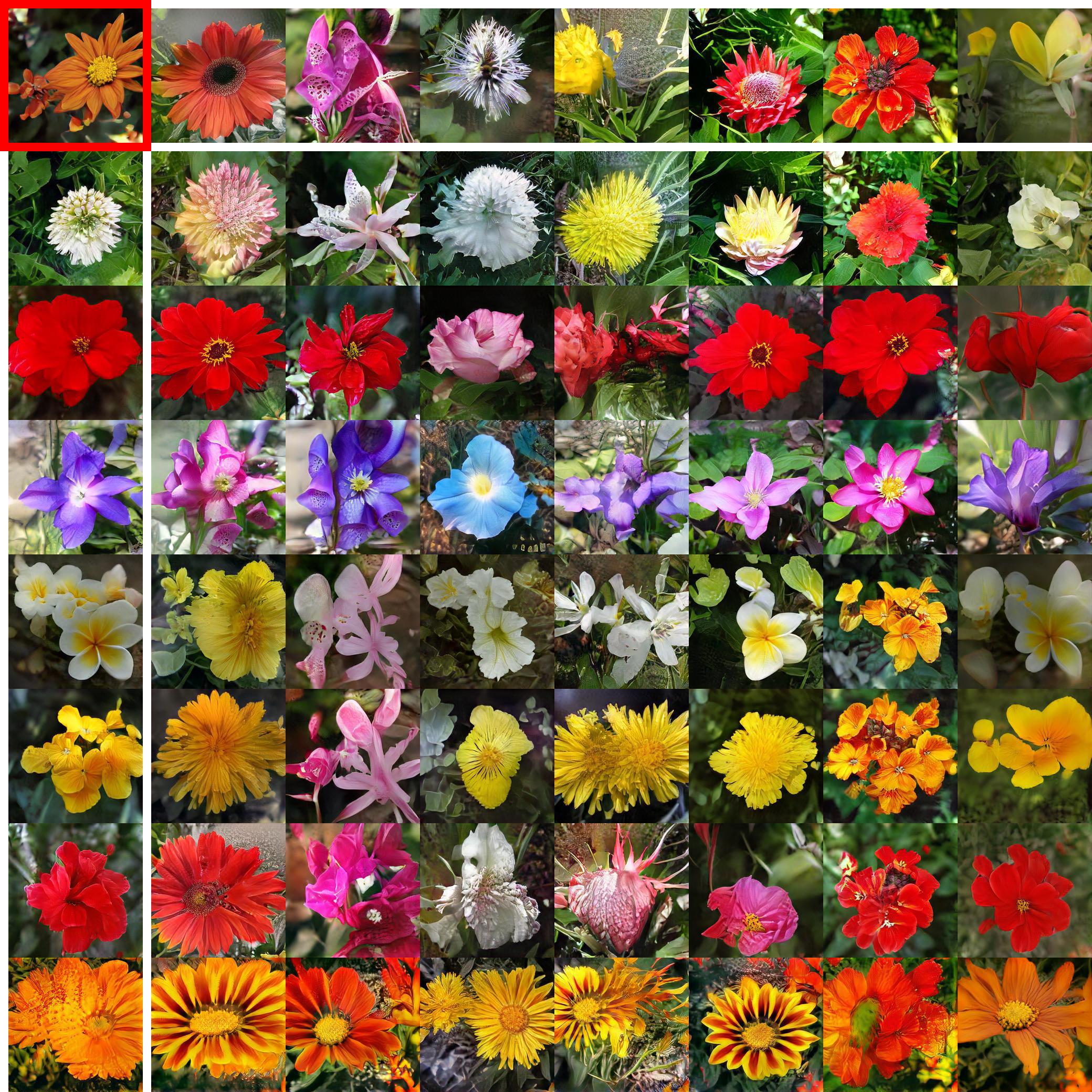}
\caption{MSG-StyleGAN+LCSA}
\end{subfigure}
\caption{Generated images using the mixing of different input noises on
Oxford-102 Flowers ($256\times256$). Two sets of images are generated from their
respective latent codes (the first row and the first column, respectively). The top left image
(red box) is generated using the averaged latent inputs from images in the first
row and the first column. The rest of images are generated by averaging the two
latent inputs of images chosen from the first row and the first column \ie,
$\mathbf{I}_{i,j}=G((\vz_{i,1}+\vz_{1,j})/2)$. We input the same noise codes to both
of the two models. Our MSG-StyleGAN+LCSA can generate more
diverse and accurate flowers than MSG-StyleGAN alone.}
\label{fig:flowers_interpolation}
\end{figure*}

\begin{figure*}[h]
\newcommand{\picspace}{\hspace{0.015\linewidth}}
\centering
\begin{subfigure}{0.485\linewidth}
\includegraphics[width=\linewidth]{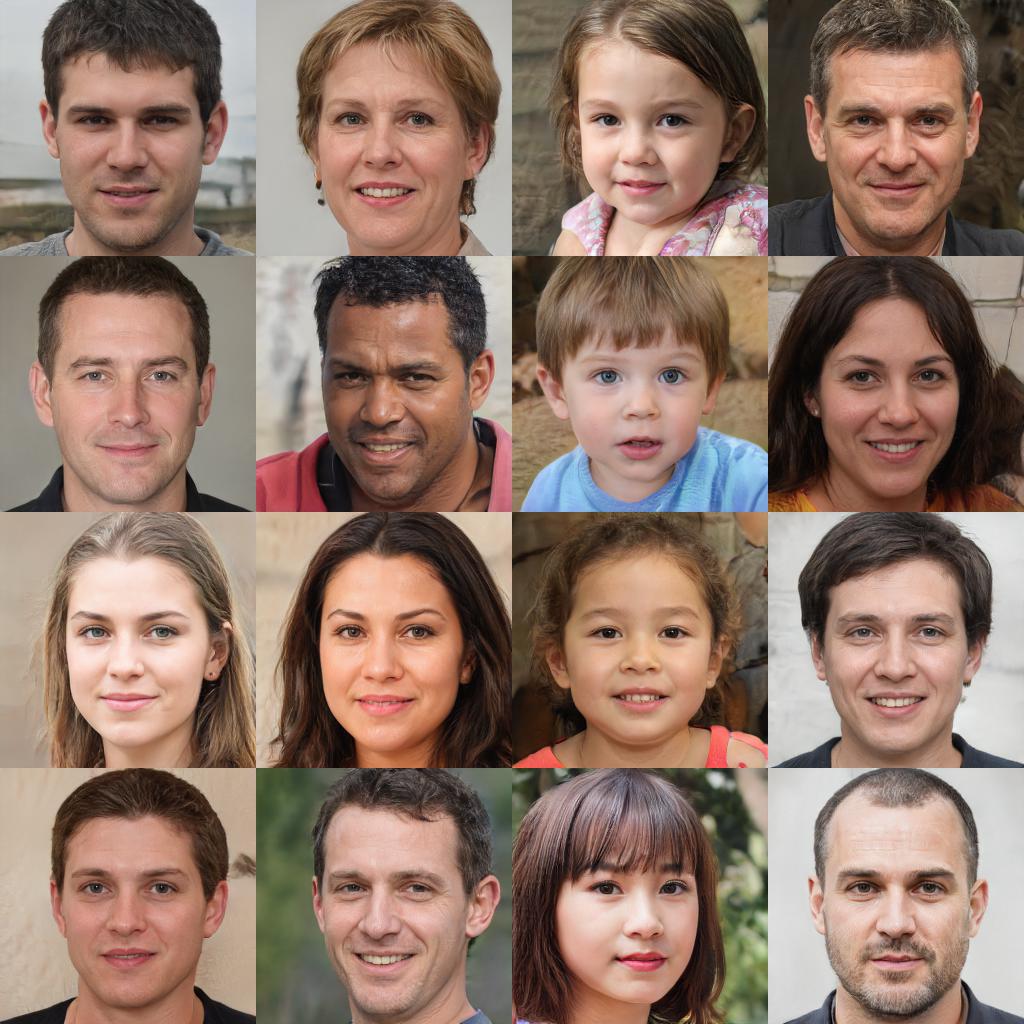}
\caption{StyleGAN2}
\end{subfigure}\picspace
\begin{subfigure}{0.485\linewidth}
\includegraphics[width=\linewidth]{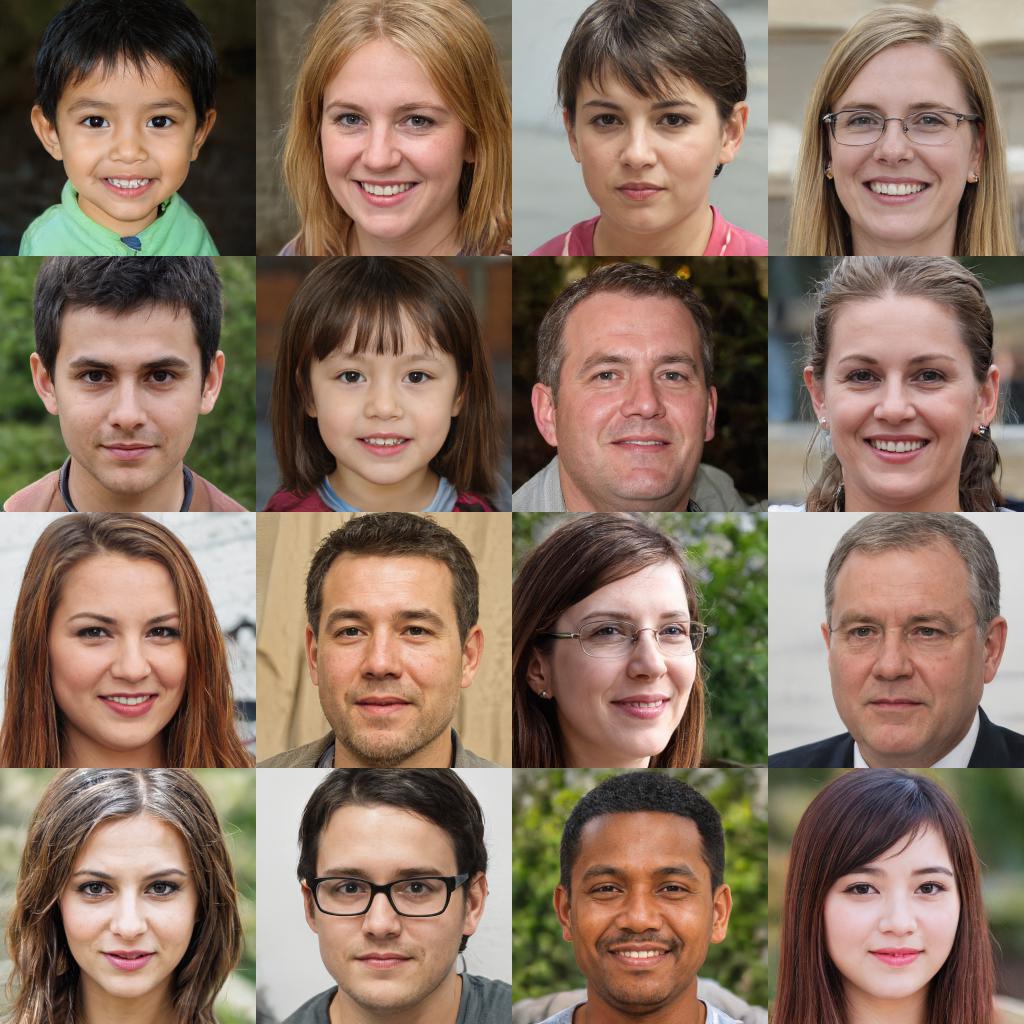}
\caption{StyleGAN2+LCSA}
\end{subfigure}
\caption{Examples of images generated by training on FFHQ ($256\times256$) $70K$ dataset with truncation trick using $\psi\!=\!0.5$ \cite{StyleGAN}.}
\label{fig:ffhq_trunc}
\end{figure*}

\begin{figure*}[h]
\newcommand{\picspace}{\hspace{0.015\linewidth}}
\centering
\begin{subfigure}{0.485\linewidth}
\includegraphics[width=\linewidth]{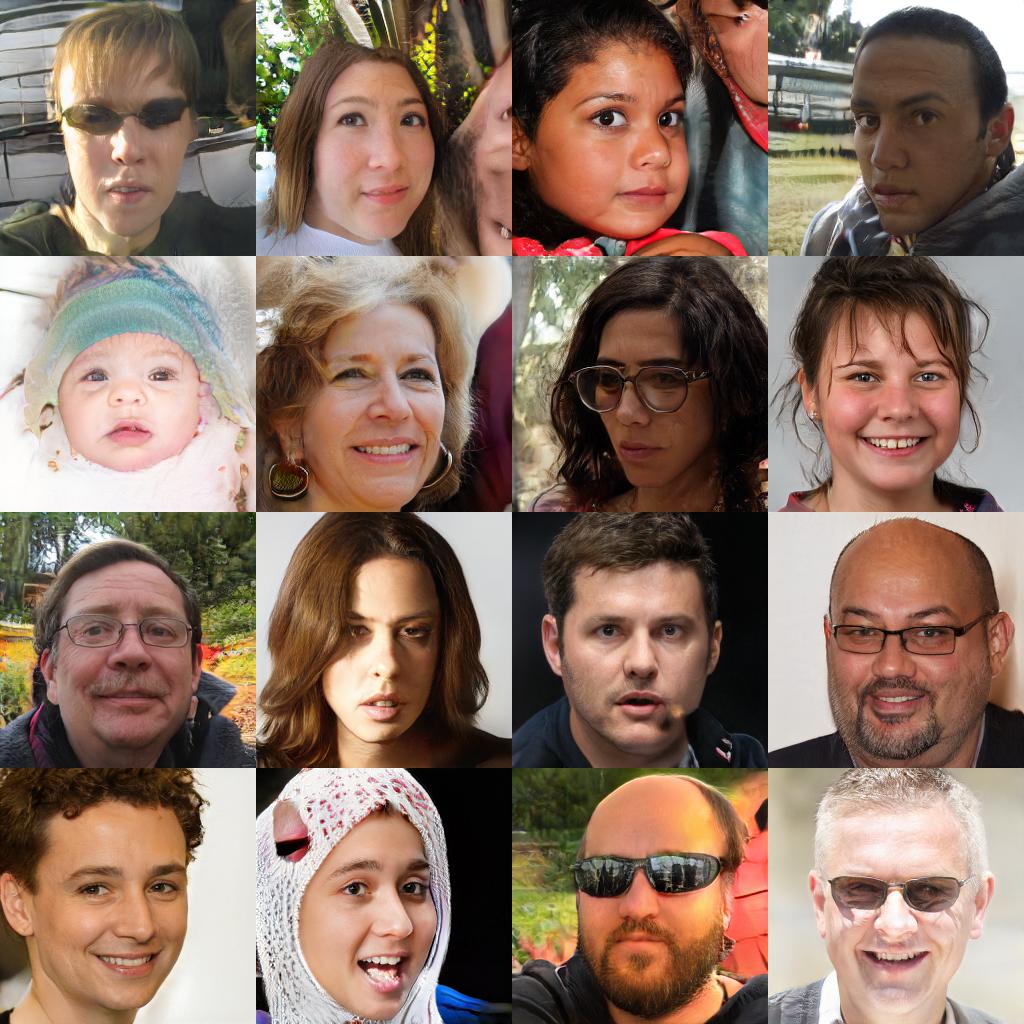}
\caption{StyleGAN2}
\end{subfigure}\picspace
\begin{subfigure}{0.485\linewidth}
\includegraphics[width=\linewidth]{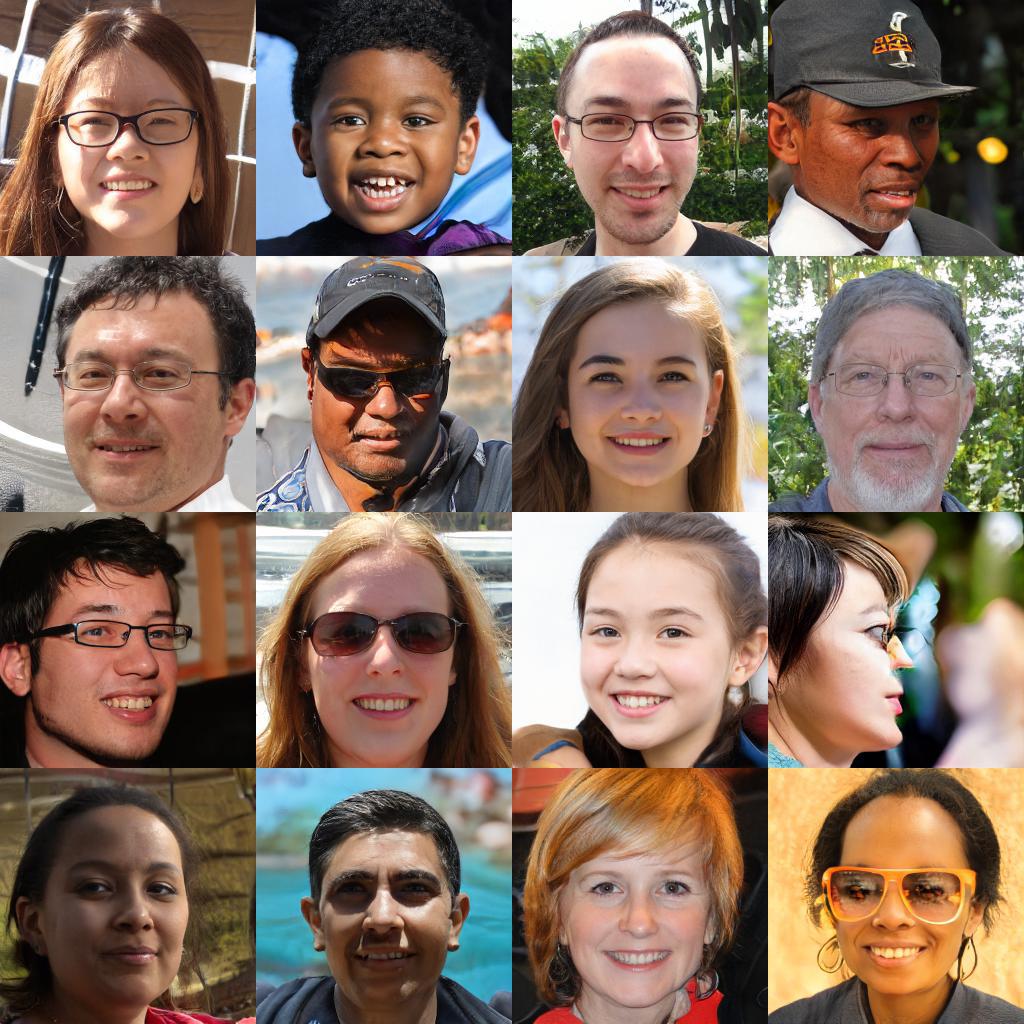}
\caption{StyleGAN2+LCSA}
\end{subfigure}
\caption{Examples of images generated by training on FFHQ ($256\times256$) $70K$ dataset without using the truncation trick. We notice that our StyleGAN2+LCSA can generate more realistic images than StyleGAN2 alone.}
\label{fig:ffhq}
\end{figure*}

\begin{figure*}[!ht]
\newcommand{\picspace}{\hspace{0.01\linewidth}}
\centering
\begin{subfigure}{0.24\linewidth}
\includegraphics[width=\linewidth]{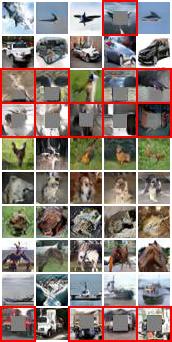}
\caption{DA (leak cutout)}
\end{subfigure}\picspace
\begin{subfigure}{0.24\linewidth}
\includegraphics[width=\linewidth]{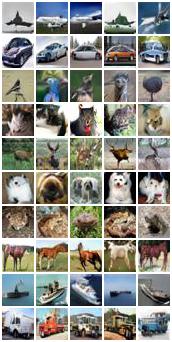}
\caption{DA+LCSA}
\end{subfigure}\picspace
\begin{subfigure}{0.24\linewidth}
\includegraphics[width=\linewidth]{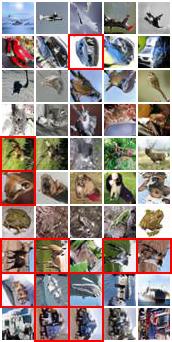}
\caption{ADA (leak rotation)}
\end{subfigure}\picspace
\begin{subfigure}{0.24\linewidth}
\includegraphics[width=\linewidth]{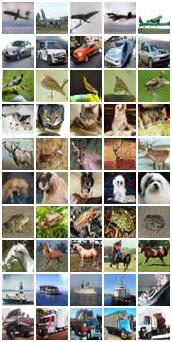}
\caption{ADA+LCSA}
\end{subfigure}
\caption{Generated images of different methods limiting overfitting on 10\%
CIFAR-10 data. \textbf{\color{red}DA leaks the cutout augmentation artifacts.} See  images enclosed with  red frames. \textbf{\color{red}ADA also leaks the
rotation augmentation artifacts. Combining LCSA with ADA or DA prevents the leakage of augmentation artifacts.}}
\label{fig:C10_lim10}
\end{figure*}

Figures \ref{fig:gen_C10}, \ref{fig:gen_C100}, \ref{fig:imagenet64_proj}, \ref{fig:imagenet64_omni}, and \ref{fig:imagenet128} show
examples of images generated with OminGAN \vs OmniGAN+LCSA after training on
CIFAR-10 ($32\!\times\!32$), CIFAR-100 ($32\!\times\!32$), ImageNet
($64\!\times\!64$) (we also include BigGAN \& BigGAN+LCSA) and ImageNet ($128\!\times\!128$). Figure \ref{fig:flowers_interpolation} shows examples of images generated with MSG-StyleGAN \vs
MSG-StyleGAN+LCSA  after training on Oxford-102 Flowers ($256\!\times\!256$). Figures \ref{fig:ffhq_trunc} and \ref{fig:ffhq} shows examples of images generated with StyleGAN2 \vs
StyleGAN2+LCSA for training on FFHQ ($256\!\times\!256$).

\begin{figure*}
\newcommand{\colwidth}{0.053\linewidth}
\newcommand{\blkspace}{\hspace{0.15cm}}
\newcommand{\picspace}{\hspace{0.03cm}}
\centering
\includegraphics[width=\colwidth]{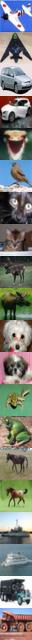}
\blkspace
\includegraphics[width=\colwidth]{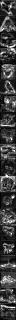}\picspace
\includegraphics[width=\colwidth]{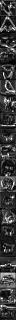}\picspace
\includegraphics[width=\colwidth]{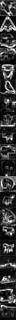}\picspace
\includegraphics[width=\colwidth]{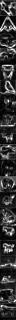}
\blkspace
\includegraphics[width=\colwidth]{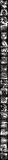}\picspace
\includegraphics[width=\colwidth]{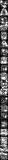}\picspace
\includegraphics[width=\colwidth]{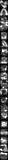}\picspace
\includegraphics[width=\colwidth]{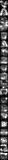}
\blkspace
\includegraphics[width=\colwidth]{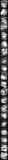}\picspace
\includegraphics[width=\colwidth]{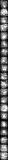}\picspace
\includegraphics[width=\colwidth]{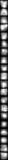}\picspace
\includegraphics[width=\colwidth]{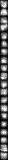}
\blkspace
\includegraphics[width=\colwidth]{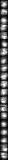}\picspace
\includegraphics[width=\colwidth]{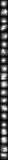}\picspace
\includegraphics[width=\colwidth]{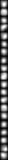}\picspace
\includegraphics[width=\colwidth]{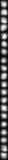}\\
\vspace{-0.3cm}
\caption{B1(LCSA LLC NNSC OMP), B2(LCSA LLC NNSC OMP), B3(LCSA\ \ LLC \ \ NNSC\ \
OMP), B4(LCSA\ \ LLC \ \ NNSC \ \ OMP). B1, B2, B3 and B4 denote the consecutive blocks of the discriminator.  The first column shows the chosen images and remaining columns show the variance of coefficients of each $\boldsymbol{\alpha}$. Notice that LCSA results in a slightly larger variance on foregrounds and edges of objects than other coders while still maintaining a low variance on fairly basic visual appearances such as uniform backgrounds or simple textures. \textbf{\color{red}We believe this is the example of the ability of LCSA to perform the quantization (backgrounds) \vs the approximate linear coding (foregrounds)  trade-off which is controlled with the Lipschitz constant $K$ (de facto $\sigma$).}}
\label{fig:var1}
\end{figure*}

\begin{figure*}
\newcommand{\colwidth}{0.093\linewidth}
\newcommand{\blkspace}{\hspace{0.15cm}}
\newcommand{\picspace}{\hspace{0.1cm}}
\centering
\begin{subfigure}{\colwidth}
\includegraphics[width=\linewidth]{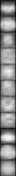}
\caption{LCSA}
\end{subfigure}
\begin{subfigure}{\colwidth}
\includegraphics[width=\linewidth]{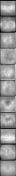}
\caption{LLC}
\end{subfigure}
\begin{subfigure}{\colwidth}
\includegraphics[width=\linewidth]{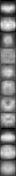}
\caption{NNSC}
\end{subfigure}
\begin{subfigure}{\colwidth}
\includegraphics[width=\linewidth]{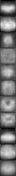}
\caption{OMP}
\end{subfigure}
\vspace{-0.2cm}
\caption{Mean over variance of each $\valpha$ on block B1.
The first 10 rows correspond
to the mean over variance of coefficients of each $\boldsymbol{\alpha}$ codes of each class, the last row is the mean over variance of coefficients of each $\boldsymbol{\alpha}$  over all classes. 
Notice that the combined variance of LCSA is again somewhat larger compared to NNSC and OMP but also lower than LLC in the areas of background. This indicates a good trade-off between quantization and reconstruction capacity of LCSA.
}
\label{fig:var2}
\end{figure*}

\input{proof}

\end{document}

%% file: proof.tex
\onecolumn

\section{Sketch Proofs of Claims from Section \ref{sec:lcsa}}
\label{sup:proofs}


\newcommand{\expect}{\mathbb{E}}
\newcommand{\mpp}{\boldsymbol{p}}
\newcommand{\mrr}{\boldsymbol{r}}
\newcommand{\mone}{\boldsymbol{1}}
\newcommand{\mq}{\boldsymbol{q}}
\newcommand{\ma}{\boldsymbol{a}}
\newcommand{\mb}{\boldsymbol{b}}
\newcommand{\mvv}{\boldsymbol{v}}
\newcommand{\mVV}{\boldsymbol{V}}

\begin{figure}[b]
\vspace{-0.5cm}
    \centering
    \begin{subfigure}{0.48\linewidth}
    \includegraphics[width=\linewidth]{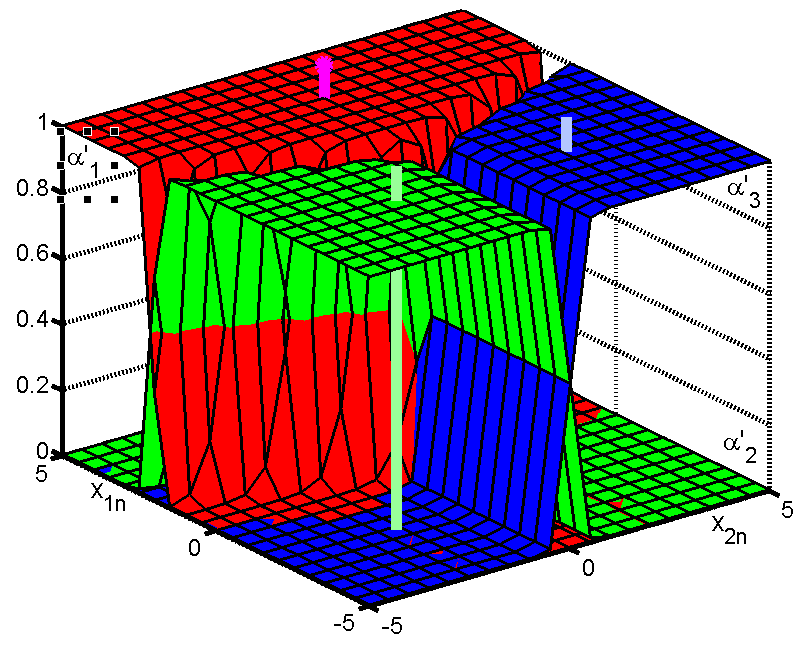}
    \caption{$\sigma^2\!=\!1$}
    \end{subfigure}
    \begin{subfigure}{0.48\linewidth}
    \includegraphics[width=\linewidth]{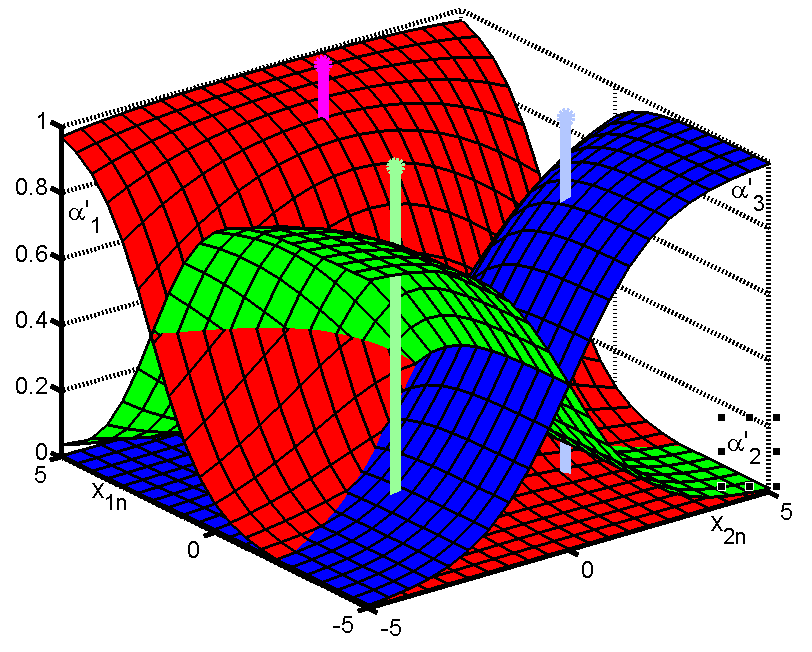}
    \caption{$\sigma^2\!=\!9$}
    \end{subfigure}
    \caption{Illustration of $\valpha$ of LCSA. Two-dimensional anchors
$[\vm_1,\vm_2,\vm_3]$ were used (three poles). LCSA was applied \wrt
$\vx\!\in\![-5,5]^2$. Responses $\valpha'\!=\![\alpha'_1,\alpha'_2,\alpha'_3]$
are given in three colors.}
    \label{fig:sig}
\end{figure}

\vspace{0.5cm}
Below we provide proofs and proof sketches of properties of LCSA. We note that
some of mathematics for $\valpha'$ functions is cumbersome, thus we resort to
key intuitions where necessary. We use one column display to ease readability of
some math formulas. Before embarking on the proofs, we develop key quantities
about $\valpha'$. Since it lives on the probability simplex, we compute
Shannon's entropy $S(\cdot) = -\expect_{\alpha'}[\log \alpha']$ of $\valpha'$
($\valpha$ being $\valpha'$ composed with a linear operator, its fundamental
geometric properties follow from $\valpha'$). We recall
\begin{eqnarray}
\valpha'(\vx;\mM,\sigma) & = & \frac{1}{Z(\vx; \mM,
\sigma)}\big[e^{-\frac{1}{2\sigma^2}||\vx\!-\!\vm_1||_2^2},\cdots,e^{-\frac{1}{2
\sigma^2}||\vx\!-\!\vm_k||_2^2}\big]^T
\label{eq:sa_rn}
\end{eqnarray}
After some algebra that we skip for readability, we arrive at the statistical
form of the gradient $\nabla S$ and Hessian $\mathrm{H}S$ at any $\vx$ (skipped
from notations for readability)
\begin{eqnarray*}
  \nabla S (\vx) & = & \frac{\expect_p [\delta]}{2 \sigma^2} \cdot (\expect_q
[\vm]-\expect_p [\vm]),\\
  \mathrm{H}S (\vx) & = & \frac{\expect_p [\delta]}{2 \sigma^4} \cdot \expect_q
[(\vm-\expect_p[\vm]) (\vm-\expect_p[\vm])^T] - \frac{2 + \expect_p [\delta]}{2
\sigma^4} \cdot \expect_p [(\vm-\expect_p[\vm]) (\vm-\expect_p[\vm])^T],
\end{eqnarray*}
where expectations are with respect to dictionary's columns, and for any such
column $\vm_i$, we let $\delta_i = \frac{ \|\vx\!-\!\vm_i\|_2^2}{\sigma^2}$ and
$p, q$ are the distributions in the $k$-dim probability simplex with $p_i
\propto \exp\left(-\frac{\|\vx\!-\!\vm_i\|_2^2}{2\sigma^2}\right)$ and $q_i
\propto p_i \delta_i$. From the shape operator of the entropy (see \textit{e.g.} \cite{pskcbTG}),
\begin{eqnarray*}
  \mathrm{s} S & = & -\frac{1}{\sqrt{1+\|\nabla S\|_2^2}} \cdot (\mathrm{I} +
\nabla S\nabla S^T)^{-1} \mathrm{H}S,
\end{eqnarray*}
we deduce the general formula for the mean curvature of $S$ (using
Sherman-Morrison Lemma to get rid of the inverse and properties of the trace),
\begin{eqnarray*}
  h S & = & \frac{1}{k \sqrt{1+\|\nabla S\|_2^2}} \cdot
\mathrm{tr}\left(\left(\mathrm{I} -\frac{1}{1+\|\nabla S\|_2^2} \cdot \nabla S
\nabla S^T\right)\cdot  -\mathrm{H}S\right)\\
  & = & \frac{1}{k (1+\|\nabla S\|_2^2)^{\frac{3}{2}}} \cdot \left[\nabla S^T
\mathrm{H}S \nabla S - \|\nabla S\|_2^2\cdot \mathrm{tr}(\mathrm{H}S)
-\mathrm{tr}(\mathrm{H}S)\right]. 
\end{eqnarray*}
For any vector $\ma$ on the $k$-dim probability simples, let us use the
shorthand $\vmu_a= \expect_a[\vm]$. Let us define, for any two such vectors
$\ma, \mb$, 
\begin{eqnarray}
  J(\ma,\mb) & = & \expect_a [\|\vm\|_2^2]\|\vmu_b-\vmu_a\|_2^2 - \expect_a
[((\vmu_a-\vmu_b)^T \vm)^2]+ (\vmu_b^\top \vmu_a)^2 - \|\vmu_a\|_2^2
\|\vmu_b\|_2^2.
\end{eqnarray}
Notice that Cauchy-Schwartz inequality brings
\begin{eqnarray}
  \expect_a [((\vmu_a-\vmu_b)^T \vm)^2] & \leq & \expect_a
[\|\vm\|_2^2]\|\vmu_b-\vmu_a\|_2^2,\label{cs1}\\
   (\vmu_b^\top \vmu_a)^2 & \leq & \|\vmu_a\|_2^2 \|\vmu_b\|_2^2, \label{cs2}
\end{eqnarray}
so we get a useful interval to which $ J(\ma,\mb)$ belongs: $ J(\ma,\mb) \in
\mathrm{I}(\vx;\mM,\sigma)$ with:
\begin{eqnarray}
 \mathrm{I}(\vx;\mM,\sigma) & = & \left[ - \left(\|\vmu_a\|_2^2 \|\vmu_b\|_2^2 -
(\vmu_b^\top \vmu_a)^2\right), \expect_a [\|\vm\|_2^2]\|\vmu_b-\vmu_a\|_2^2 -
\expect_a [((\vmu_a-\vmu_b)^T \vm)^2]\right],
\end{eqnarray}
and we note that those Cauchy-Schwartz inequalities \eqref{cs1}, \eqref{cs2}
also yield
\begin{eqnarray}
  0 & \in & \mathrm{I}(\vx;\mM,\sigma).
\end{eqnarray}
After some more derivations, we arrive at the expression 
\begin{eqnarray}
  h S & = & \frac{\left\{\begin{array}{c}
\left(\frac{\expect_p [\delta]}{2 \sigma^2}\right)^2 \cdot\frac{2 + \expect_p
[\delta]}{2 \sigma^4}\cdot \left(\expect_p [\|\vm\|_2^2]\|\vmu_q-\vmu_p\|_2^2 -
\expect_p [((\vmu_p-\vmu_q)^T \vm)^2]+ (\vmu_q^\top \vmu_p)^2 - \|\vmu_p\|_2^2
\|\vmu_q\|_2^2 \right) \nonumber\\
  -  \left(\frac{\expect_p [\delta]}{2 \sigma^2}\right)^2 \cdot\frac{\expect_p
[\delta]}{2 \sigma^4} \cdot  \left(\expect_q [\|\vm\|_2^2]
\|\vmu_q-\vmu_p\|_2^2- \expect_q [((\vmu_p-\vmu_q)^T \vm)^2]  +(\vmu_p^T \vmu_q
)^2-\|\vmu_p\|_2^2 \|\vmu_q\|_2^2 \right)
                         \end{array}\right\}}{k \left(1 + \left(\frac{\expect_p
[\delta]}{2 \sigma^2}\right)^2 \cdot \|\vmu_q-\vmu_p\|_2^2\right)^{\frac{3}{2}}}
\nonumber\\
  & = & \frac{(\expect_p [\delta])^2 }{k\sigma^2} \cdot \frac{(2 + \expect_p
[\delta])\cdot J(\mpp, \mq) -  \expect_p [\delta]  \cdot J(\mq,
\mpp)}{\left(4\sigma^4 + (\expect_p [\delta])^2 \cdot
\|\vmu_q-\vmu_p\|_2^2\right)^{\frac{3}{2}}}.\label{bmeanc}
\end{eqnarray}
We are now ready to develop the proofs (or proof sketches) to our various
Claims:\\

\vspace{0.5cm}
\noindent \textbf{Proof of Claim 1.} Since
\begin{eqnarray*}
  \frac{1}{(z+a)^{\frac{3}{2}}} & = & \frac{1}{a^{\frac{3}{2}}} - \frac{3z}{2
a^{\frac{5}{2}}} + o(z),
\end{eqnarray*}
we get the Taylor expansion of $h S$ in \eqref{bmeanc} as $\sigma\rightarrow 0$
($Q, R \neq 0$):
\begin{eqnarray*}
  h S & = & \frac{Q}{k\sigma^2} + \sigma^2 R + o(\sigma^2).
\end{eqnarray*}
So $hS$ diverges at rate $1/\sigma^2$ as $\sigma\rightarrow 0$, which shows that
the entropy's support approaches a vertex of the probability simplex (where its
curvature is maximal) and we get the HA solution.

\vspace{0.5cm}
\noindent \textbf{Proof of Claim 2.} We analyze cases for which $h S \rightarrow
0$. Looking at \eqref{bmeanc}, we see several scenarii: (i) $\expect_p [\delta]$
is small. This only happens when all the dictionary columns are close to each
other \textit{and} $\vx$ is close to them. In Fig. \ref{fig:sig}, this would
amount to brings all poles close to each other and would thus bring regions of
linearity,  i.e. the intersection of Voronoi cells, close to each other. Another
case, (ii), is when $\|\vmu_q - \vmu_p\|_2 \leq \epsilon$ with a decreasing
$\epsilon$, because then both bounds of interval $\mathrm{I}(\vx;\mM,\sigma)$
are $O(\epsilon^2)$ and so the mean curvature vanishes as well. Looking at
$\vmu_q - \vmu_p$, one can see that the norm of their differences vanishes in
particular when there are two sets of dictionary anchors, one which are far away
from $\vx$, and thus with low weight on both $p$ and $q$, and a second set,
closer to $\vx$ and such that all norms $\|\vm_. - \vx\|_2$ are approximately
constant, which precisely mean that $\vx$ is close to the center of the Voronoi
cell defined by those anchors. There is also a simple visual argument: Figure
\ref{fig:sig} shows that the rapid slope change takes place at
$\vmu_{12}\!=\!\frac{1}{2}(\vm_1\!+\!\vm_2),
\vmu_{13}\!=\!\frac{1}{2}(\vm_1\!+\!\vm_3),
\vmu_{23}\!=\!\frac{1}{2}(\vm_2\!+\!\vm_3)$ and
$\vmu_{123}\!=\!\frac{1}{3}(\vm_1\!+\!\vm_2\!+\!\vm_3)$. In fact, maximizing
over the absolute value of derivatives of each $\alpha'_i$ \wrt $\vx$, that is,
$\text{argmax}{_t}_{\vx'}\left|\frac{\partial\alpha'_i}{\partial\vx'}\right|$,
yields maximum at locations $\vmu_{\Lambda(k')}$, where operator
$\text{argmax}{_t}$ returns top $t\!=\!|\Lambda|$ maxima, $\Lambda$ returns all
possible ordered subsets of non-zero sizes of anchor indexes
$i\!=\!1,\cdots,{k'}$ \eg, for $k'\!=\!3$ we have $\{(1,2), (1,3),(2,3),
(1,2,3)\}\!=\!\Lambda(3)$. At $\{\vmu_{\Lambda(k')}\}$  locations, the maximum
indicates the largest slope changes which coincide with the linear slope regime
of sigmoid function.


\vspace{0.5cm}
\noindent\textbf{Proof of Claim 3.} To this end, we seek the stationary points of $\epsilon^2=\lVert\vx- \sum_k \vm_k \alpha_k(\vx; \mM,\sigma)\rVert_2^2$ 
where
\[
\alpha_k = \frac{\exp(-\| \vx - \vm_k \|^2 / 2\sigma^2)}{Z},
\]
just as in \eqref{eq:sa_rn} and $\vm_k$ are dictionary atoms defining the  Voronoi cell of $\vx$. By setting 
$\frac{\partial \epsilon^2}{\partial \sigma}=0$ 
we obtain
\[
-\frac{2}{\sigma^3}(\vx-\mM\valpha(x))\Big(\sum_i\vm_i\alpha_i(\vx)\big(\lVert\vx-\vm_i\rVert_2^2) - c\big)  \Big) =0 
\]
%
where $c=\sum_k\lVert\vx-\vm_k\rVert_2^2\alpha_k(x)$.
It is trivial to see that the first maximum is at the stationary point $\sigma=\infty$. Moreover, for $\sigma=0$ we have also the stationary point as
\begin{align}
&-\frac{2}{\sigma^3}(\vx-\mM\valpha(x))\Big(\sum_i\vm_i\alpha_i(\vx)\big(\lVert\vx-\vm_i\rVert_2^2) - c\big)  \Big) =0\\
&-\frac{2}{\sigma^3}\big(\vx-\mM\valpha(x))\Big((\sum_i\vm_i\alpha_i(\vx)\lVert\vx-\vm_i\rVert_2^2\big) - \big(\sum_i\vm_i\alpha^2_i(\vx)\lVert\vx-\vm_i\rVert_2^2\big)\Big)=0\\
& \text { because if $\sigma=0$ then }\nonumber\\
& \sum_i\vm_i\alpha_i(\vx)\lVert\vx-\vm_i\rVert_2^2= \sum_i\vm_i\alpha^2_i(\vx)\lVert\vx-\vm_i\rVert_2^2.
\end{align}

The last condition follows from the simplification of $c$ due to the fact that for $\sigma=0$, $\valpha$ codes obey the Hard Assignment, that is only one coefficient of $\valpha$ is equal one, the rest are equal zero. 

It is easy to see the other stationary point occurs if $\vx=\mM\valpha(x)$ which we already identified as the case of perfect reconstruction.

The last set of stationary points needs to satisfy
\[
\sum_i\vm_i\alpha_i(\vx)\Big(d^2_i-\sum_k d^2_k\alpha_k(\vx)\Big)  =0,
\]
where $d^2_i=\lVert\vx-\vm_i\rVert_2^2$ (for $d^2_k$ just substitute $k$ in place of $i$), 
and for these points we have checked via simulations that they yield the minimum reconstruction error.

\vspace{0.5cm}
\noindent\textbf{Proof of Claim 4 (Lipschitz condition).} 
Define $\vy = \mM \valpha(\vx) = h (\vx)$, where $\valpha$ is the LCSA mapping.
The task is to compute the Jacobian matrix $\partial \vy / \partial \vx$. Therefore
we compute $\partial y_i / \partial x_j$.

We have 
\begin{equation}
\label{eq:vy}
\vy = \sum_k \vm_k \alpha_k
\end{equation}
where
\[
\alpha_k = \frac{\exp(-\| \vx - \vm_k \|^2 / 2\sigma^2)}{Z}
\]
and $Z$ is defined so that $\sum_k \alpha_k = 1$.
From this we get
\begin{align}
\begin{split}
\label{eq:da_kZ/dx_j}
\frac{\partial (\alpha_k Z)}{\partial x_j} &= \exp(-\| \vx - \vm_k \|^2 / 2\sigma^2) ~~ (m_{kj} - x_j) / \sigma^2\\
&= Z \alpha_k (m_{kj} - x_j) / \sigma^2,
\end{split}
\end{align}
where $m_{kj}$ is the $j$-th component of $\vm_k$.

In addition, since $Z = \sum_k \alpha_k Z$, summing \eqref{eq:da_kZ/dx_j} we get
\begin{equation}
\label{eq:dZ/dx_j}
\frac{\partial Z}{\partial x_j} = \sum_k Z \alpha_k (m_{kj} - x_j)/ \sigma^2 ~.
\end{equation}

From \eqref{eq:vy} we have,
\begin{align*}
y_i &= \frac{\sum_k m_{ki} (\alpha_k Z) }{Z} ~,
\end{align*}
where we have multiplied top and bottom by $Z$, for convenience.
We differentiate this as a quotient, using \eqref{eq:da_kZ/dx_j} and
\eqref{eq:dZ/dx_j}, giving
\begin{align}
\sigma^2 \frac{\partial y_i}{\partial x_j} &= \frac{Z \big(\sum_k m_{ki} Z\alpha_k (m_{kj} - x_j)\big) - \big(\sum_k m_{ki} (\alpha_k Z)\big) 
\big(\sum_k Z \alpha_k (m_{kj} - x_j)\big)}{Z^2} \nonumber\\
&= \sum_k m_{ki} \alpha_k (m_{kj} - x_j) - \Big(\sum_k m_{ki}\alpha_k\Big) 
\Big(\sum_k  \alpha_k (m_{kj} - x_j)\Big)  &\text{cancelling~} Z^2 \nonumber\\
&= \sum_k m_{ki} \alpha_k (m_{kj} - x_j) - \Big(\sum_k m_{ki}\alpha_k\Big) 
\Big(\big(\sum_k  \alpha_k m_{kj}\big) - x_j\Big) & \text{since ~} \sum_k \alpha_k x_j = x_j \nonumber\\
&= \sum_k m_{ki} \alpha_k m_{kj} - \Big(\sum_k m_{ki}\alpha_k\Big) 
\Big(\sum_k  \alpha_k m_{kj}\Big)  &\text{cancelling~} x_j\nonumber\\
&= \sum_k \alpha_k m_{ki} m_{kj} - y_iy_j ~. \label{eq:dyi/dxj}
\end{align}
Finally, writing $\tilde{\vm}_k = \vm_k - \vy$,
this gives 
\begin{equation}
\label{eq:jacobian-simple}
\frac{\partial y_i}{\partial x_j} = \frac{1}{\sigma^2}\sum_k \alpha_k \tilde m_{ki} \tilde m_{kj}~.
\end{equation}
This shows the interesting fact that the Jacobian is symmetric positive-semidefinite,
and can be written in a way depending only on $\vy = h(\vx)$, but not on $\vx$ directly.

A further interesting result can be derived from \eqref{eq:jacobian-simple}.
Let $d\vx$ be an infinitessimal change in $\vx$.  The corresponding infinitessimal
change in $\vy$ is computed by
\begin{align*}
d y_i &= \frac{\partial y_i}{\partial \vx} ~ d\vx \\
      &= \sum_j \frac{\partial y_i}{\partial x_j} ~ d x_j \\
      &= \frac{1}{\sigma^2} \sum_k \alpha_k \tilde m_{ki} \sum_j \tilde m_{kj} dx_j & \text{after interchanging summation order}\\
      &= \frac{1}{\sigma^2} \sum_k \alpha_k \tilde m_{ki} ~\big<\tilde{\vm}_k,   d \vx \big> ~.
\end{align*}
Now, if $d\vx$ is perpendicular to the simplex $\Delta$, then it is perpendicular to
each $\tilde{\vm}_k$, so the inner product vanishes.  Consequently, $d y_i$ vanishes for all $i$,
and $d \vy = 0$.  This shows that if $\vx$ varies in a direction perpendicular to
the simplex, then the value of $\vy = h(\vx)$ does not change.  This is a verification
of the fact that $h(\vx)$ is constant on {\em fibres} perpendicular to the simplex $\Delta$.

The formula \eqref{eq:dyi/dxj}
can be written neatly using matrices.  Suppose that $\mM$ is the matrix
with columns $\vm_k$.  Then we have
\begin{align}
\begin{split}
\sigma^2 \frac{\partial \vy}{\partial \vx} &=  \mM {\rm diag}(\valpha) \mM^T - \vy \vy^T\\
  &= \mM ({\rm diag}(\valpha) - \valpha \valpha^T) \mM^T ~,
\end{split}
\end{align}

\paragraph{Lipschitz condition  -- $L_1$. }
A Lipschitz constant for a function $\vy = h(\vx)$ with respect to some norm $\| \cdot \|$
is a constant $K$ such that
\[
\| \vy - \vy'\| \le K \| \vx - \vx'\| ~.
\]
For a differentiable function a Lipschitz constant for the $L_1$ vector norm is given by 
\[
K = \max_j \| \partial \vy / \partial x_j \|_1
\]
evaluated over the domain of the function (a single Voronoi cell for instance).

From \eqref{eq:jacobian-simple} we have 
\begin{align*}
K &= (1/\sigma^2) \max_j \Big\| \sum_k \alpha_k \tilde {\vm}_k~ \tilde m_{kj} \Big\|_1  \\
  &\le (1/\sigma^2) \max_j \sum_k \|\alpha_k \tilde {\vm}_{k} \tilde m_{kj}  \|_1 & \text{triangle inequality}\\
  &\le (1/\sigma^2) \sum_k \max_j \|\alpha_k \tilde {\vm}_{k} \tilde m_{kj}  \|_1& 
  \text{interchanging } \max \text{ and sum}\\
  &= (1/\sigma^2) \sum_k  \max_j \alpha_k |\tilde m_{kj}|~ \| \tilde {\vm}_{k}   \|_1& 
  \text{taking scalars outside the norm}\\
    &= (1/\sigma^2) \sum_k  \alpha_k \| \tilde {\vm}_{k}   \|_1 \max_j |\tilde m_{kj}| & 
  \text{taking terms outside } \max_j \\
   &= (1/\sigma^2)  \sum_k \alpha_k \| \tilde {\vm}_{k}\|_1 ~\| \tilde {\vm}_{k}\|_\infty \\
   &\le (1/\sigma^2)  \max_k  \| \tilde {\vm}_{k}\|_1 ~\| \tilde {\vm}_{k}\|_\infty &\text{ since } \sum_k \alpha_k = 1
\end{align*}
This leads to
\[
K  \le \frac{\max_k  \| \tilde {\vm}_{k}\|_1^2}{\sigma^2}
\]

Since $\tilde{\vm}_k = \vm_k - \vy$ and $\vy$ can vary, we see that
$\|\tilde{\vm}_k \|^2 \le D^2$ where $D$ (with respect to the given norm) is the diameter of the simplex with 
vertices $\vm_k$.  (The diameter of a set is the supremum of distances between two points in the set.)
This gives finally
\[
K \le \frac{D^2}{\sigma^2} ~.
\]
This will be true for any norm that is greater than the $\infty$-norm for all points.
However, it therefore applies to any norm that is greater than a multiple of the $\infty$-norm. 
Therefore we can conclude Proposition 2.4 as follows.
%
%

\vspace{0.5cm}
\noindent
\textbf{Proposition 2.4}: 
For any norm $\|\cdot \|$ on $\mbr{d}$ equivalent to 
the $2$-norm, a Lipschitz constant for the LCSA mapping on a given Voronoi region
is given by
\[
K = \frac{D^2}{\sigma^2}
\]
where $D$ is the diameter of the simplex $\Delta$.

\paragraph{Lipschitz condition -- $L_2$. }

For the $L_2$ norm a Lipschitz constant is given by
\[
K = \max_{\|v\|} ~ \frac{\|J \vv \|}{\| \vv \|} 
\]
where $\vv$ is a vector and $J = \partial y_i / \partial x_j$ is the Jacobian.
In the case where $J$ is symmetric positive definite (such as in the
present case), this is equal to 
\begin{equation}
K = \max_{\|v\| = 1}  \vv^T J \vv ~.
\label{eq:jakob1}
\end{equation}
With $J = \partial y_i / \partial x_j = \sum_k \alpha_k \tilde m_{ki} \tilde m_{kj} / \sigma^2$ as in \eqref{eq:jacobian-simple} this becomes
\begin{align*}
\vv^T J \vv &= (1/\sigma^2)~\sum_k \alpha_k \big<\tilde{\vm}_k, \vv \big>^2 \\
              &\le (1/\sigma^2)~ \max_k \big<\tilde{\vm}_k, \vv \big>^2 ~.
\end{align*}
This quantity is maximized, over vectors $\vv$ of norm $1$ when
$\vv = \arg \max_k \| \tilde{\vm}_k \|$ (up to scale), in which case it is equal to $\max_k \| \tilde{\vm}_k \|^2/\sigma^2$.
This shows that when $\| \vv\| = 1$, 
\begin{align}
\vv^T J \vv  &\le \max_k \| \tilde{\vm}_k\|^2 / \sigma^2\nonumber\\
               &\le D^2 / \sigma^2.\label{eq:jakob2}
\end{align}
This shows that the same Lipschitz constant $K = D^2 / \sigma^2$ holds for the $L_2$ as for the $L_1$ Lipschitz condition.

\vspace{0.5cm}
\noindent\textbf{Proof of Claim 5.} The LCSA encoding $\mM \valpha(\vx)$ is non-continuous at the boundaries of the Voronoi regions simply because the set of $k'$ nearest neighbor anchors of $\vx$ changes as $\vx$ crosses from  one Voronoi region to another the Voronoi region.


\vspace{0.5cm}
\noindent\textbf{Proof of Claim 7.} 
 We penalise the spectral norm of Jacobian matrix $\Big\lVert\frac{\partial\mM\boldsymbol{\alpha}(\vx)}{\partial\vx}\Big\rVert_2\!=\!K$ via de facto controlling the Lipschitz constant $K\!=\!{D^2}/{\sigma^2}$ because this is shown in \eqref{eq:jakob1} and \eqref{eq:jakob2}.

\vspace{0.5cm}
 Claims 1--8  highlight that LCSA can perform the feature quantization in some regions of the feature space (proximity of dictionary atoms), as well as it can  perform the approximate
linear coding akin to linear coding methods (the proximity of the mean of atoms of Voronoi cell). The  Lipschitz constant highlights that our $\sigma$ controls denoising achieved via LCSA due to the mechanism akin to  DAE (whose denoising effect is controlled by its $\sigma'$). Our blocks in the GAN
discriminator can guide the dictionary atoms and conv. features towards quantization
 or reconstruction with higher linear fidelity, also selecting some intermediate denoising hypothesis as  the  Lipschitz constant poses the upper bound controlling denoising effect.

\twocolumn